%% file: acl_latex.tex
\pdfoutput=1

\documentclass[11pt]{article}

\usepackage[preprint]{acl}

\usepackage{times}
\usepackage{latexsym}


\usepackage{url}

\usepackage[utf8]{inputenc} 
\usepackage[T1]{fontenc}    
\definecolor{cvprblue}{rgb}{0.21, 0.49, 0.74}
\usepackage{booktabs}       
\usepackage{amsfonts}       
\usepackage{nicefrac}       
\usepackage{microtype}      

\usepackage{inconsolata}

\usepackage{threeparttable}
\usepackage{graphicx}
\usepackage{multirow}
\usepackage{makecell}
\usepackage{enumitem}
\usepackage{amsmath}
\usepackage{amsthm}
\usepackage{wrapfig}
\usepackage{subcaption}

\usepackage{tikz}
\usepackage{arydshln}

\title{Can Graph Neural Networks Learn Language \\with Extremely Weak Text Supervision?}

\author{
 \textbf{Zihao Li\textsuperscript{1}},
 \textbf{Lecheng Zheng\textsuperscript{1}},
 \textbf{Bowen Jin\textsuperscript{1}},
 \textbf{Dongqi Fu\textsuperscript{2}},
\\
 \textbf{Baoyu Jing\textsuperscript{1}},
 \textbf{Yikun Ban\textsuperscript{1}},
 \textbf{Jingrui He\textsuperscript{1}},
 \textbf{Jiawei Han\textsuperscript{1}}
\\
\\
 \textsuperscript{1}University of Illinois Urbana-Champaign,
 \textsuperscript{2}Meta AI
\\
 \small{
   \textbf{Correspondence:} \href{mailto:zihaoli5@illinois.edu}{zihaoli5@illinois.edu}
 }
}

\newcommand{\cm}{\mathcal}
\newcommand{\bm}{\mathbf}

\begin{document}
\maketitle
\begin{abstract}
    While great success has been achieved in building vision models with Contrastive Language-Image Pre-training (CLIP) over internet-scale image-text pairs, building transferable Graph Neural Networks (GNNs) with CLIP pipeline is challenging because of the scarcity of labeled data and text supervision, different levels of downstream tasks, and the conceptual gaps between domains. 
    In this work, to address these issues, we propose a multi-modal prompt learning paradigm to effectively adapt pre-trained GNN to downstream tasks and data, given only a few semantically labeled samples, each with extremely weak text supervision. 
    Our new paradigm embeds the graphs directly in the same space as the Large Language Models (LLMs) by learning both graph prompts and text prompts simultaneously.dai
    We demonstrate the superior performance of our paradigm in few-shot, multi-task-level, and cross-domain settings. Moreover, we build the first CLIP-style zero-shot classification prototype that can generalize GNNs to unseen classes with extremely weak text supervision. The code is available at \url{https://github.com/Violet24K/Morpher}.
\end{abstract}

\input{0_sections/a_introduction}

\input{0_sections/background}

\input{0_sections/Improved_graph_grompt}

\input{0_sections/multimodal}

\input{0_sections/experiment}

\input{0_sections/related_work}

\input{0_sections/z_conclusion_discussion}

\bibliography{reference}

\clearpage

\begin{figure*}[t!]
    \centering
    \includegraphics[width=0.9\textwidth]{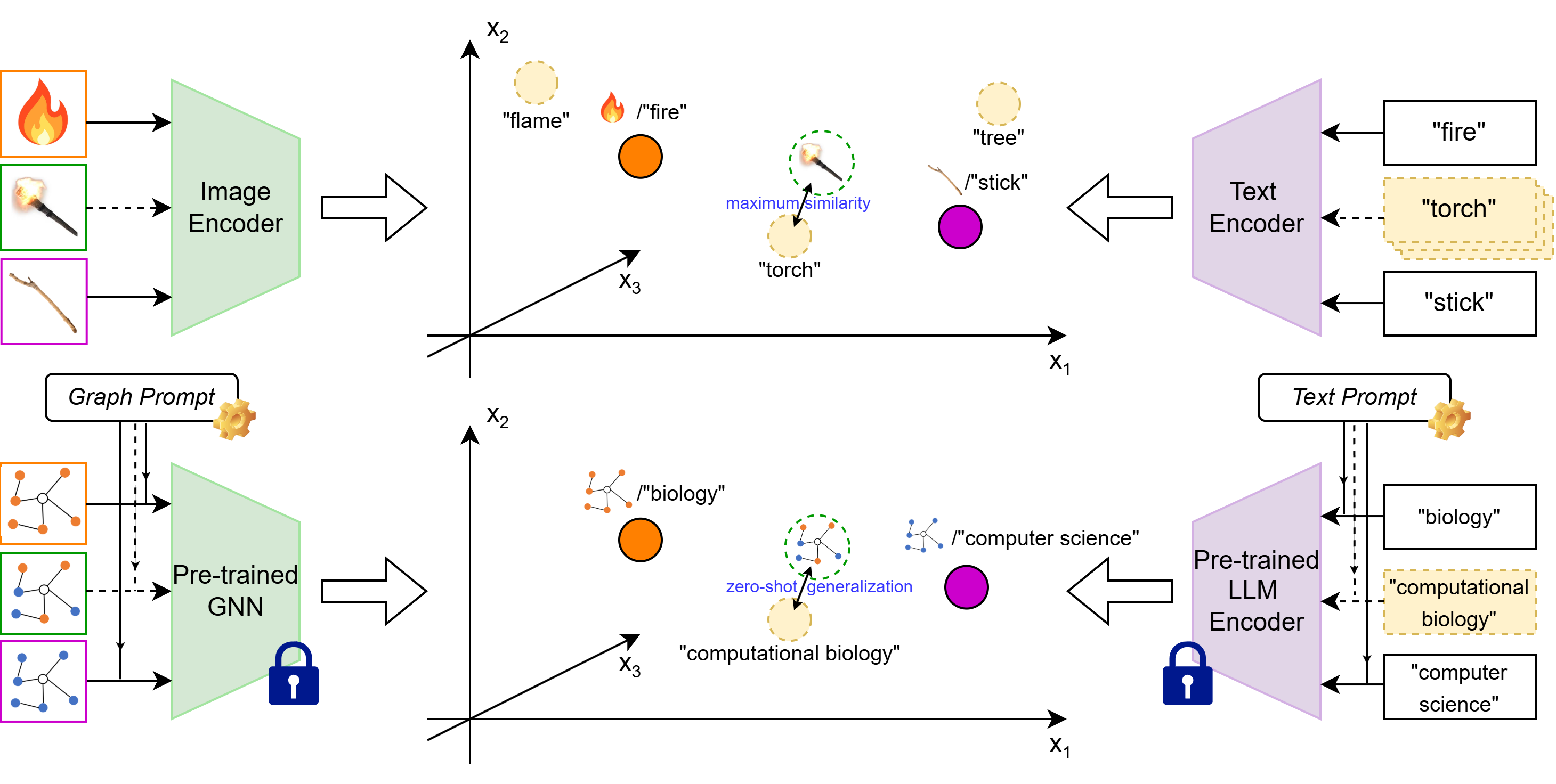}
    \caption{CLIP backbone (top) and this work (bottom). If a research paper cites many papers from biology and computer science, we realize this paper will likely be about computational biology, even if we do not know what exactly computational biology is. CLIP builds image encoders that learn such language dependency by Contrastive Language-Image Pre-training in the same embedding space according to Internet-scale data. However, text supervision is often extremely weak for graphs. This work leverages Multi-modal Prompt Learning for Graph Neural Networks that can effectively teach GNNs language dependency given few training samples with weak text supervision.}
\end{figure*}

\appendix

\input{appendix}

\end{document}

%% file: 0_sections/a_introduction.tex
\section{Introduction}
Graphs are constructed from real scenarios, but GNNs, optimized according to numerical labels, still do not \textit{understand} what a label represents in the real world. To solve the issue of predetermined numerical categories, CLIP \citep{DBLP:conf/icml/RadfordKHRGASAM21} leverages natural language supervision by jointly training an image encoder and a text encoder in the same embedding space at scale. CLIP has demonstrated the ability to train high-quality, generalizable vision models \citep{DBLP:conf/icml/RadfordKHRGASAM21, DBLP:conf/icml/JiaYXCPPLSLD21, DBLP:conf/icml/0001LXH22}, which can adapt to diverse downstream tasks. Similar frameworks have been successfully extended to video \citep{DBLP:conf/emnlp/XuG0OAMZF21}, 3D images 
\citep{DBLP:conf/wacv/HessTPAS24},
speech \citep{DBLP:conf/slt/ShihWCBLH22} and audio \citep{DBLP:conf/icassp/GuzhovRHD22}, consistently demonstrating that alignment with text enhances the transferability of encoders. 
As for graphs, so far, such graph-text alignment has only been explored in the molecular domain \citep{DBLP:journals/corr/abs-2307-09484, DBLP:conf/emnlp/LiuLL00K0C23} and on text-attributed graphs \citep{DBLP:conf/sigir/Wen023, DBLP:conf/emnlp/0001DL23, DBLP:conf/acl/JinZZ000023, DBLP:conf/nips/YanLLY0ZYZHSDZ023}, where the paired graph-text data is relatively sufficient for joint pre-training. 

However, extending this paradigm to more general graph data poses significant challenges due to three facts. First, compared with language or vision data
, graph data is very scarce and the text supervision is extremely weak \citep{DBLP:journals/corr/abs-2310-11829, DBLP:journals/corr/abs-2312-10073, DBLP:conf/log/ManchandaGRB23}. Besides the number of samples being much smaller than images, many graph datasets are used for classification, where the label names consist of only a few tokens. 
Second, the task space of graph data could be on node-level, edge-level, and graph-level. 
Third, in general, language tokens and visual objects retain the same conceptual meaning across different distributions, but the same graph structure may have distinct interpretations in different domains.

Jointly pre-training graph and text encoders is impractical for graph data with extremely weak text supervision. Fortunately, we can deal with the two modalities separately for pre-training: large language models have already been extensively pre-trained, and tremendous efforts have been devoted to pre-train GNNs through self-supervision \citep{DBLP:conf/iclr/HuLGZLPL20, DBLP:journals/tkde/LiuJPZZXY23, DBLP:conf/kdd/ZhengJLTH24, DBLP:journals/corr/abs-2412-21151}. However, even with a pre-trained graph model, effectively adapting it to both the semantic embedding space for text alignment and diverse downstream tasks remains non-trivial. This raises a critical question:
\begin{center}
\emph{How to adapt pre-trained GNNs to the semantic embedding space given limited downstream data, i.e., few samples and weak text supervision?}
\end{center}
This paper aims to answer this question based on the following observations: (1) Semantic text embedding spaces do not necessarily result from joint pre-training. In fact, the embedding spaces of encoder LLMs are inherently semantic and high-quality, as LLMs are trained on massive text data and demonstrate strong performances. 
(2) When the downstream data are limited, prompt learning \citep{DBLP:conf/acl/LiL20, DBLP:conf/icml/HoulsbyGJMLGAG19, DBLP:conf/iclr/ZhangLCDBTHC22, DBLP:conf/emnlp/LesterAC21} provides a better option than fine-tuning as much fewer parameters not only makes the optimization more efficient but also requires less resource than fine-tuning a large model. Notably, some works have explored prompt learning for better alignment and obtained improvement in vision prediction \citep{DBLP:conf/cvpr/ZhouYL022, DBLP:conf/cvpr/KhattakR0KK23}. Inspired by these observations, 
we propose a prompting-based paradigm with an LLM that aligns the GNN representations in the semantic embedding space, while keeping the parameters of both GNN and LLM frozen.

When adapting the representation from one modality to another, solely prompting a single modality could be sub-optimal, as it limits the adjustment to downstream tasks in the other modality \citep{DBLP:conf/cvpr/KhattakR0KK23}. To this end, we propose \underline{M}ulti-m\underline{o}dal P\underline{r}ompt Learning for Gra\underline{\smash{ph}} N\underline{e}ural Netwo\underline{r}ks (Morpher). Given a pre-trained GNN and few-shot semantically labeled graph data with weak text supervision, we assume zeroth-order access to a pre-trained LLM. Then, to leverage its high-quality semantic embedding space, Morpher connects and aligns the graph embeddings to it through prompting on both modalities with a cross-modal projector. Nonetheless, designing such a paradigm is more challenging than vision-language models. First, we lack jointly pre-trained encoders for the two modalities; instead, we only have two encoders pre-trained independently in each modality. Second, determining how to prompt the graph modality is non-trivial and remains a trending research topic. Third, the downstream data for GNN usually have much fewer labeled classes and labeled samples than Vision-Language models, and the text supervision is extremely weak.  
Our contributions towards tackling these challenges are:
\begin{itemize}[leftmargin=*]
  \item We analyze that, state-of-the-art graph prompt \citep{DBLP:conf/kdd/SunCLLG23} is often unable to learn good representations of the downstream data. We further improve it to prevent unstable optimization.
  \item To connect and adapt the pre-trained GNN with LLM effectively with extremely weak text supervision, we propose Morpher, the first graph-text multi-modal prompt learning paradigm to align the representations of GNN and LLM without fine-tuning any of their parameters.
  \item With extremely weak text supervision, we demonstrate our improved graph prompt and Morpher under few-shot, multi-task, and cross-domain settings. To show that GNN learns language dependency through Morpher, we present the first CLIP-style zero-shot generalization prototype where the GNN can predict unseen classes.
\end{itemize}

%% file: 0_sections/background.tex
\section{Background}
\label{sec: background}

\begin{figure*}[t]
\centering
\includegraphics[width=\linewidth]{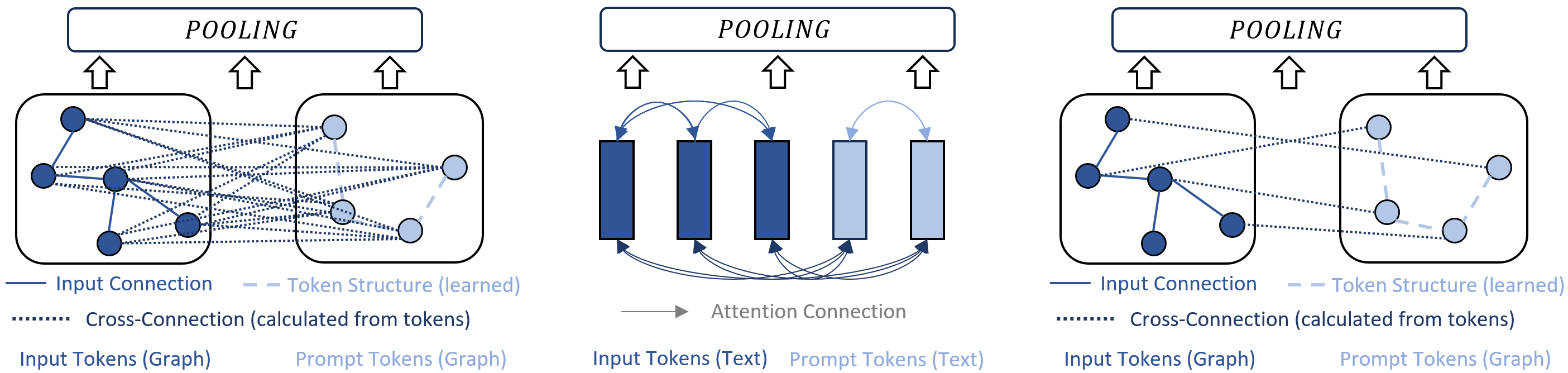}
\caption{
\small {Cross-connections overwhelm inner-connections in current graph prompt design, which may be unstable during training (left); attention in NLP where $3\times 2 = 6$ cross-connections and $3 + 1 = 4$ inner-connections are balanced (middle); and our balanced graph prompt design (right). \bf The cross-connections between input and prompt should have a consistent scale with the input connections.}
}
\label{fig: mechanism}
\end{figure*}

We use calligraphic letters (e.g., $\mathcal{A}$) for sets, and specifically $\cm{G}$ for graphs. We use bold capital letters for matrices (e.g., $\bm{A}$). For matrix indices, we use $\bm{A}(i, j)$ to denote the entry in the $i^{th}$ row and the $j^{th}$ column. $\bm{A}(i, :)$ is the $i^{th}$ row in $\bm{A}$.

\paragraph{Graph Neural Networks.} We use $\cm{G} = (\bm{A}, \bm{X})$ to denote a graph with node set $\cm{V}$ and edge set $\cm{E}$,
where $\bm{A} \in \mathbb{R}^{|\cm{V}| \times |\cm{V}|}$ is the adjacency matrix and $\bm{X}\in\mathbb{R}^{|\cm{V}|\times d}$ is the node feature matrix. $\bm{A}(u, v)=1$ if there is an edge connecting $u$ and $v$; otherwise $\bm{A}(u, v) = 0$.
A graph neural network $f_\phi^g(\cdot)$ with hidden dimension $d_g$ encodes $\cm{G}$ into the embedding space: $f_\phi^g(\cm{G})\in\mathbb{R}^{|\cm{V}| \times d_g}$, which could preserve both feature and structure information of $\cm{G}$.

\paragraph{Few-shot Prompt Learning.}
Let $f_\phi^t(\cdot)$ denote the LLM encoder with embedding dimension $d_t$. 
For a series of input tokens $\{x_k\}_{k=1}^K$, the LLM encoder embeds it as a matrix $\bm{X}_t$ = $f_\phi^t(\{x_k\}_{k=1}^K) \in \mathbb{R}^{K\times d_t}$.
Prompt learning initializes a tunable matrix $\bm{P}_{\theta}^t \in \mathbb{R}^{n_t \times d_t}$, where $n_t$ denotes the number of text prompt tokens.
Then, this tunable matrix is concatenated with the input tokens' embeddings to form $[\bm{P}_{\theta}^t; \bm{X}_t]_{dim=0} \in \mathbb{R}^{(K+n_t) \times d_t}$.

\paragraph{Our Problem Set-up.} Given a pre-trained GNN $f_\phi^g(\cdot)$ with embedding dimension $d_g$ and a pre-trained LLM encoder $f_\phi^t(\cdot)$ with embedding dimension $d_t$. Without loss of generality, we assume the downstream task is graph-level classification, as node-level or edge-level GNN tasks can be reformulated as graph-level by inducing ego-graphs within neighbor distance $\gamma$. For $L$-shot graph classification, we are given limited text-labeled pairs $\{(\cm{G}_i, t_c)\}_{i=1}^L$ for each class $c$. Each text label $t_c$ consists of only a few tokens.
Assuming $\cm{T}$ is the set of all text labels $t_c$, we are provided a set of test graphs $\{\cm{G}_j\}_{j=1}^{L_{test}}$ and want to predict the text label $t_j \in \cm{T}$ for each test graph $\cm{G}_j$.

%% file: 0_sections/Improved_graph_grompt.tex
\section{Improving Single-modal Graph Prompt}
\label{sec: graph prompt}

Unlike prompting text data, prompting graph data presents a significant challenge due to the non-euclidean nature of graphs. The pioneering work \citep{DBLP:conf/kdd/SunCLLG23} designs the graph prompt still as a graph and then inserts it into the original graph. 

\paragraph{Current Graph Prompt Design.} To prompt a graph $\cm{G}$, each prompt token is a new node. 
Let $n_g$ denote the number of prompt tokens and $\cm{P} = \{p_i\}_{i=1}^{n_g}$ denote the set of prompt tokens. The graph prompt is formulated by a tunable matrix $\bm{P}_\theta^g \in \mathbb{R}^{n_g \times d}$, where $d$ is the node feature dimension. Each row vector $\bm{P}_\theta^g(i, :)$ is the feature of the prompt token $p_i$. Then, the mechanism to prompt a graph $\cm{G} = (\bm{A}, \bm{X})$ with $n$ nodes and $d$ feature dimension is \cite{DBLP:conf/kdd/SunCLLG23}
\begin{itemize}[leftmargin=*]
    \item Compute inner-connections to construct the prompt graph $\cm{G}_p = (\bm{A}_p, \bm{X}_p) = (\bm{A}_p, \bm{P}_\theta^g)$. For two prompt tokens $p_i$ and $p_j$, $\bm{A}_p(i, j) = 1 \iff \sigma(\bm{P}_\theta^g(i, :) \bm{P}_\theta^g(j, :)^\top) > \delta_{inner}$, where $\sigma(\cdot)$ is the sigmoid function.
    \item Compute cross-connections to insert the prompt graph $\cm{G}_p$ into $\cm{G}$. Similarly, for $x_i \in \cm{G}$ and $p_j \in \cm{G}_p$, there is an edge between them if and only if $\sigma(\bm{X}(i, :) \bm{P}_\theta^g(j, :)^\top) > \delta_{cross}$.
    \item Construct the prompted graph (i.e., manipulated graph) $\cm{G}_m = (\bm{A}_m, \bm{X}_m)$. 
    The overall adjacency matrix $\bm{A}_m \in \mathbb{R}^{(n + n_g) \times (n + n_g)}$ is constructed from the original adjacency matrix $\bm{A}$, the inner edges $\bm{A}_p$ and the cross edges. The overall node feature matrix is concatenated from the prompt token features and the original input node features: $\bm{X}_m = [\bm{P}_{\theta}^g; \bm{X}]_{dim=0} \in \mathbb{R}^{(n+n_g) \times d}$.
\end{itemize}

\paragraph{Issues associated with the current design.} The input node features of most real-world datasets are sparse, resulting from the construction process \citep{DBLP:conf/icml/YangCS16, DBLP:journals/corr/abs-2007-08663, DBLP:journals/jmlr/DwivediJL0BB23}. As shown in Appendix Table \ref{tab: data}, $||\bm{X}(i, :)||_1$ is typically 1.
As the initialization of each token feature tensor $\bm{P}_\theta^g(i, :)$ is close to $\vec{\boldsymbol{0}}$ to stabilize gradients, for any node $i$ and token $p_j$, the dot products $\bm{X}(i, :) \bm{P}_\theta^g(j, :)^\top$ is close to $0$, and the sigmoid value is very close to $0.5$. 
Consequently, if we want the graph prompt to have cross-connections, we must set $\delta_{cross} < 0.5$. Then, as the sigmoid values are close to 0.5, the cross-connections will be dense, i.e., almost every node in the original graph is connected with every node token in the prompt graph. For two different graphs $\cm{G}_1$ and $\cm{G}_2$ in the same task, the prompt graph $\cm{G}_p$ is identical. 
Since the GNNs aggregate the node features, their embeddings $f_\phi^g(\cm{G}_1)$ and $f_\phi^g(\cm{G}_2)$ are approximately the same because the features in the prompt graph overwhelm the features in the original graphs due to the dense cross-connections. Then, even if $\cm{G}_1$ and $\cm{G}_2$ have different labels, the task head classifier cannot be trained to distinguish them
\footnote{In fact, similar training instability problems have been observed by another work \citep{DBLP:conf/kdd/ZhaoYCRZDKZ024}.}. 

In Appendix \ref{ap: current graph prompt design}, we show that initializing graph prompt token feature tensor with higher variance cannot effectively address this problem.

\newcommand{\sursf}{\sigma}
\newcommand{\Attn}{{\rm Attn}}
\newcommand{\wt}{\widetilde}
\newcommand{\defeq}{\mathrel{\mathop:}=}
\newcommand{\paren}[1]{{\left( #1 \right)}}
\newcommand{\R}{\mathbb{R}}
\newcommand{\sets}[1]{{\{ #1 \}}}

\begin{figure*}[t]
\centering
\includegraphics[width=\linewidth]{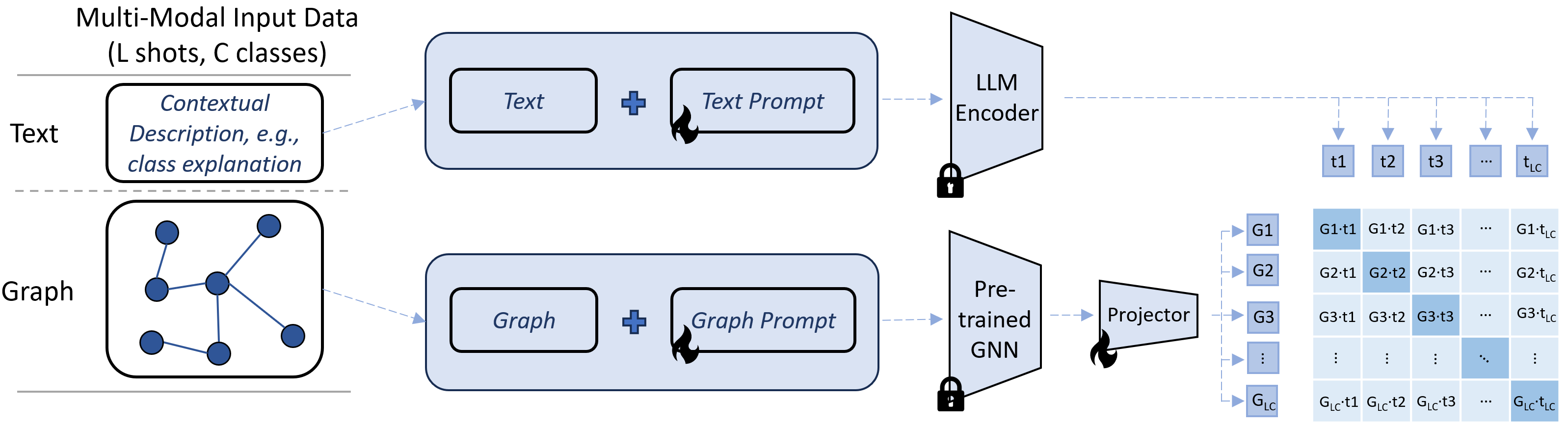}
\caption{
\small {Similar to CLIP backbone, Morpher adapts the graph representations to semantic space through multi-modal prompt learning, even if the GNN and LLM are not jointly trained and are kept frozen.}
}
\label{fig: morpher illustration}
\end{figure*}

\paragraph{Improved Graph Prompt Design.}
The issue of the current graph prompt lies in the significant imbalance between original connections within the input graph and the input-prompt cross-connections, as illustrated in Figure \ref{fig: mechanism} (left).
We also visualize the standard NLP attention mechanism \citep{DBLP:conf/nips/VaswaniSPUJGKP17} in Figure \ref{fig: mechanism} (middle). 
When a sequence of text prompt tokens ${p_i^t}$ is prepended to the input sequence, the features of the prompt tokens will be aggregated with those of the input tokens through a dense ``cross-connection'', i.e., attention.
Simultaneously, the features within the input sequence are also densely aggregated via attention, maintaining a balance with the prompt-input aggregation to prevent overwhelming. 
Inspired by this, we deem that a balance could be achieved by approximately equalizing the number of cross-connections with that of input graph connections, i.e., $n_e$. Since the connection of a graph dataset is often sparse, we constrain the cross-connections to be sparse as well. Therefore, we set the number of cross-connections to at most $n_e$ by connecting each node in the input graph with at most $\left \lfloor{\frac{n_e}{a}}\right \rfloor$ prompt tokens.
Then, we can safely use a small $\delta_{cross}$ and cosine similarity $\frac{\bm{X}(i, :)\cdot\bm{P}_\theta^g(j, :)^\top}{\|\bm{X}(i, :)\|_2 \|\bm{P}_\theta^g(j, :)\|_2}$ instead of $\sigma(\bm{X}(i, :) \bm{P}_\theta^g(j, :)^\top)$ to calculate the cross-connections.

%% file: 0_sections/multimodal.tex
\newcommand{\Proj}{{\rm Proj}}
\newcommand{\Token}{{\rm Tokenize}}
\newcommand{\readout}{{\rm readout}}

\section{GNN Multi-modal Prompt Learning}
To adapt the GNN embeddings to the LLM's semantic embedding space and leverage the additional weak supervision provided by the text associated with graph labels, we explore the potential of multi-modal prompt learning for both graphs and language. This approach is motivated by the intuition that only prompting on the graph data may limit the flexibility to adjust the LLM representation space. The overall paradigm of Morpher is illustrated in Figure \ref{fig: morpher illustration}. 
Given the data $\{(\cm{G}_i, t_i)\}_{i=1}^{L\times C}$, we aim to align graph embedding $\readout(f_\phi^g(\cm{G}_i))$ with $\readout(f_\phi^t(\Token(t_i)))$. Yet one direct issue is that, $\readout(f_\phi^g(\cm{G}_i)) \in \R^{1\times d_g}$ and $\readout(f_\phi^t(\Token(t_i))) \in \R^{1\times d_t}$ may have distinct dimensions. To address this issue, we adopt a cross-modal projector that learns to map the graph embedding space to the text embedding space. For an input $d_g$-dimensional graph embedding $\bm{v}$, the projector maps it to a vector $\wt{\bm{v}}$ in the $d_t$-dimensional text embedding space:
\begin{equation}
    \wt{\bm{v}} = \Proj_\theta(\bm{v}) \defeq \tanh (\bm{W}\bm{v} + \bm{b}) \in \R^{1 \times d_t}
\end{equation}
As discussed in Sections \ref{sec: background} and \ref{sec: graph prompt}, we introduce the text prompt $\bm{P}_{\theta}^t \in \mathbb{R}^{n_t \times d_t}$ with $n_t$ text prompt tokens and the graph prompt $\bm{P}_\theta^g \in \mathbb{R}^{n_g \times d}$ with $n_g$ graph prompt tokens. The graph prompting function $\psi_g(\cdot, \bm{P}_{\theta}^g)$ modifies a given graph $\cm{G}$ into a manipulated graph $\cm{G}m = \psi_g(\cm{G}, \bm{P}_{\theta}^g)$.

Let $\omega_t(\cdot, \bm{P}_{\theta}^t)$ be the prompted text embedding given input text $t$. For the text prompt methods we choose, the prompted embedding is
\begin{equation}
    \omega_t(t, \bm{P}_{\theta}^t) = [\bm{P}_{\theta}^t; f_\phi^t(\Token(t))]_{dim=0}
\end{equation}
Let $\omega_g(\cdot, \bm{P}_{\theta}^g)$ be the prompted graph embedding given input graph $\cm{G}$, then we have:
\begin{equation}
    \omega_g(\cm{G}, \bm{P}_{\theta}^g) = f_\phi^g(\cm{G}_m) = f_\phi^g(\psi_g(\cm{G}, \bm{P}_{\theta}^g))
\end{equation}
For the whole prompted text and the whole prompted graph of the sample $(\cm{G}_i, t_i)$, we apply readout (e.g., mean-pooling, max-pooling, etc.) to get their embedding:
\begin{equation}
    \bm{h}_i^t = \readout(\omega_t(t_i, \bm{P}_{\theta}^t)) \in \R^{1 \times d_t}
\end{equation}
\begin{equation}
    \bm{h}_i^\cm{G} = \readout(\omega_g(\cm{G}_i, \bm{P}_{\theta}^g)) \in \R^{1 \times d_g}
\end{equation}
For the given data $(\cm{G}_i, t_i)$, we compute the normalized embedding of prompted $\cm{G}_i$ and project it to the text embedding space through the projector:
\begin{equation}
    \bm{z}_{norm, i}^{\cm{G}} =\frac{\bm{h}_i^\cm{G}}{||\bm{h}_i^\cm{G}||_2} = \frac{\readout(\omega_g(\cm{G}_i, \bm{P}_{\theta}^g))}{||\readout(\omega_g(\cm{G}_i, \bm{P}_{\theta}^g))||_2}
\end{equation}
\begin{equation}
    \bm{z}_i^\cm{G}=\Proj_\theta(\bm{z}_{norm, i}^{\cm{G}})
\end{equation}
For the text embeddings, since for limited data the set $\cm{T} = \{t_i\}_{i=1}^C$ may contain texts that are semantically close as discussed in Appendix \ref{ap: text labels}, we extract a subspace in the text embedding space by normalizing the embedding as follows. We further normalize the text embeddings to the unit sphere.
\begin{equation}
    \mu = \frac{1}{L} \sum_{i=1}^L \bm{h}_i^t, \quad \bm{h}_{norm, i}^t = \bm{h}_i^t - \mu
\end{equation}
\begin{equation}
\label{eq: mu of text}
    \bm{z}_i^t =\frac{\bm{h}_{norm, i}^t}{||\bm{h}_{norm, i}^t||_2} = \frac{\readout(\omega_t(t_i, \bm{P}_{\theta}^t)) - \mu}{||\readout(\omega_t(t_i, \bm{P}_{\theta}^t)) - \mu||_2}
\end{equation}
Finally, we use the in-batch similarity-based contrastive loss with temperature $\tau$ to train text prompts, graph prompts, and the projector.
\begin{equation}
\label{eq: z of text}
    \mathcal{L}_{G\rightarrow T} = -\frac{1}{B} \sum_{i=1}^B \log \frac{\exp(\bm{z}_i^\cm{G} \cdot \bm{z}_i^t / \tau)}{\sum_{j=1}^B\exp(\bm{z}_i^\cm{G} \cdot \bm{z}_j^t / \tau)}
\end{equation}
During inference stage, for an input graph $\cm{G}_i$ and text label candidates $\cm{T} = \{t_i\}_{i=1}^C$, we compute the embedding $\bm{z}_i^\cm{G} = \Proj_\theta(\frac{\readout(\omega_g(\cm{G}_i, \bm{P}_{\theta}^g))}{||\readout(\omega_g(\cm{G}_i, \bm{P}_{\theta}^g))||_2})$ using trained $\bm{P}_{\theta}^g$ and $\Proj_\theta(\cdot)$. Then, we compute $\bm{z}^t_i$ as Equations \ref{eq: mu of text} and \ref{eq: z of text}. Finally, $\cm{G}_i$ will be classified to associate with text label ${\arg \max}_{1 \leq i \leq C} (\bm{z}_i^\cm{G} \cdot \bm{z}_i^t)$.

%% file: 0_sections/experiment.tex
\section{Experiments}

\input{tables/few_shot_learning}
We show that both Morpher and the improved graph prompt more effectively adapt pre-trained GNNs to the specific downstream classification task.
We use RoBERTa \citep{DBLP:journals/corr/abs-1907-11692} as the LLM encoder for Morpher in the main experiments. We also validate the performance of Morpher with ELECTRA \citep{DBLP:conf/iclr/ClarkLLM20} and DistilBERT \citep{DBLP:journals/corr/abs-1910-01108} in Appendix \ref{ap: other llm}.

\textit{Datasets.} We use real-world graph datasets from PyTorch Geometric \citep{DBLP:journals/corr/abs-1903-02428}, including one molecular dataset MUTAG \citep{DBLP:journals/corr/abs-2007-08663}; two bioinformatic datasets ENZYMES and PROTEINS \citep{DBLP:conf/ismb/BorgwardtOSVSK05}; one computer vision dataset MSRC\_21C \citep{DBLP:journals/ml/NeumannGBK16}; three citation network datasets Cora, CiteSeer and PubMed \citep{DBLP:conf/icml/YangCS16}. We use real-world class names as text labels. The text supervision is extremely weak, as each text label contains no more than five words. More details are summarized in Appendix \ref{ap: dataset}. 

\textit{Pre-training algorithms and GNN backbones.} To pretrain GNNs for evaluation, we adopt GraphCL \citep{DBLP:conf/nips/YouCSCWS20} and SimGRACE \citep{DBLP:conf/www/XiaWCHL22} to pre-train three widely used GNN backbones: GCN \citep{DBLP:journals/corr/KipfW16}, GAT \citep{DBLP:conf/nips/YunJKKK19} and GraphTransformer (GT) \citep{DBLP:conf/nips/KhoslaTWSTIMLK20}. Additionally, in Appendix \ref{ap: other pretrain}, we verify the effectiveness of our methods on GNNs pre-trained using GraphMAE \citep{DBLP:conf/kdd/HouLCDYW022} and MVGRL \citep{DBLP:conf/icml/HassaniA20}, two other representative GNN self-supervised learning algorithms. For each dataset, to pre-train GNNs, we leverage self-supervised learning methods on all the graphs without any label information. 

\textit{Baselines.} We compare our methods with the following baselines: (1) training a GNN from scratch supervised by few-shot data (\textit{``supervised''}); (2) fine-tuning a task head together with pre-trained GNN (\textit{``fine-tune''}). We allow GNNs to be tunable for ``\textit{supervised}'' and ``\textit{fine-tune}''; (3) state-of-the-art graph prompting algorithms: All-in-one (\textit{``AIO''}) \citep{DBLP:conf/kdd/SunCLLG23}, which is the only graph prompting algorithm that supports multiple tasks in node-level, edge-level and graph-level to the best of our knowledge; GPF-plus \citep{DBLP:conf/nips/FangZYWC23} which prompt on graph features and Gprompt \citep{DBLP:conf/www/LiuY0023} which is based on subgraph similarity.

\subsection{Few-shot Learning}
\label{sec: few-shot}
We investigate our improved graph prompt (``\textit{ImprovedAIO}'') and Multimodal prompt (``\textit{Morpher}'') to adapt frozen pre-trained GNNs using few-shot data. We focus on graph-level classification here and will further investigate the few-shot learning ability at other task levels in Section \ref{sec: multi-task}. Our few-shot learning setting is more challenging than existing works \citep{DBLP:conf/kdd/SunCLLG23, DBLP:conf/kdd/SunZHWW22} as we only allow no more than 10 labeled training and validation samples for each class. 
The results are shown in Table \ref{TB: main comparison}. By observations, given the same pre-trained GNN, our ImprovedAIO outperforms all the existing baseline methods. This improvement is attributed to its design, which restricts cross-connections, ensuring stable training and optimization. Moreover, Our Morpher can achieve an absolute further accuracy improvement over the baselines across all datasets. Its superior performance, even under extremely weak text supervision, stems from its ability to dynamically adapt and align the graph and language representation spaces with prompt learning. This flexibility enables Morpher to better leverage the semantic information from weakly-supervised text labels while preserving the structural integrity of the graph embeddings, resulting in more robust and accurate predictions.

\subsection{Morpher Supports Multiple-level Tasks}
\label{sec: multi-task}

\begin{table}[]
        \centering
        \vspace{-2mm}
        \resizebox{0.45\textwidth}{!}{%
        \begin{tabular}{@{}cccccc@{}}
        \toprule
        \multicolumn{2}{c}{\makecell[c]{Dataset}}    & \multicolumn{2}{c}{Cora}    & \multicolumn{2}{c}{CiteSeer}  \\ \midrule
        Tasks                        & Methods & Acc & F1 & Acc & F1  \\ \midrule
        \multirow{5}{*}{\makecell[c]{Node\\Level}}  
                                    & Supervised    &  52.83    & 47.73     & 63.91  & 64.82      \\
                                    & Fine-tune     & 56.37     & 55.04     & 64.87  & 66.42      \\
                                    & AIO \citep{DBLP:conf/kdd/SunCLLG23}     & 14.69         & 7.10     & 18.93  & 6.92         \\ 
                                    & ImprovedAIO   & \underline{58.46}     & \underline{55.10}     & \underline{66.44}  & \underline{66.53}       \\ 
                                    & Morpher       & \textbf{61.26}     & \textbf{62.36}     & \textbf{68.20}  & \textbf{68.56}        \\ 
        \midrule
        \multirow{5}{*}{\makecell[c]{Edge\\Level}} 
                                    & Supervised    & 51.78          & 50.62     & 52.14  & 50.81        \\
                                    & Fine-tune     & 52.50         & 51.00     & 52.50  & 51.12      \\
                                    & AIO \citep{DBLP:conf/kdd/SunCLLG23}     & 50.00     & 33.33    & 50.00  & 33.33         \\ 
                                    & ImprovedAIO   & \underline{54.64}     & \underline{54.57}     & \underline{53.92}  & \underline{53.55}      \\ 
                                    & Morpher       & \textbf{55.71}     & \textbf{55.05}     & \textbf{55.35}  & \textbf{55.05}        \\
        \midrule
        \end{tabular}%
        }
        \vspace{-2mm}
        \caption{Node-level, edge-level performance. Best results are bolded and second-best results are underlined.}\label{TB: multi-task}
\end{table}
Inherited from AIO, our ImprovedAIO and Morpher also support adaptation to downstream tasks at node-level and edge-level, because they can be reformulated into graph-level tasks.
After reformulating node classification and task classification as graph classifications by inducing ego-graphs, we use GraphCL+GCN to pre-train the GNN and report the performance in Table \ref{TB: multi-task}. The results are consistent with graph level, where ImprovedAIO and Morpher outperform existing methods, with Morpher achieving slightly better performance than ImprovedAIO. Notably, as analyzed in Section \ref{sec: graph prompt}, AIO is unstable during training, for example in Table \ref{TB: multi-task} (Node level) and in certain cases of Table \ref{TB: main comparison} (e.g., on MSRC\_21C with GraphCL or with SimGRACE+GAT). The performance fluctuations observed are consistent and reflect the same underlying issue of AIO. In contrast, our proposed ImprovedAIO effectively addresses this issue.

\begin{figure*}[t!] 
{
\setlength{\tabcolsep}{1pt}
\begin{tabular}{c:c:c}
\begin{subfigure}{0.56\textwidth}
\includegraphics[width=\linewidth]{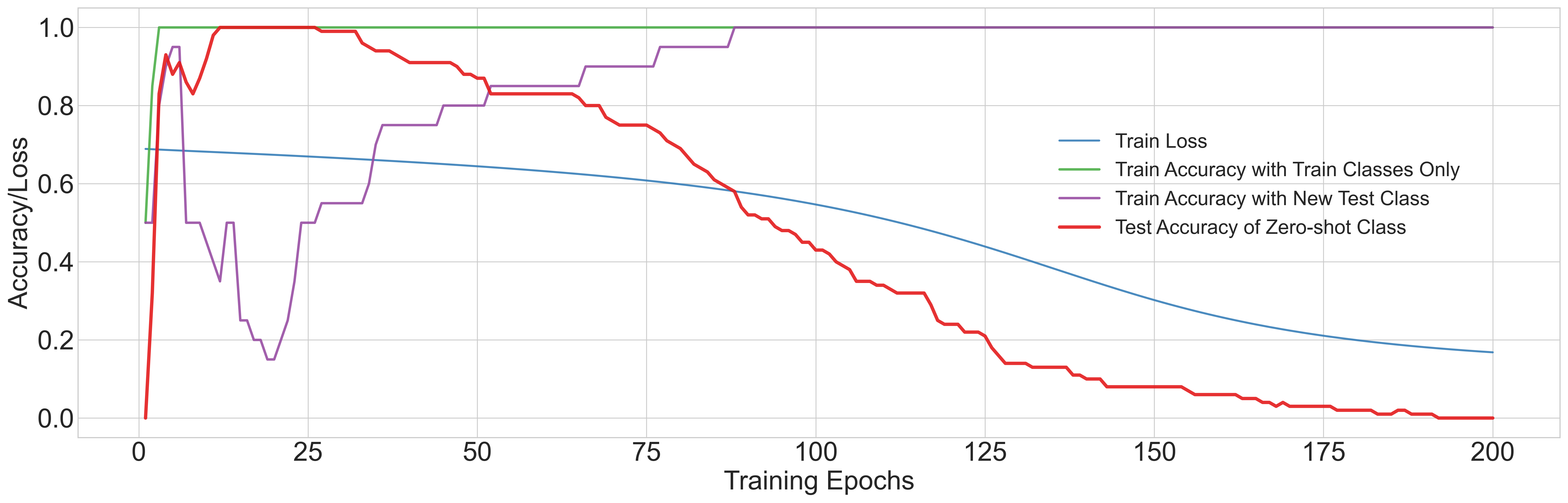}
\end{subfigure}\hspace*{\fill}
&
\begin{subfigure}{0.215\textwidth}
\includegraphics[width=\linewidth]{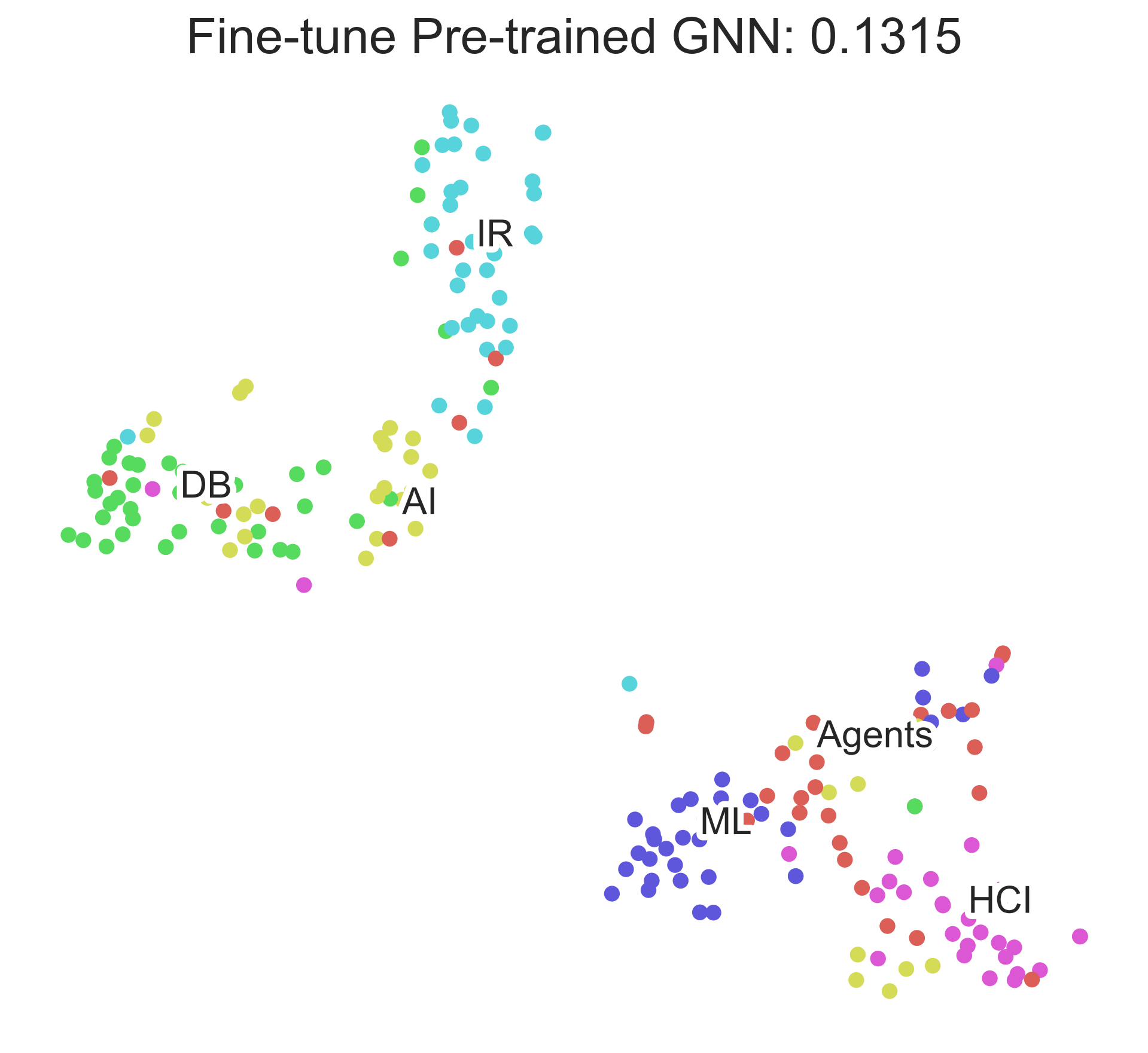}
\end{subfigure} 
&
\begin{subfigure}{0.215\textwidth}
\includegraphics[width=\linewidth]{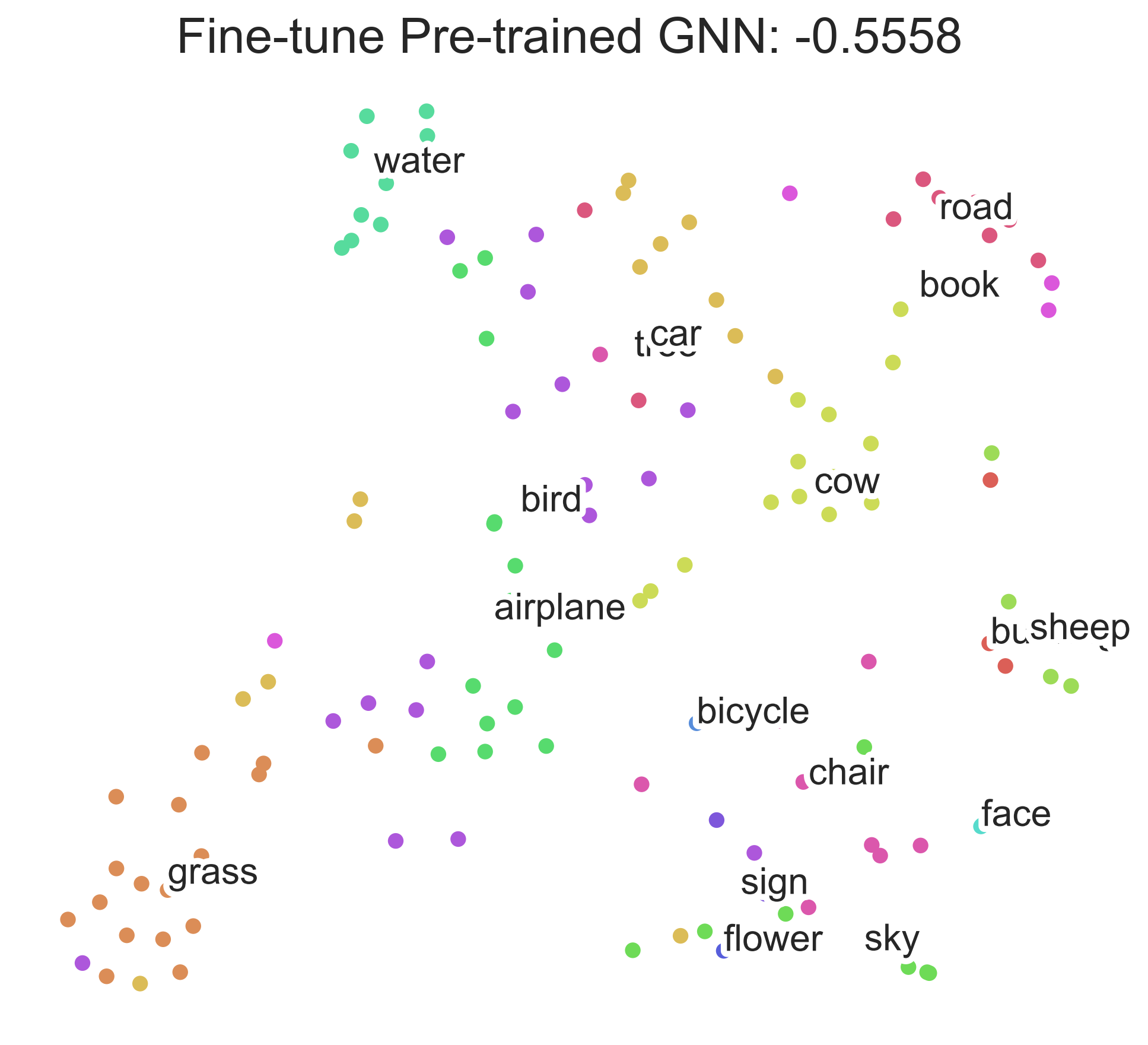}
\end{subfigure}
\\
\begin{subfigure}{0.56\textwidth}
\includegraphics[width=\linewidth]{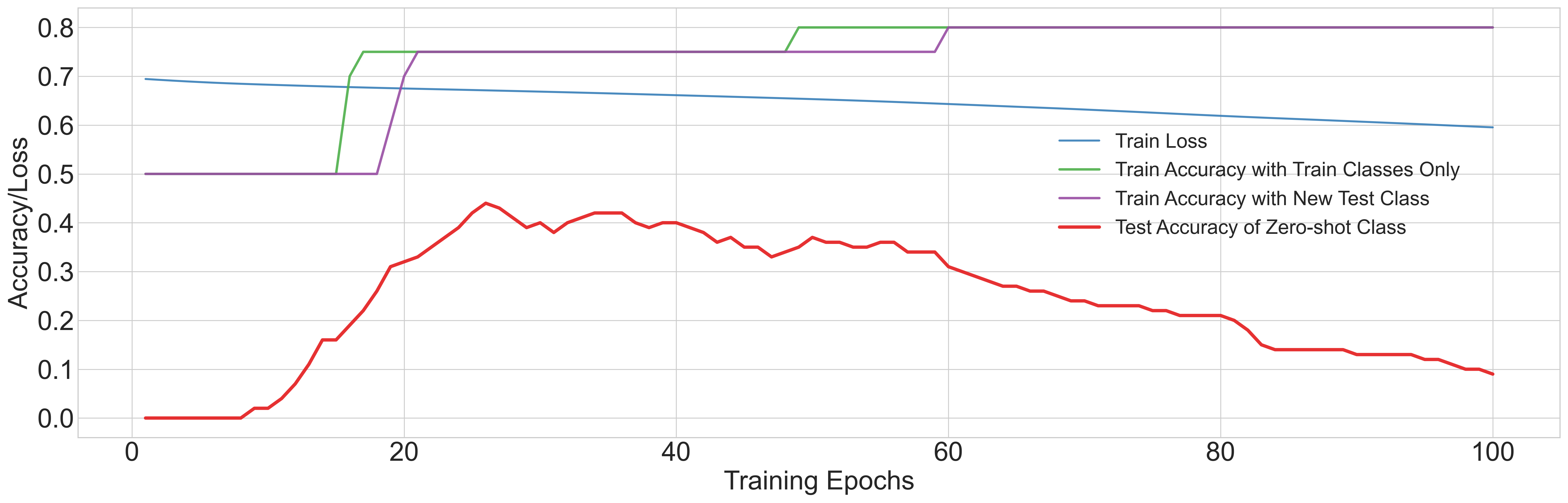}
\end{subfigure}\hspace*{\fill}
&
\begin{subfigure}{0.215\textwidth}
\includegraphics[width=\linewidth]{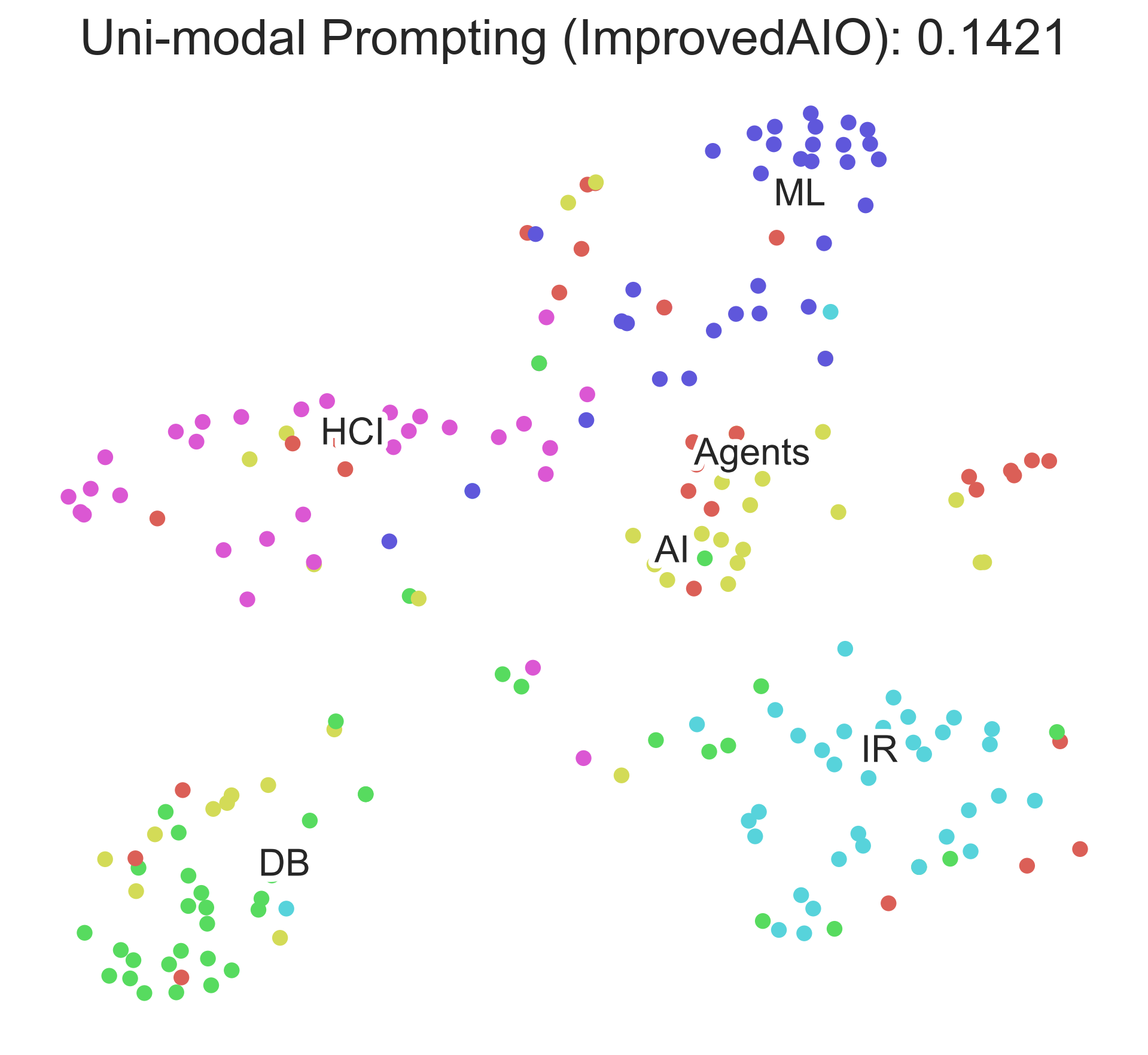}
\end{subfigure}
&
\begin{subfigure}{0.215\textwidth}
\includegraphics[width=\linewidth]{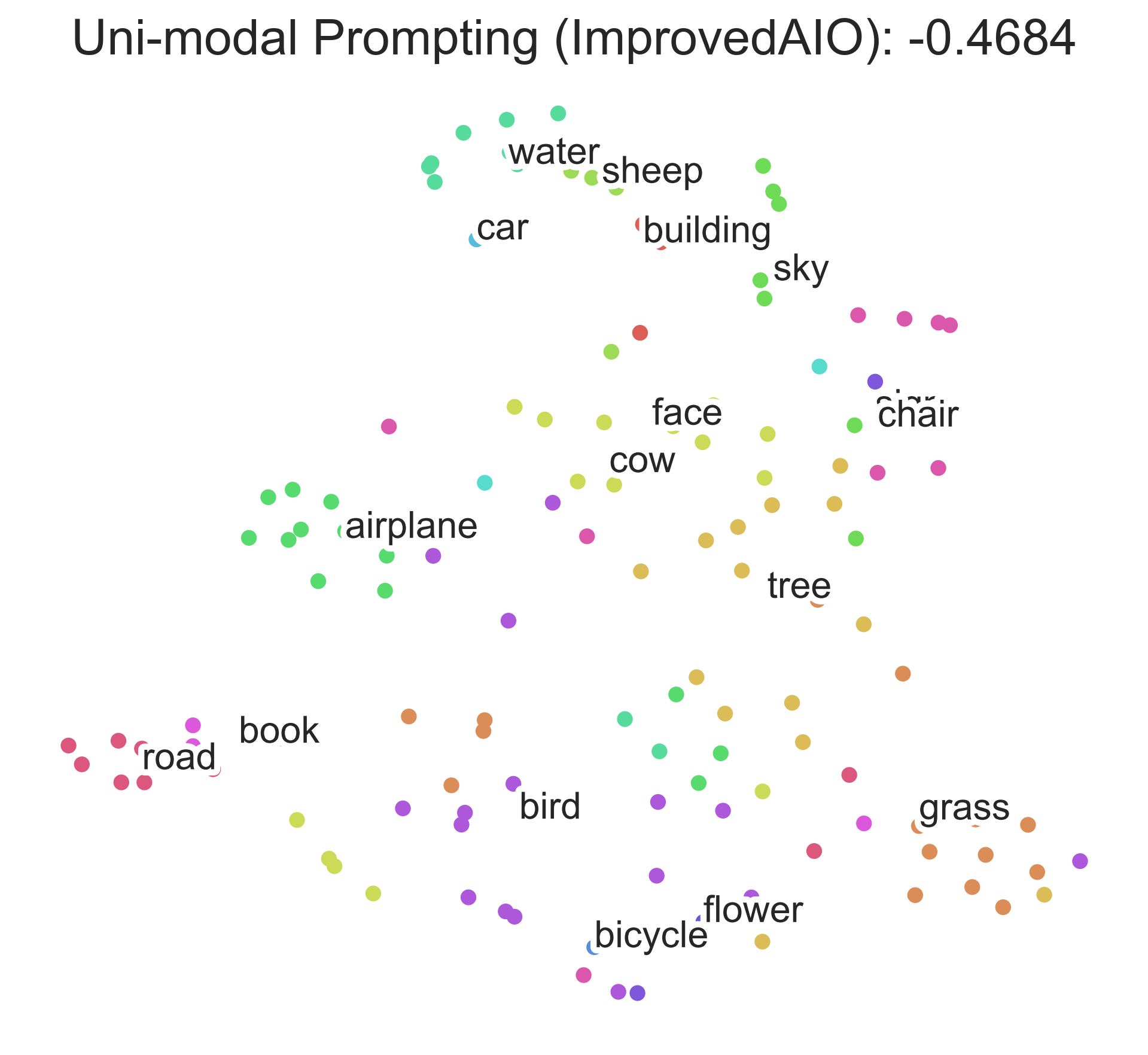}
\end{subfigure}
\\
\begin{subfigure}{0.56\textwidth}
\includegraphics[width=\linewidth]{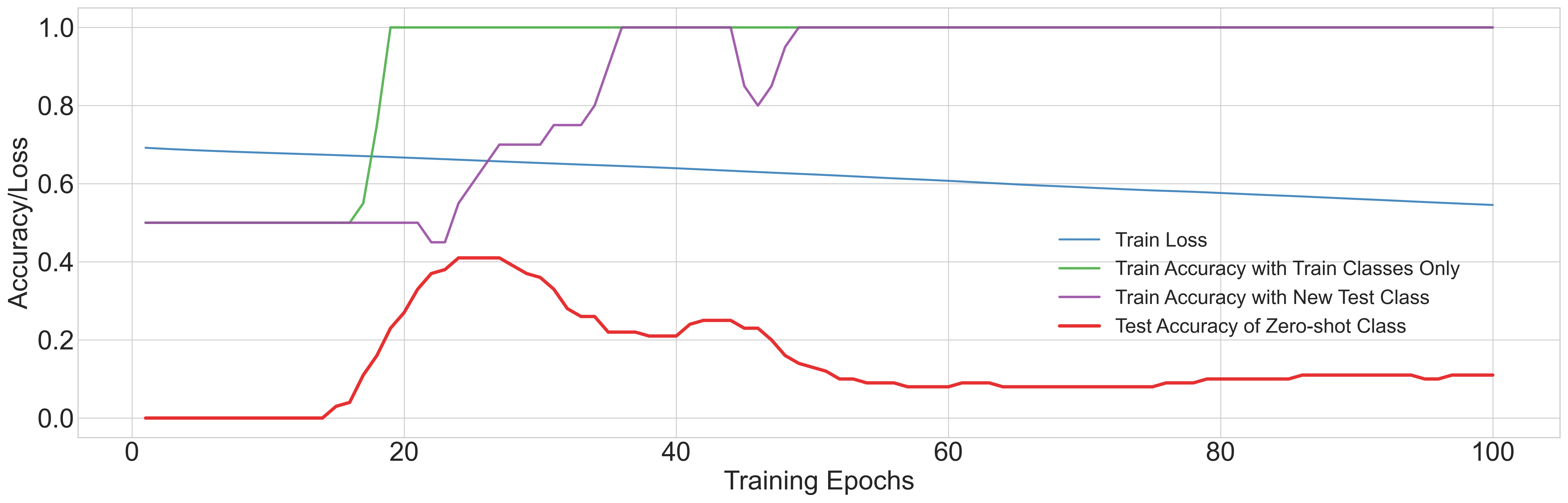}
\end{subfigure}\hspace*{\fill}
&
\begin{subfigure}{0.215\textwidth}
\includegraphics[width=\linewidth]{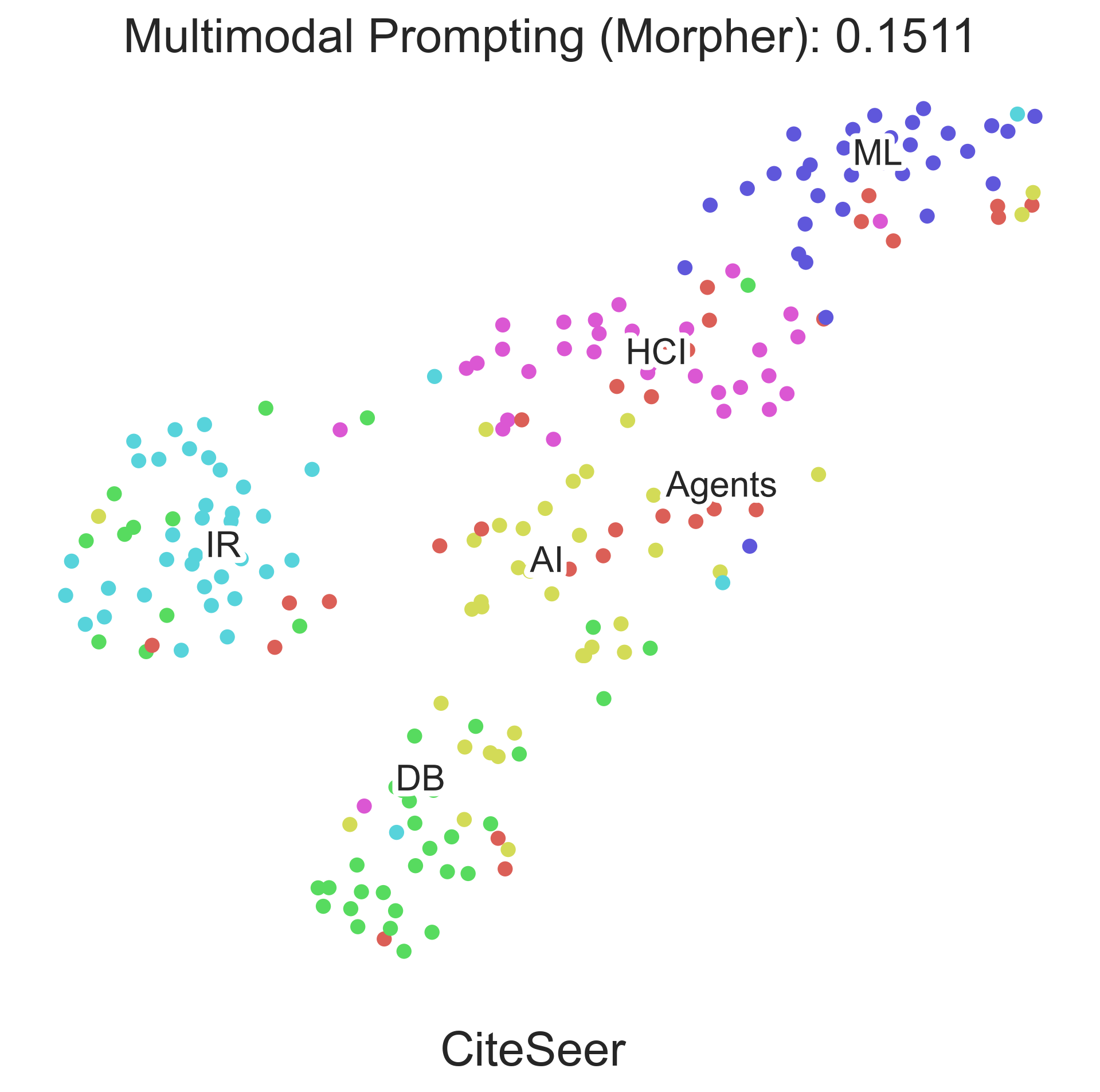}
\end{subfigure}
&
\begin{subfigure}{0.215\textwidth}
\includegraphics[width=\linewidth]{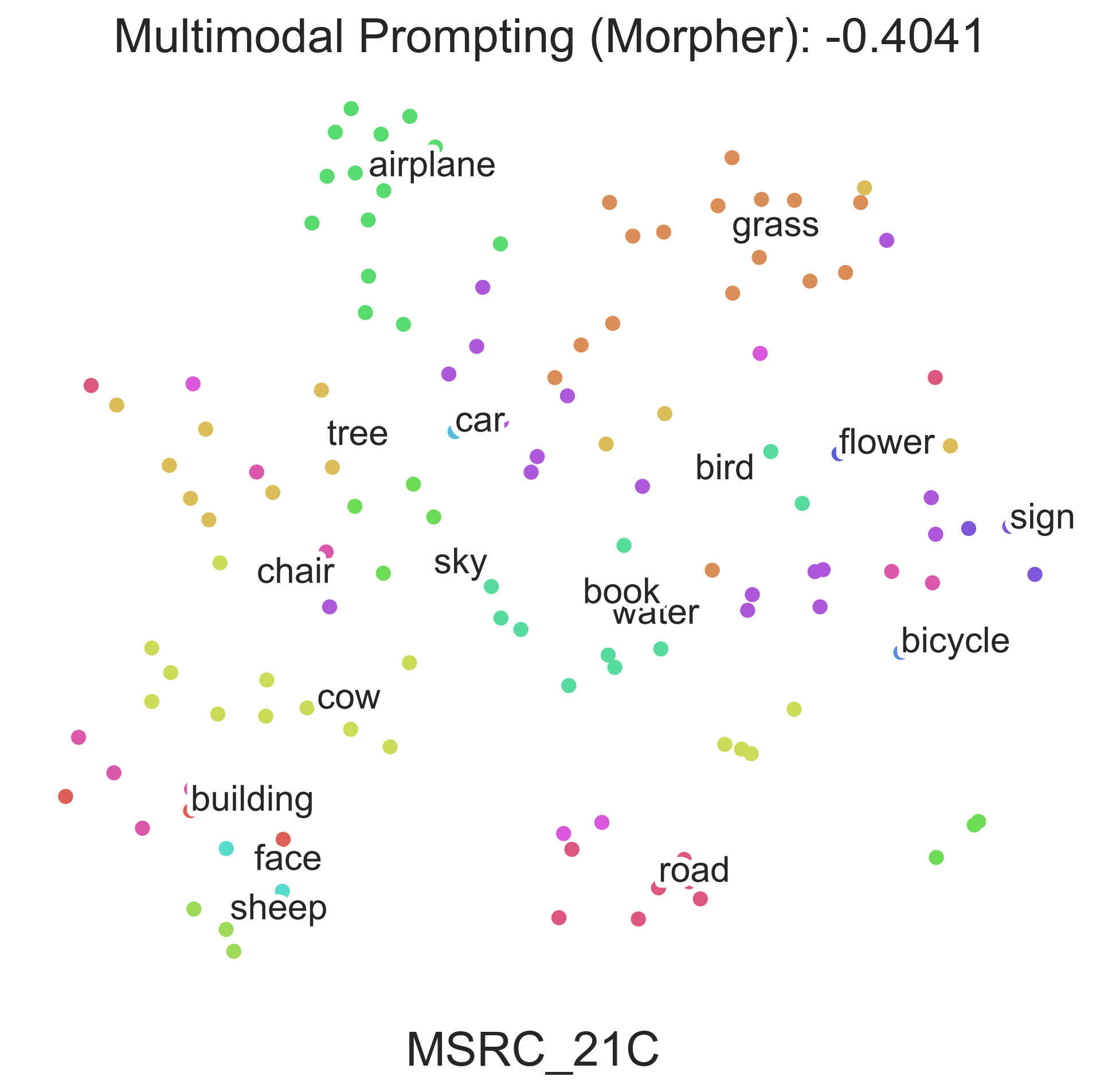}
\end{subfigure}

\end{tabular}
}

\caption{Results of novel class generalization (left); t-SNE embedding plots on CiteSeer, MSRC\_21C (right). Train accuracy with train classes only is the accuracy of predicting the training graphs from the two training classes. Train accuracy with new test classes is the accuracy of predicting the training graphs from all three classes. Test Accuracy of zero-shot class is the accuracy of predicting the testing graphs from all three classes. Full-resolution figures can be found in Appendix \ref{ap: full resolution figures}.} 
\label{fig: zero and emb}
\end{figure*}

\subsection{Domain Transfer}
We explore the potential of using Morpher for domain adaptation.
From the previous experiments, we have demonstrated that our ImprovedAIO outperforms the original AIO. Therefore, in the subsequent pages, we focus on comparing Morpher with ImprovedAIO to avoid redundancy. We pre-train GNNs on ENZYMES or CiteSeer datasets, then test the classification performance on MUTAG and PubMed and report the results in Table \ref{TB: transfer}. We unify the pre-train feature dimension with the downstream feature dimension by padding zeros or SVD reduction. From the results, Morpher demonstrates the best transferability, followed by ImprovedAIO. 
\begin{table}[h]
        \centering
        \resizebox{0.42\textwidth}{!}{%
        \begin{tabular}{@{}cccccccc@{}}
        \toprule
        \multicolumn{2}{c}{\makecell[c]{Target Domain}}    & \multicolumn{2}{c}{MUTAG}    & \multicolumn{2}{c}{PubMed}    \\ \midrule
        \multicolumn{2}{c}{\makecell[c]{Target Task}}    & \multicolumn{2}{c}{graph-level}    & \multicolumn{2}{c}{node-level}    \\ \midrule
        Source                        & Methods & Acc & F1 & Acc & F1  \\ \midrule
        \multirow{4}{*}{\makecell[c]{ENZYMES \\ (graph-level)}} 
                                     & Fine-tune      & 68.00         & 55.04     & 47.57  & 36.07      \\
                                     & ImprovedAIO     & \underline{70.67}         & \underline{64.07}     & \underline{50.28}  & \underline{50.51}      \\ 
                                     & Morpher     & \textbf{72.67}         & \textbf{73.29}     & \textbf{54.42}  & \textbf{53.96}         \\ 
        \midrule
        \multirow{4}{*}{\makecell[c]{CiteSeer \\ (node-level)}} 
                                     & Fine-tune      & 71.33         & 62.19     & 48.71  & 40.66         \\
                                     & ImprovedAIO     & \underline{74.00}         & \underline{73.76}     & \underline{52.57}  & \underline{51.29}       \\ 
                                     & Morpher     & \textbf{76.67}         & \textbf{77.04}     & \textbf{58.29}  & \textbf{57.54}         \\
        \midrule
        \end{tabular}%
        }
        \vspace{-2mm}
        \caption{Domain Transfer Performance. Best results are bolded and second-best results are underlined.}\label{TB: transfer}
\end{table}

\subsection{Zero-shot Classification Prototype}
\label{sec: zero-shot}
An advantage of adapting pre-trained GNNs to the semantic embedding space is that GNNs might be empowered to ``reasoning'' in a CLIP style. Here, we conduct a novel experiment that generalizes GNN to an unseen class. Since no real-world data is available for this setting, we synthetically create three datasets, ZERO-Cora, ZERO-CiteSeer, and ZERO-PubMed, all from real-world connections.
We aim to simulate a citation network with two research areas and an interdisciplinary research area in between.  
For each citation network, we randomly sample 120 nodes and induce their 2-hop ego-graphs and then replace the node features in 10 ego-graphs with $[1, 0]$ and another 10 ego-graphs with $[0, 1]$ to construct 20 training graph samples. 
For the remaining ego-graphs, we uniformly randomly replace the node features with $[1, 0]$ and $[0, 1]$ to construct 100 testing graph samples. We assign text labels of the first research area (e.g., ``biology'') to the $[1, 0]$ training graphs, the second research area (e.g., ``informatics'') to the $[0, 1]$ training graphs, and the interdisciplinary area (e.g., ``bioinformatics'') to the testing graphs. Intuitively, the nodes with feature $[1, 0]$ are papers in the first area, and nodes with feature $[0, 1]$ are in the second area, which makes the datasets rational. 

\begin{figure*}[t]
\begin{subfigure}{0.32\textwidth}
\includegraphics[width=\linewidth]{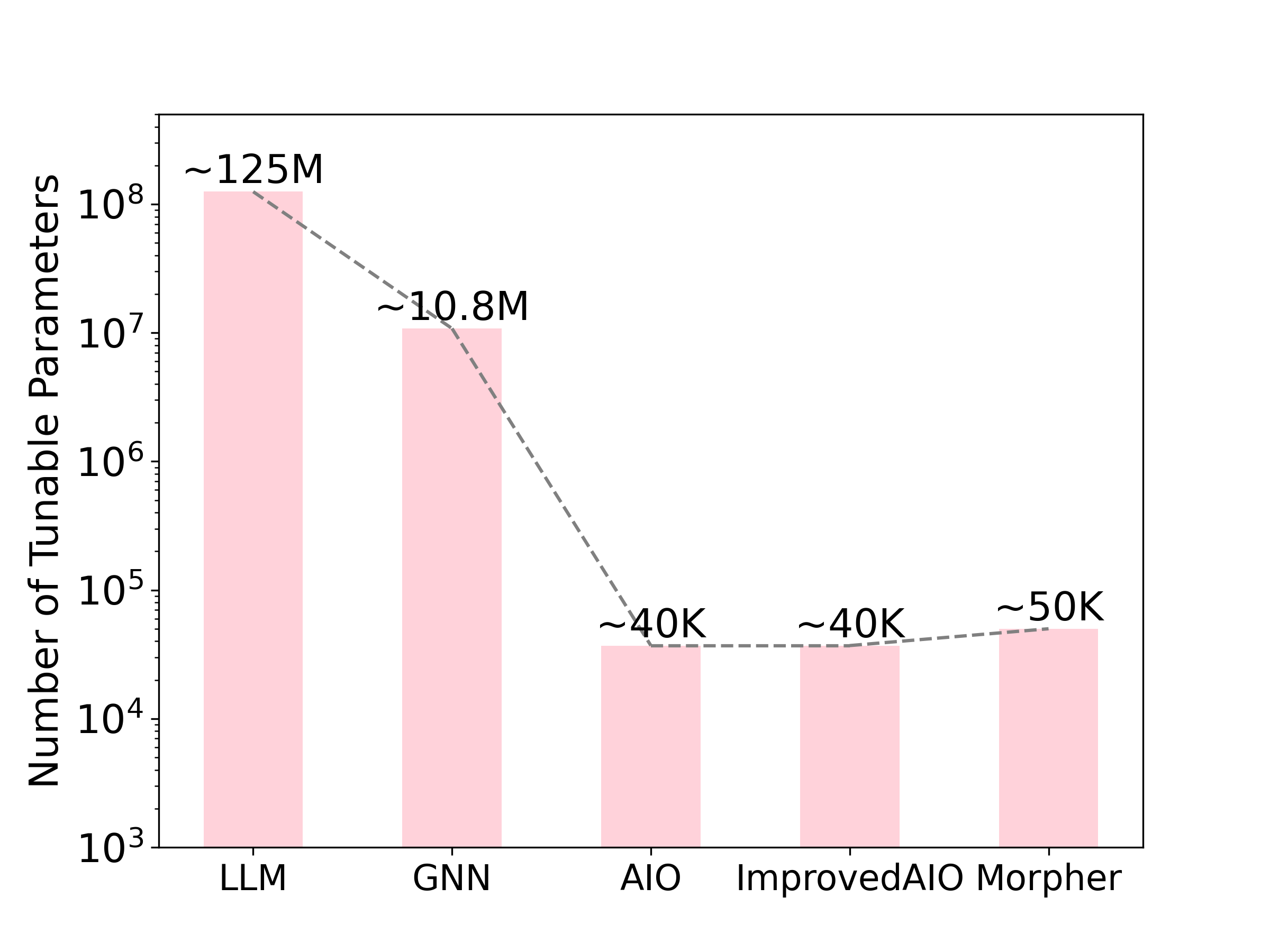}
\end{subfigure}\hspace*{\fill}
\begin{subfigure}{0.32\textwidth}
\includegraphics[width=\linewidth]{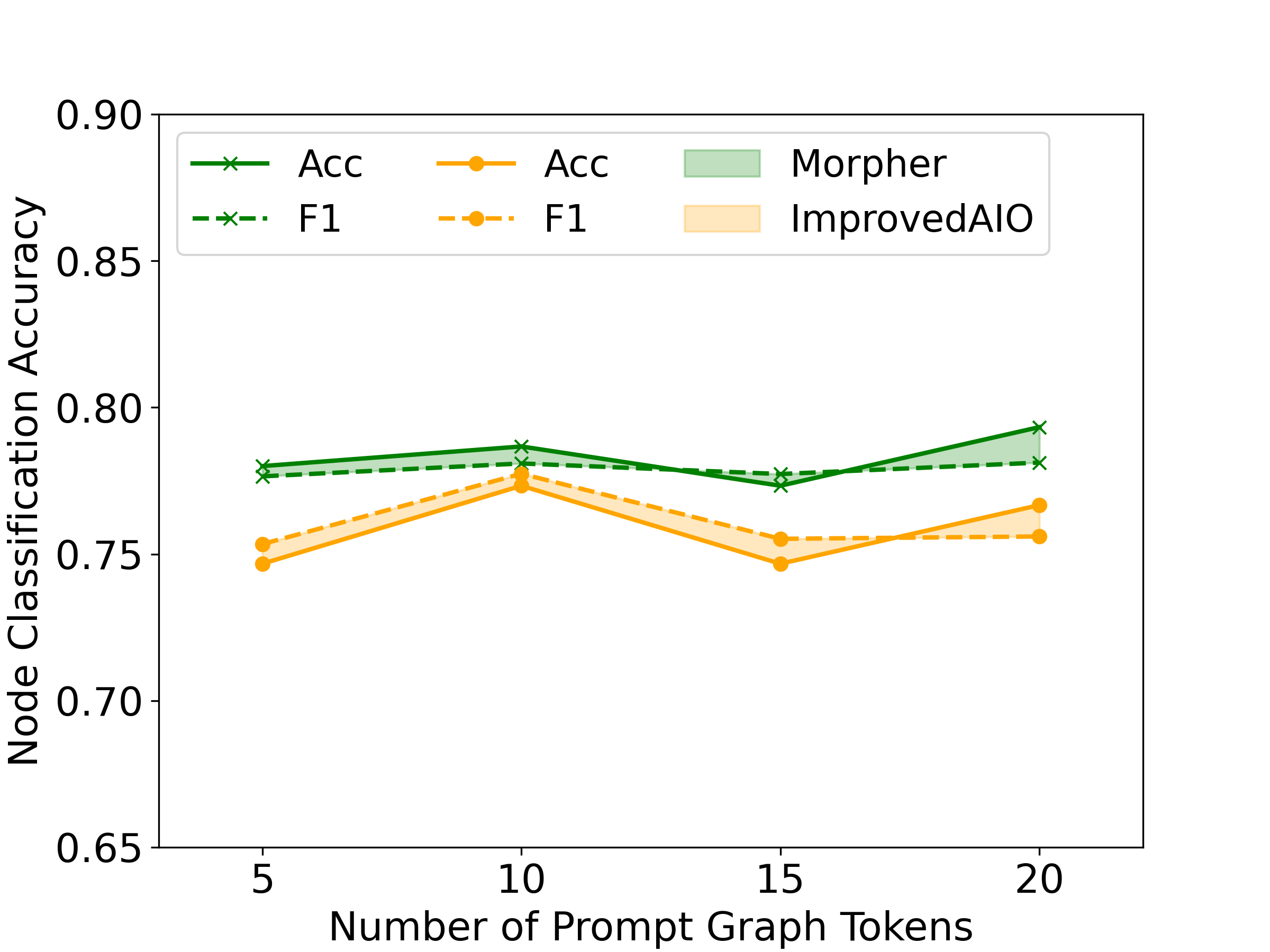}
\end{subfigure}\hspace*{\fill}
\begin{subfigure}{0.32\textwidth}
\includegraphics[width=\linewidth]{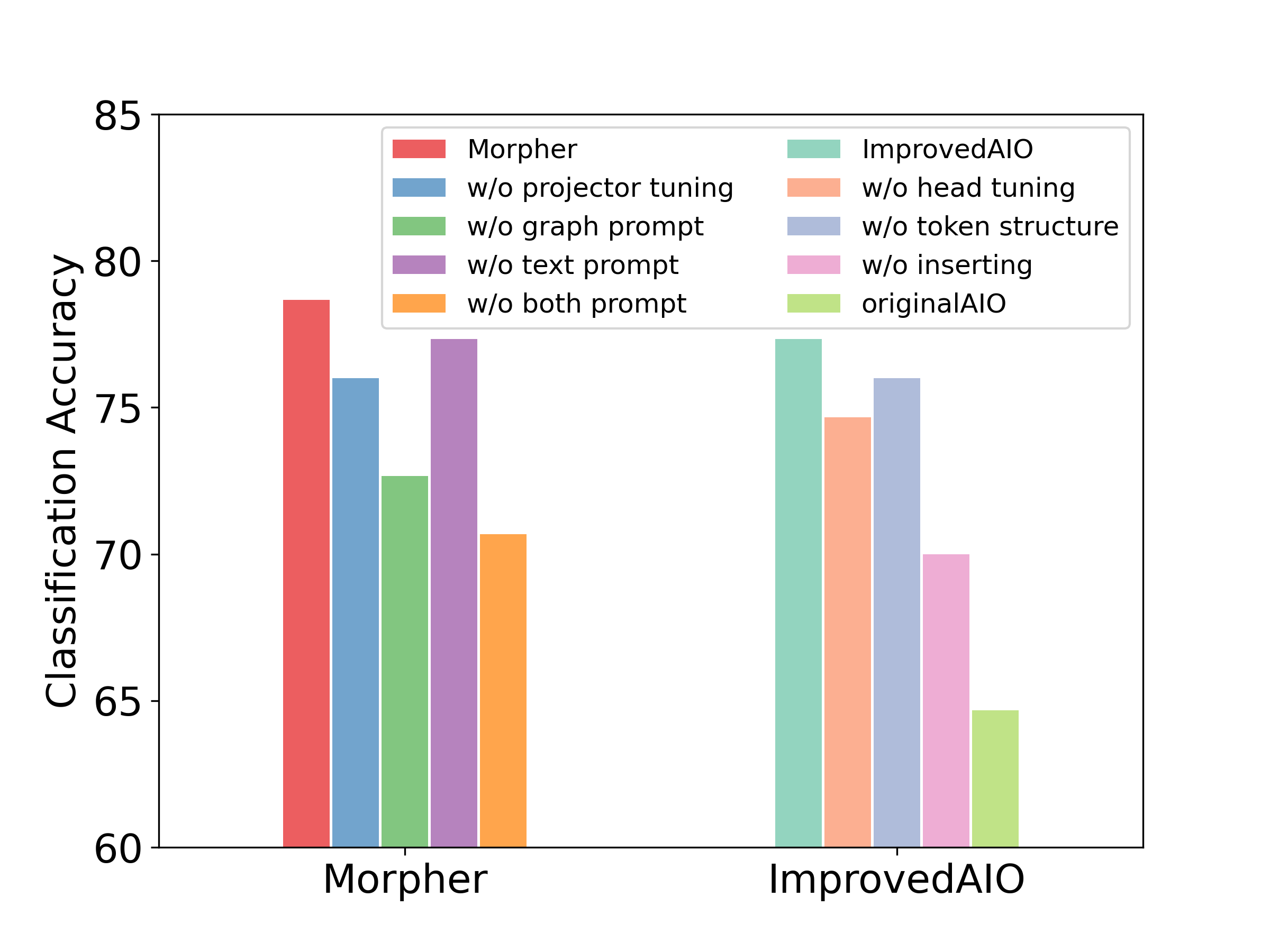}
\end{subfigure}
\vspace{-2mm}
\caption{Efficiency comparison (left), parameter study (middle) and ablation study (right).}
\label{fig: efficiency parameter and ablation}
\end{figure*}

For each dataset, using GraphCL+GCN, we pre-train GNNs on all graphs. Then, we train Morpher on the training graphs, only knowing the text labels of the two training classes. Since we do not have validation data in zero-shot learning, we report the results
of each epoch in Figure \ref{fig: zero and emb} (left). We observe that, while Morpher quickly adapts the GNN to downstream training data, the CLIP-like framework can predict the graphs in the novel class with good accuracy (red curve). Moreover, the training samples can be classified correctly from training and novel classes. Before the training overfits, there is a period when Morpher can distinguish all the graphs from the training and novel classes with high accuracy. 
Such zero-shot novel-class generalization ability validates Morpher's alignment between graph embeddings and text embeddings. When Morpher is trained on two classes of graphs with text labels of biology and informatics, a graph-in-the-middle will be classified as text-in-the-middle: bioinformatics, even if “bioinformatics” is an unseen label.

\subsection{Efficiency and Embedding Analysis}
Without fine-tuning the GNN or LLM, the prompt-based methods have better parameter efficiency. As shown in Figure \ref{fig: efficiency parameter and ablation} (left), our ImprovedAIO and Morpher require similar numbers of parameters with AIO \citep{DBLP:conf/kdd/SunCLLG23}, which is $0.032\%$ to $0.46\%$ compared to either tune the LLM (RoBERTa) or GNN (GCN). Due to such parameter efficiency, our methods learn better graph representations given few-shot data. We visualize the graph embeddings of CiteSeer and MSRC\_21C in Figure \ref{fig: zero and emb} and calculate the silhouette score, a metric for cluster quality ($\uparrow$) ranged in $[-1, 1]$. It turns out that our Morpher leads to better adaptation.

\subsection{Hyperparameter and Ablation Study}
\label{subsec: hyper and ablation}
We conduct the hyperparameter study by choosing and testing various numbers of graph prompt tokens for both ImprovedAIO and Morpher. The results are shown in Figure \ref{fig: efficiency parameter and ablation} (middle), from which we can observe that both methods are generally stable, and Morpher constantly outperforms ImprovedAIO under different choices. 
To verify the necessity of each component in our design, we compare Morpher and ImprovedAIO with multiple variants, respectively, and report the result in Figure \ref{fig: efficiency parameter and ablation} (right). We observe that removing any component would result in a performance drop. 

We also conduct experiments to verify the effectiveness of our proposed Morpher with ELECTRA \citep{DBLP:conf/iclr/ClarkLLM20} and DistilBERT \citep{DBLP:journals/corr/abs-1910-01108} as the text encoder in Appendix \ref{ap: other llm} due to space limitation. In general, Morpher is robust with respect to the language encoder. 
As for the robustness with respect to the pre-trained GNNs, we further conduct experiments using GNNs pre-trained from GraphMAE \citep{DBLP:conf/kdd/HouLCDYW022} and MVGRL \citep{DBLP:conf/icml/HassaniA20}. Due to the space limitation, the results are in Appendix \ref{ap: other pretrain}.

%% file: tables/few_shot_learning.tex
\begin{table*}
\centering
\resizebox{0.7\textwidth}{!}{%
\begin{tabular}{
cccccccccc
}
\toprule
\midrule
\multirow{2}{*}{\makecell[c]{Training\\ schemes}}     & \multirow{2}{*}{GNN pretraining} & \multicolumn{2}{c}{MUTAG}       & \multicolumn{2}{c}{ENZYMES}    & \multicolumn{2}{c}{PROTEINS}     & \multicolumn{2}{c}{MSRC\_21C}  \\
                                      &                          & Acc & F1 & Acc & F1 & Acc & F1 & Acc & F1\\ \midrule
\multirow{3}{*}{Supervised}           
& N/A + GCN      & 66.00    & 66.67    & 16.67    & 8.68    & 65.89    & 60.77      & 38.85    & 35.32 \\
& N/A + GAT      & 66.00    & 65.69    & 16.45    & 4.65    & 64.75    & 64.08    & 41.14    & 39.86 \\
& N/A + GT       & 66.66    & 66.26    & 15.62    & 4.22    & 62.81    & 57.12    & 38.28    & 41.62 \\
\midrule
\multirow{6}{*}{\makecell[c]{Pre-train \\+\\  Fine-tune}} 
& GraphCL+GCN   & 70.00    & 70.23    & 17.91     & 11.82    & 65.89    & 61.23     & 40.00     & 43.89 \\
& GraphCL+GAT   & 70.00    & 69.73    & 17.91     & 10.46    & 65.16    & 63.92     & 44.57     & 45.74 \\
& GraphCL+GT    & 68.00    & 67.81    & 17.70     & 8.99    & 63.28    & 56.41     & 41.71    & 43.73 \\
& SimGRACE+GCN  & 66.67    & 67.27    & 17.29     & 8.78    & 66.82    & 64.70 & 40.57    & 43.84 \\
& SimGRACE+GAT  & 70.67    & 69.10    & 16.87     & 7.18    & 65.42    & 63.65     & 42.85    & 42.37 \\
& SimGRACE+GT   & 69.33    & 69.77    & 16.24     & 6.08    & 65.98    & 62.31     & 39.42    & 40.78 \\
\midrule
\multirow{6}{*}{\makecell{AIO\\ \citep{DBLP:conf/kdd/SunCLLG23}}}               
& GraphCL+GCN   & 64.67 & 39.27 & 17.50 & 4.97  & 61.35 & 44.93 & 3.59 & 10.09\\
& GraphCL+GAT   & 64.67 & 39.27 & 17.50 & 4.97  & 59.21 & 37.19 & 14.37 & 3.11\\
& GraphCL+GT    & 73.33 & 72.06 & 18.33 & 9.09 & 40.79 & 28.97 & 17.96 & 8.30\\
& SimGRACE+GCN  & 64.67 & 39.27 & 16.04 & 4.61  & 67.42 & 60.87 & 34.73 & 18.16\\
& SimGRACE+GAT  & 64.67 & 39.27 & 16.04 & 4.61  & 59.21 & 37.19 & 7.78 & 1.79\\
& SimGRACE+GT   & 36.00 & 27.26 & 17.50 & 8.15 & 50.56 & 49.34 & 32.34 & 15.13\\
\midrule
\multirow{6}{*}{\makecell[c]{ImprovedAIO\\(Ours)}}               
& GraphCL+GCN   & 77.33    & 77.74    & 18.13     & 11.98    & 65.89    & 65.97     & 42.85     & 45.91 \\
& GraphCL+GAT   & 74.67    & 75.51    & 18.33     & 11.26    & 65.76    & 66.05     & \underline{46.85}     & \underline{51.39} \\
& GraphCL+GT    & 74.67    & 74.67    & 19.16     & 9.04    & 68.12    & 68.18     & 42.85     & 43.54 \\
& SimGRACE+GCN  & 68.00    & 69.01    & 17.91     & 9.02    & 66.82    & 66.40     & 44.57     & 49.24 \\
& SimGRACE+GAT  & 77.33    & 77.20    & 18.75     & 9.39    & 66.91    & 65.49     & 45.14     & 42.31 \\
& SimGRACE+GT   & 71.33    & 72.06    & 18.95     & 11.25    & 68.59    & 68.84     & 40.57     & 42.82\\
\midrule
\multirow{6}{*}{\makecell[c]{Morpher\\(Ours)}}               
& GraphCL+GCN   & \underline{78.67}    & \underline{78.09}    & \underline{20.41}     & 15.20    & 67.47    & 66.40     & 45.14     & 49.62 \\
& GraphCL+GAT   & \textbf{79.33}    & \textbf{79.15}    & \textbf{23.12}     & \textbf{18.01}    & 70.89    & 70.30     & \textbf{50.85}     & \textbf{54.48} \\
& GraphCL+GT    & 76.00    & 76.51    & 19.58     & 13.28    & \textbf{73.53}    & \textbf{72.48}     & 45.71    & 48.41 \\
& SimGRACE+GCN  & 69.33    & 70.27    & 19.79     & 14.94    & 67.10    & 66.15     & 45.71    & 51.24 \\
& SimGRACE+GAT  & 78.00    & 77.65    & 20.21     & \underline{16.27}    & 68.12    & 67.26     & 45.71   & 51.13 \\
& SimGRACE+GT   & 74.00    & 74.84    & 19.16     & 14.29    &\underline{71.76}    & \underline{71.75}     & 44.00    & 48.16 \\
\midrule
\rowcolor{gray!10} \multicolumn{2}{c}{\textit{IMP of ImprovedAIO (\%)}}   & 3.89 $\uparrow$ &	4.67 $\uparrow$ &	0.90 $\uparrow$ &	1.07 $\uparrow$  &	1.41 $\uparrow$ &	4.64 $\uparrow$ &	2.29 $\uparrow$ &	2.34  $\uparrow$ \\     
\midrule
\rowcolor{gray!15} \multicolumn{2}{c}{\textit{Further IMP of Morpher (\%)}}     &                    2.00 $\uparrow$ & 	1.72 $\uparrow$  & 	1.84 $\uparrow$ & 	5.01 $\uparrow$  & 	2.8 $\uparrow$ & 	2.24 $\uparrow$ & 	2.38 $\uparrow$ & 	4.64  $\uparrow$\\  
\bottomrule
\end{tabular}%
}
\caption{Few-shot graph classification performance (\%). \textit{IMP (\%)}: the average improvement (absolute value) compared to the \textbf{best result} among all the baseline methods. Best results are bolded and second-best results are underlined. We also compared with GPF-plus \citep{DBLP:conf/nips/FangZYWC23} and Gprompt \citep{DBLP:conf/www/LiuY0023}, in Appendix Table \ref{TB: main comparison full} due to space limit.}
\vspace{-4mm}
\label{TB: main comparison}
\end{table*}

%% file: 0_sections/related_work.tex
\section{Related Work}
\paragraph{GNN Pre-training.} 
Recently, a surge of graph pre-training strategies have emerged~\citep{DBLP:conf/iclr/HuLGZLPL20, DBLP:conf/aaai/LuJ0S21, jing2021hdmi, DBLP:conf/cikm/ZhouZF0H22}. The main idea of pre-trained graph models is to capture general graph information across different tasks and transfer this knowledge to the target task using techniques such as contrastive predictive coding~\citep{DBLP:conf/nips/KhoslaTWSTIMLK20, DBLP:conf/www/XiaWCHL22}, context prediction~\citep{DBLP:conf/kdd/HuZiniu20}, and mutual information maximization~\citep{DBLP:conf/iclr/Infograph}. Different from these approaches, this paper aims to adapt pre-trained GNNs by leveraging multi-modal prompt learning techniques.

\paragraph{Graph Prompt Learning.}
Recent studies exploring prompt learning for GNNs mark a thriving research area \citep{DBLP:journals/corr/abs-2311-16534, DBLP:journals/corr/abs-2303-07275}. It is a promising way to adapt GNNs to downstream tasks through token-level \citep{DBLP:conf/nips/FangZYWC23, DBLP:conf/kdd/TanGDL23, DBLP:journals/corr/abs-2310-14845, DBLP:conf/kdd/SunZHWW22, DBLP:journals/corr/abs-2302-12449} or graph-level \citep{DBLP:conf/kdd/SunCLLG23, DBLP:conf/nips/HuangRCK0LL23, DBLP:journals/corr/abs-2310-17394} prompting. Among all the existing methods, All-in-one (AIO) \citep{DBLP:conf/kdd/SunCLLG23} is the only algorithm to learn tunable graph prompts for multi-level downstream tasks (Table \ref{tb: graph prompts comparison}). Based on our improved AIO, we present a pioneer study to explore learning prompts in multiple modalities simultaneously while keeping the pre-trained models frozen.

\paragraph{LLM on Graphs.} 
LLMs' potential for graph-related tasks \citep{jin2023large} has been explored recently.
The first category employs LLMs as pre-trained feature extractors to enhance GNNs \citep{duan2023simteg,chien2021node,zhu2021textgnn}.
The second category focuses on integrating graph structures directly into LLM architectures \citep{yang2021graphformers, zhang2022greaselm, jin2023heterformer, jin2023patton}.
Despite these advancements, none of them have explored the collaboration between LLMs and GNNs with extremely weak text supervision and under graph prompt learning.

%% file: 0_sections/z_conclusion_discussion.tex
\section{Conclusion}
We present Morpher, the first multimodal prompt learning paradigm leveraging LLMs to semantically adapt pre-trained GNNs to downstream tasks with extremely weak text supervision. Addressing limitations of existing graph prompting techniques, we demonstrate through extensive experiments that Morpher excels in few-shot, multi-level tasks, and domain transfer. Notably, Morpher enables generalization to novel testing classes.

\section*{Acknowledgments}
Research was supported in part by US DARPA INCAS Program No. HR0011-21-C0165 and BRIES Program No. HR0011-24-3-0325, National Science Foundation Award No. IIS-19-56151 and Award No. IIS-2117902, the Molecule Maker Lab Institute: An AI Research Institutes program supported by NSF under Award No. 2019897, the Institute for Geospatial Understanding through an Integrative Discovery Environment (I-GUIDE) by NSF under Award No. 2118329, and IBM-Illinois Discovery Accelerator Institute - a new model of an academic-industry partnership designed to increase access to technology education and skill development to spur breakthroughs in emerging areas of technology. The views and conclusions are those of the authors and should not be interpreted as representing the official policies of the funding agencies or the government.

\section{Limitations}
Our Multimodal graph prompt learning paradigm assumes the ``pre-train + prompt'' framework to learn transferable graph representations, yet there could be other paths to achieve graph-related foundation models. Also, graph prompt learning only works on the graph neural network architecture, and might not work for other architectures that are proposed in the future. Another limitation of this work is the requirement of language encoder. While RoBERTa is one of the most advanced encoder-only language models and can be considered an LLM with over 0.1B parameters, more recent LLMs such as Llama or Mistral cannot be used in Morpher because they are decoder-only LLMs and do not explicitly have an encoder.

%% file: appendix.tex
\section{Dataset Details}
\label{ap: dataset}

\begin{table*}[t]
\centering
\caption{Dataset statistics}
\label{tab: data}
\resizebox{0.99\textwidth}{!}{%
\begin{tabular}{@{}ccccccccc@{}}
\toprule
Dataset & task level & \# graphs & average \# nodes & average \# edges  & \# feature dimension & \# classes & \# shots per class & feature characteristic         \\ \midrule
MUTAG  & graph  & 188 & 17.9    & 39.6     & 7       & 2     & 10     & one-hot, sparse   \\
ENZYMES & graph & 600 & 32.6   & 124.3    & 3        & 6   & 10  & one-hot, sparse \\
PROTEINS  & graph & 1113 & 39.1  & 145.6 & 3        & 2    & 10  & one-hot, sparse     \\
MSRC\_21C  & graph & 209 & 40.28  & 96.60 & 22        & 17    & 1   & one-hot, sparse    \\
Cora     & node, edge & 1    & 2708  & 10556 & 1433  & 7 & 2 (node), 20 (edge)  & sum 1, sparse  \\
CiteSeer     & node, edge & 1    & 3327  & 9104 & 3703  & 6 & 2 (node), 20 (edge)  & sum 1, sparse   \\
PubMed  & node & 1 & 19,717 & 88648 & 500& 3 & 10 & TF-IDF value, dense \\
\bottomrule
\end{tabular}%
}
\end{table*}

\begin{table*}[t]
\centering
\caption{Synthetic Zero-shot Class Generalization Dataset statistics}
\label{tab: zero data}
\resizebox{0.98\textwidth}{!}{%
\begin{tabular}{@{}ccccccc@{}}
\toprule
Dataset & \# graphs & average \# nodes & average \# edges  & \#feature dimension & \# classes & \# shots per class          \\ \midrule
ZERO-Cora    & 120 & 8.41    & 10.38     & 2       & 2     & 10       \\
ZERO-CiteSeer & 120 & 10.03   & 21.31    & 2        & 2   & 10   \\
ZERO-PubMed  & 120 & 20.33  & 41.75 & 2        & 2    & 10       \\
\bottomrule
\end{tabular}%
}
\end{table*}

\subsection{Dataset Statistics}
Table \ref{tab: data} summarizes the statistics of the public real-world datasets, which we used in the few-shot experiments. For our synthetic datasets in the zero-shot prototype, we summarize their statistics in Table \ref{tab: zero data}. As discussed in Section \ref{sec: zero-shot}, the connections of our synthetic datasets are real, and we only replace the node feature by $[1, 0]$ and $[0, 1]$. The code to download the public data and the code to create synthetic data are provided in the supplementary materials.

\subsection{Text Labels}
\label{ap: text labels}
When created, real-world graph datasets are usually coupled with textual meanings, but a common practice is to convert the textual meanings into numbers to create labels, which weakens the supervision of the graph data. For each real-world dataset, we convert the numerical labels back to text labels and feed into Morpher Language encoder through ``[learnable text prompt] + [text label]''. The mapping from the numbers to text labels for each dataset are provided as follows:

\paragraph{MUTAG.} MUTAG is a dataset of nitroaromatic compounds, aiming to predict their mutagenicity on Salmonella typhimurium. Therefore, the mapping from numerical labels to text labels is: 
    \{0: non-mutagenic on Salmonella typhimurium, 1: mutagenic on Salmonella typhimurium\}.

\paragraph{ENZYMES.} ENZYMES aims to predict which subcategory each enzyme belongs to. The subcategories are:
    {0: oxidoreductases, 1: transferases, 2: hydrolases, 3: lyases, 4: isomerases, 5: ligases}.

\paragraph{PROTEINS.} PROTEINS is a dataset comprising proteins classified as either enzymes or non-enzymes. Therefore, the mapping is: {0: 'enzyme', 1: 'non-enzyme'}.

\paragraph{MSRC\_21C.} Each graph in MSRC is constructed according to an image. The graph label is the image label. MSRC\_21C contains 20 classes in MSRC, and ``C'' here means ``Challenging'' as the graphs(images) that are easy to classify has been filtered. The mapping from the numerical labels to text labels is:
    \{0: building, 1: grass, 2: tree, 3: cow, 4: sheep, 5: sky, 6: airplane, 7: water, 8: face, 9: car, 
                                10: bicycle, 11: flower, 12: sign, 13: bird, 14: book, 15: chair, 16: road\}.

\paragraph{Cora.} Cora is a citation network of papers in seven research areas. Each paper is labeled according to its corresponding research area. The mapping from the numerical labels to text labels is:
    \{0: case based, 1: genetic algorithms, 2: neural networks, 3: probabilistic methods, 4: reinforcement learning, 5: rule learning, 6: theory\}.

\paragraph{CiteSeer.} CiteSeer is a citation network of papers, each labeled according to one of six research areas. The mapping from the numerical labels to text labels is: 
\{0: Agents, 1: AI, 2: DB, 3: IR, 4: ML, 5: HCI\}. We note that using abbreviations of the research area is not an issue because these abbreviations frequently appear, and the LLM tends to tokenize each of them as one token. 

\paragraph{PubMed.} PubMed is a collection of scientific publications from the PubMed database related to diabetes, classified into one of three categories. The mapping from the numerical labels to text labels is: 
\{0: Diabetes Mellitus Experimental, 1: Diabetes Mellitus Type 1, 2: Diabetes Mellitus Type 2\}.

\paragraph{Edge-level tasks.} Cora, CiteSeer and PubMed can also be used as link prediction datasets. For link prediction, the mapping from the numerical labels to text labels is: \{0: not connected, 1: connected\}.

\paragraph{Synthetic Zero-shot Class Generalization Datasets.} For ZERO-Cora, we synthetic three classes of ego-graph in a citation network. The first and second classes, respectively, have text labels "machine learning" and "theory", and the third (novel) class to generalize is "machine learning theory". For ZERO-CiteSeer, we synthetic three classes of ego-graph in a citation network. The first and second classes, respectively, have text labels "biology" and "informatics", and the third (novel) class to generalize is "bioinformatics". For ZERO-PubMed, we synthetic three classes of ego-graph in a citation network in the medical domain. The first and second classes, respectively, have text labels "cardiology" and "neurology", and the third (novel) class to generalize is "neurocardiology".





\definecolor{LightCyan}{rgb}{0.88,1,1}

\renewcommand\arraystretch{1.4}

\begin{table*}
\caption{Comparison of graph prompts.}
\label{tb: graph prompts comparison}
\begin{center}
\begin{small}
\scalebox{0.93}{
\begin{tabular}{lcccccc}
\toprule
\multirow{2}{*}{\makecell{Method}} &  \multirow{2}{*}{\makecell{prompt level}}  & \multicolumn{3}{c}{level of supported downstream tasks} &  \multirow{2}{*}{\makecell{learnable prompt}} & \multirow{2}{*}{\makecell{semantic}} \\
                &  & node-level  & edge-level  & graph-level \\
\midrule
GPF-Plus \citep{DBLP:conf/nips/FangZYWC23}   & token-level   & $\surd$ & $\times$ & $\times$ & $\surd$ & $\times$ \\
Gprompt \citep{DBLP:conf/www/LiuY0023}  & token-level & $\surd$ & $\times$ & $\surd$ & $\surd$ & $\times$ \\
VNT \citep{DBLP:conf/kdd/TanGDL23}   & token-level   & $\times$ & $\times$ & $\surd$ & $\surd$ & $\times$ \\
ULTRA-DP \citep{DBLP:journals/corr/abs-2310-14845}   & token-level & $\surd$ & $\times$ & $\times$ & $\surd$ & $\times$ \\
GPPT \citep{DBLP:conf/kdd/SunZHWW22}   & token-level & $\surd$ & $\times$ & $\times$ & $\surd$ & $\times$ \\
SGL-PT \citep{DBLP:journals/corr/abs-2302-12449}   & token-level   & $\surd$ & $\times$ & $\times$ & $\surd$ & $\times$ \\
SAP \citep{DBLP:journals/corr/abs-2310-17394}  & graph-level    & $\surd$ & $\times$ & $\surd$ & $\surd$ & $\times$  \\
PRODIGY \citep{DBLP:conf/nips/HuangRCK0LL23} & graph-level  & $\surd$& $\surd$& $\surd$ & $\times$ & $\times$ \\
All-in-one (AIO) \citep{DBLP:conf/kdd/SunCLLG23} & graph-level  & $\surd$& $\surd$& $\surd$ & $\surd$ & $\times$ \\
\midrule
ImprovedAIO (ours) & graph-level  & $\surd$& $\surd$& $\surd$ & $\surd$ & $\times$ \\
Morpher (ours) & graph-level  & $\surd$& $\surd$& $\surd$ & $\surd$ & $\surd$ \\
\bottomrule
\end{tabular}
}
\end{small}
\end{center}
\vskip -0.1in
\end{table*}

\renewcommand\arraystretch{1}

\section{Experiment Details}
\subsection{Reproducibility}
\label{ap: reproducibility}
\paragraph{Code.} 
The code for the experiments is provided in the supplementary material with a well-written README file. We also provide the commands and instructions to run the code. The datasets used will be automatically downloaded when the code is executed.

\paragraph{Environment.}
We run all our experiments on a Windows 11 machine with a 13th Gen Intel(R) Core(TM) i9-13900H CPU, 64GB RAM, and an NVIDIA RTX A4500 GPU. We have also tested the code on a Linux machine with NVIDIA TITAN RTX GPU. All the code of our algorithms is written in Python. The Python version in our environment is 3.9.18. In order to run our code, one has to install some other common libraries, including PyTorch, PyTorch Geometric, pandas, numpy, scipy, etc. Please refer to our README in the code directory for downloading instructions.

We have optimized our code and tested that the space cost of the CPU memory is less than 16 GB, and the space cost of the graphics card is less than 6 GB. The execution time to run an experiment is less than 20 minutes on our machine.


\subsection{Implementation Details}
\label{ap: exp details}
We provide the configuration files for the experiments to reproduce the results. We initialize the graph prompt using kaiming\_initialization, and we initialize the text prompts through real token embeddings. We have tested multiple initializations, and they would not affect the overall results. Specifically, we initialize the text prompt for each dataset as follows.

MUTAG: ``a graph with property''; ENZYMES: ``this enzyme is''; PROTEINS: ``this protein is''; MSRC\_21C: ``an image of''; Cora: ``a paper of''; CiteSeer: ``a paper of''; PubMed: ``a paper of''; Edge tasks: ``central nodes are''.

In our few-shot setting, we split the labeled data into training samples and validation samples at approximately 1:1. For all the parameters, we used the Adam optimizer, whose learning rate and weight decay are provided in the configuration files.


\subsection{Experiment with ELECTRA and DistilBERT}
\label{ap: other llm}
On the LLM pre-training side, RoBERTa is one of the most advanced encoder-only LLMs until now, and we have demonstrated the effectiveness with RoBERTa serving on the LLM side in the Morpher paradigm. Additionally, we conducted experiments with ELECTRA \citep{DBLP:conf/iclr/ClarkLLM20} and DistilBERT \citep{DBLP:journals/corr/abs-1910-01108}. Using these two LLMs, Morpher can also achieve comparable performances to RoBERTa. The results are shown in Table \ref{tb: morpher with distilbert} and in Table \ref{tb: morpher with electra}.

\begin{table*}
\centering
\caption{Few-shot graph classification performance (\%) of Morpher with ELECTRA \citep{DBLP:conf/iclr/ClarkLLM20} as language encoder. Other experiment settings are identical to the main experiment.}
\label{tb: morpher with electra}
\resizebox{0.7\textwidth}{!}{%
\begin{tabular}{
ccccccccc
}
\toprule
\multirow{2}{*}{GNN pretraining} & \multicolumn{2}{c}{MUTAG}       & \multicolumn{2}{c}{ENZYMES}    & \multicolumn{2}{c}{PROTEINS}     & \multicolumn{2}{c}{MSRC\_21C}  \\
                        & Acc & F1 & Acc & F1 & Acc & F1 & Acc & F1\\ \midrule
GraphCL + GCN      & 78.00 & 78.17 & 20.41 & 15.79 & 67.38 & 65.66 & 43.42 & 47.19 \\
GraphCL + GAT      & 76.67 & 75.75 & 20.41 & 11.37 & 66.26 & 65.66 & 44.57 & 49.01 \\
GraphCL + GT       & 76.67 & 77.04 & 19.16 & 14.68 & 73.06 & 72.70 & 42.28 & 44.09 \\
SimGRACE + GCN     & 70.00 & 70.99 & 19.79 & 12.41 & 68.96 & 67.77 & 45.71 & 48.44 \\
SimGRACE + GAT     & 77.33 & 77.51 & 18.12 & 13.31 & 68.96 & 67.78 & 44.00 & 49.43 \\
SimGRACE + GT      & 72.67 & 73.55 & 18.33 & 15.76 & 70.18 & 70.28 & 41.14 & 44.50 \\
\bottomrule
\end{tabular}%
}
\end{table*}

\begin{table*}
\centering
\caption{Few-shot graph classification performance (\%) of Morpher with DistilBERT \citep{DBLP:journals/corr/abs-1910-01108} as language encoder. Other experiment settings are identical to the main experiment.}
\label{tb: morpher with distilbert}
\resizebox{0.7\textwidth}{!}{%
\begin{tabular}{
ccccccccc
}
\toprule
\multirow{2}{*}{GNN pretraining} & \multicolumn{2}{c}{MUTAG}       & \multicolumn{2}{c}{ENZYMES}    & \multicolumn{2}{c}{PROTEINS}     & \multicolumn{2}{c}{MSRC\_21C}  \\
                        & Acc & F1 & Acc & F1 & Acc & F1 & Acc & F1\\ \midrule
GraphCL + GCN      & 78.00 & 78.61 & 20.62 & 10.00 & 66.44 & 65.54 & 43.42 & 47.98 \\
GraphCL + GAT      & 77.33 & 75.64 & 21.25 & 15.87 & 70.59 & 68.25 & 45.14 & 48.82 \\
GraphCL + GT       & 74.67 & 75.20 & 19.58 & 14.96 & 70.27 & 70.55 & 44.57 & 47.28 \\
SimGRACE + GCN     & 69.33 & 70.36 & 20.62 & 18.82 & 66.91 & 66.41 & 45.14 & 47.77 \\
SimGRACE + GAT     & 77.33 & 76.90 & 18.54 & 14.44 & 67.56 & 65.08 & 45.71 & 44.36 \\
SimGRACE + GT      & 72.67 & 73.52 & 17.91 & 11.06 & 70.55 & 70.36 & 45.14 & 44.01 \\
\bottomrule
\end{tabular}%
}
\end{table*}

In general, using ELECTRA and DistilBERT results in similar performance compared to using RoBERTa, showing the robustness of Morpher with respect to the language encoder.

\subsection{Experiment with GNNs trained using GraphMAE and MVGRL}
\label{ap: other pretrain}
In the main pages, we used GraphCL and SimGRACE to show that Morpher achieves better performance given a pre-trained GNN. Additionally, to further verify the robustness of Morpher over the pre-train method, we conducted experiments on the pre-trained GNNs using GraphMAE \citep{DBLP:conf/kdd/HouLCDYW022} and MVGRL \citep{DBLP:conf/icml/HassaniA20}. We use GCN as the GNN backbone and RoBERTa as the LLM encoder, and the results are reported in Table \ref{TB: GraphMAE results} and Table \ref{TB: MVGRL results}.

\renewcommand\arraystretch{1.2}

\begin{table*}[h]
\centering
\caption{Few-shot graph classification performance (\%) of Morpher with the GNN pre-trained by GraphMAE \citep{DBLP:conf/kdd/HouLCDYW022}. Other experiment settings are identical to the main experiment.}
\label{TB: GraphMAE results}
\resizebox{0.7\textwidth}{!}{%
\begin{tabular}{
ccccccccc
}
\toprule
\multirow{2}{*}{GNN pretraining} & \multicolumn{2}{c}{MUTAG}       & \multicolumn{2}{c}{ENZYMES}    & \multicolumn{2}{c}{PROTEINS}     & \multicolumn{2}{c}{MSRC\_21C}  \\
                        & Acc & F1 & Acc & F1 & Acc & F1 & Acc & F1\\ \midrule
Pre-train + Fine-tune   & 71.33 & 71.41 & 16.04 & 12.14 & 65.86 & 65.22 & 39.42 & 40.20 \\
ImprovedAIO             & 76.67 & 76.95 & 19.58 & 12.59 & 66.36 & 65.30 & 42.28 & 46.81 \\
Morpher                 & 78.67 & 78.67 & 20.20 & 16.95 & 67.38 & 65.66 & 45.71 & 48.49 \\
\bottomrule
\end{tabular}%
}
\end{table*}

\begin{table*}[h]
\centering
\caption{Few-shot graph classification performance (\%) of Morpher with the GNN pre-trained by MVGRL \citep{DBLP:conf/icml/HassaniA20}. Other experiment settings are identical to the main experiment.}
\vspace{1mm}
\label{TB: MVGRL results}
\resizebox{0.7\textwidth}{!}{%
\begin{tabular}{
ccccccccc
}
\toprule
\multirow{2}{*}{GNN pretraining} & \multicolumn{2}{c}{MUTAG}       & \multicolumn{2}{c}{ENZYMES}    & \multicolumn{2}{c}{PROTEINS}     & \multicolumn{2}{c}{MSRC\_21C}  \\
                        & Acc & F1 & Acc & F1 & Acc & F1 & Acc & F1\\ \midrule
Pre-train + Fine-tune   & 68.67 & 69.46 & 16.45 & 10.16 & 65.15 & 64.71 & 38.85 & 40.56 \\
ImprovedAIO             & 74.67 & 74.00 & 18.13 & 15.57 & 66.54 & 65.90 & 42.85 & 46.66 \\
Morpher                 & 78.00 & 77.81 & 18.96 & 14.97 & 67.56 & 66.79	& 44.57 & 48.67 \\
\bottomrule
\end{tabular}%
}
\end{table*}

\renewcommand\arraystretch{1}

Using GraphMAE or MVGRL to pre-train the GNN, the trend of performance is similar to that when using GraphCL or SimGRACE. Also, ImprovedAIO and Morpher's performance is similar to that of pre-trained GNNs from GraphCL or SimGRACE and can still significantly outperform the pre-train + fine-tune baseline, showing the robustness of Morpher with respect to the pre-training strategy.

\subsection{Morpher on MolecureNet with More Text Supervision}
The relatively abundant labeled graph-text pairs in medical and biological domains have significantly accelerated research into training large models specifically tailored for these areas \cite{DBLP:journals/titb/QiuLSPSZDLLXYWXL23}. For example, \cite{DBLP:conf/chil/ParkBKKC22} pretrains the multimodal medical model MedGTX for Electronic Health Records (EHR) using the open-source EHR dataset MIMIC-III. Similarly, ProGen \cite{DBLP:journals/corr/abs-2004-03497} trains a 1.2B-parameter language model on approximately 280M protein sequences. HyenaDNA \cite{DBLP:conf/nips/NguyenPFTWBMPRB23} pretrains a genomic foundation model on the human reference genome with context lengths of up to 1 million tokens. Furthermore, \cite{DBLP:journals/corr/abs-2212-13138} introduces the MultiMedQA benchmark and performs instruction-tuning on the 540-billion parameter Flan-PaLM model within the clinical domain.

We demonstrate that, though not specifically designed for any downstream applications, the Morpher framework has the potential to be used in various tasks where there is more text supervision compared to previous experiments. As for a case study, We use bace (inhibitors of human beta-secretase), tox21 (toxicology in the 21st century) and hiv (inhibit HIV replication) from MolecureNet \citep{DBLP:journals/corr/WuRFGGPLP17}. These three datasets have 1513, 7831, and 41127 graphs to classify, respectively. In these datasets, each graph label is associated with a text description. The tasks on bace and hiv are bio-activity prediction and the task on tox21 is toxicity prediction. To adopt Morpher, we use GraphCL to pre-train the GAT model and initialize the text prompts and text labels using those from GIMLET \citep{DBLP:conf/nips/ZhaoLMXFDKL23}.

\begin{table*}
\centering
\caption{AUC-ROC $(\uparrow)$ on MolecureNet (bace, tox21, hiv). Morpher-K denotes K shots.}
\resizebox{0.98\textwidth}{!}{%
\begin{tabular}{
ccccccccc
}
\toprule   Dataset  & KVPLM & MoMu & Galactica-1.3B & GIMLET-64M-50-shots & GAT-1M-supervised & Morpher-10 & Morpher-20 & Morpher-50 \\       
\midrule
bace      & 0.5126 & 0.6656 & 0.5648 & 0.729 & 0.697 & 0.6231 & 0.6513 & 0.6858 \\
tox21     & 0.4917 & 0.5757 & 0.4946 & 0.652 & 0.754 & 0.6769 & 0.7275 & 0.7459 \\
hiv       & 0.6120 & 0.5026 & 0.3385 & 0.721 & 0.729 & 0.5742 & 0.7034 & 0.7283 \\
\bottomrule
\end{tabular}%
}
\end{table*}

KVPLM \citep{zeng2022deep}, MoMu \citep{DBLP:journals/corr/abs-2209-05481}, Galactica-1.3B \citep{DBLP:journals/corr/abs-2211-09085} are zero-shot predictors for the three tasks; GIMLET-64M-50-shots is the GIMLET \citep{DBLP:conf/nips/ZhaoLMXFDKL23} model fine-tuned on 50 additional training samples\footnote{the performance of GIMLET and other baselines are directly from the GIMLET paper \citep{DBLP:conf/nips/ZhaoLMXFDKL23}.}; GAT-1M-fully-supervised uses all the training data to train a GAT. Our Morpher-k-shots uses only k training samples. From the results, first, using only 10 training samples, Morpher can outperform the zero-shot baselines KVPLM, MoMu, and Galactica-1.3B. Second, using only 50 shots, Morpher can achieve similar performance with the fully supervised GAT. Third, using the same amount of few-shot data (50 shots), Morpher-50 outperforms GIMLET-64M-50-shots on tox21 and hiv, the two largest datasets among the three. This means our graph-text multi-modal prompt learning, with much fewer learnable parameters ($\sim50K$), is more sample-efficient than fine-tuning language model encoder.

\subsection{Full Table for Few-shot Experiment}
Due to the page limitation of the main pages, Table \ref{TB: main comparison full} shows the full table for the few-shot experiment.

\begin{table}[h]
    \centering
    \resizebox{0.47\textwidth}{!}{%
    \begin{tabular}{lcccc}
        \toprule
        & MUTAG & ENZYMES & PROTEINS & MSRC\_21C \\
        \midrule
        default variance  & 64.67 & 17.50 & 59.21 & 14.37 \\
        3$\times$ variance       & 64.67 & 17.50 & 61.79 & 13.17 \\
        5$\times$ variance       & 68.00 & 17.70 & 59.21 & 11.37 \\
        10$\times$ variance      & 67.33 & 16.45 & 58.65 & 17.96 \\
        \bottomrule
    \end{tabular}
    }
    \caption{Accuracy results of high-variance initialization experiments in few-shot learning.}
    \label{TB: high_variance_results}
\end{table}

\subsection{Full Table for Domain Transfer}
In the main pages, through the experiments presented in Tables 1 and 2, we have already demonstrated that fine-tune outperforms supervised, and ImprovedAIO significantly outperforms the original AIO. Therefore, in Table \ref{TB: transfer}, we focus on comparing Morpher, ImprovedAIO and fine-tune methods to avoid redundancy. Here, we report the full comparison with the performance of supervised and AIO baselines. in Table \ref{TB: transfer full}. The result is consistent that fine-tune outperforms supervised, and our improvedAIO outperforms AIO.

\begin{table}[h]
        \centering
        \resizebox{0.42\textwidth}{!}{%
        \begin{tabular}{@{}cccccccc@{}}
        \toprule
        \multicolumn{2}{c}{\makecell[c]{Target Domain}}    & \multicolumn{2}{c}{MUTAG}    & \multicolumn{2}{c}{PubMed}    \\ \midrule
        \multicolumn{2}{c}{\makecell[c]{Target Task}}    & \multicolumn{2}{c}{graph-level}    & \multicolumn{2}{c}{node-level}    \\ \midrule
        Source                        & Methods & Acc & F1 & Acc & F1  \\ \midrule
        \multirow{4}{*}{\makecell[c]{ENZYMES \\ (graph-level)}} 
                                     & Supervised      & 66.00         & 56.67     & 47.57  & 36.07      \\
                                     & Fine-tune      & 68.00         & 55.04     & 47.57  & 36.07      \\
                                     & AIO     & 64.00         & 54.50     & 44.85  & 34.13        \\ 
                                     & ImprovedAIO     & \underline{70.67}         & \underline{64.07}     & \underline{50.28}  & \underline{50.51}      \\ 
                                     & Morpher     & \textbf{72.67}         & \textbf{73.29}     & \textbf{54.42}  & \textbf{53.96}         \\ 
        \midrule
        \multirow{4}{*}{\makecell[c]{CiteSeer \\ (node-level)}} 
                                     & Supervised      & 66.00         & 56.67     & 47.57  & 36.07      \\
                                     & Fine-tune      & 71.33         & 62.19     & 48.71  & 40.66         \\
                                     & AIO     & 65.33         & 57.20     & 45.71  & 34.39       \\ 
                                     & ImprovedAIO     & \underline{74.00}         & \underline{73.76}     & \underline{52.57}  & \underline{51.29}       \\ 
                                     & Morpher     & \textbf{76.67}         & \textbf{77.04}     & \textbf{58.29}  & \textbf{57.54}         \\
        \midrule
        \end{tabular}%
        }
        \vspace{-2mm}
        \caption{Domain Transfer Performance. Best results are bolded and second-best results are underlined.}\label{TB: transfer full}
\vspace{-3mm}
\end{table}

We would like to acknowledge that AIO demonstrates state-of-the-art performance under the experimental settings reported in its original paper \cite{DBLP:conf/kdd/SunCLLG23}. However, our evaluation is conducted under more challenging conditions, particularly with fewer training and validation samples. In these settings, we observe that AIO's performance degrades, suggesting there is still room for improvement. To address this, we propose ImprovedAIO, which extends AIO's design to better handle these harder scenarios. Our goal is not to critique prior work, but to help advance the field of graph prompt learning and multimodal alignment by pushing toward more robust and generalizable solutions.

\input{tables/few_shot_learning_full}

\section{Further Discussions}
\subsection{Unstable Training of Current Graph Prompt Design}
\label{ap: current graph prompt design}
As analyzed in Section \ref{sec: graph prompt}, the current graph prompt design suffers from unstable training due to the imbalance of inner-connections and cross-connections because for any node $i$ and token $p_j$, the dot products $\bm{X}(i, :) \bm{P}_\theta^g(j, :)^\top$ is close to $0$, leading to ineffective prompt-token interactions. A potential solution to ensure $\bm{X}(i, :) \bm{P}_\theta^g(j, :)^\top$ has a larger nonzero value is to initialize prompt tokens with higher variance. Through further analysis and experimental validation, We tend to believe that it fails to address the root cause.

First, initializing parameters with high variance can introduce additional challenges during training, such as unstable gradients, over-reliance on the initial high-variance parameters, and ineffective weight regularization. Specifically, in the prompt learning setting, high-variance initialization of prompts may cause the training process to overly focus on the prompt embeddings, thereby overshadowing the information encoded in the input graph and hindering the model's ability to learn meaningful representations from the input graph structure.

Second, even with high-variance initialization, the computation of
$\sigma(\bm{X}(i, :) \bm{P}_\theta^g(j, :)^\top)$ would still result in approximately half of the cross-connections being established. While this is an improvement over the original AIO, where nearly all cross-connections are formed, it would still lead to the prompted graph representations being overly similar. As a result, the task head cannot effectively learn to distinguish between different graphs.

To further validate this analysis, we conducted additional experiments using high-variance initialization in the original AIO method. These experiments were performed in the few-shot learning setting (Table~\ref{TB: high_variance_results}) using GraphCL pretraining and a GAT encoder.

As shown in Table \ref{TB: high_variance_results}, increasing the initialization variance does not consistently improve performance. In some cases, it even leads to degradation, likely due to training instability. Furthermore, the overall performance of AIO remains significantly lower than that of our ImprovedAIO and Morpher, demonstrating that high-variance initialization is not a sufficient solution to the dense cross-connection issue.

Our ImprovedAIO effectively addresses the training issue of graph prompt learning without introducing any new thresholding hyperparameters. Due to our pruning mechanism, ImprovedAIO is less sensitive to the choice of $\delta_{cross}$. Specifically, a smaller $\delta_{cross}$ can safely be used in our framework, as it allows more cross edges to be initially introduced without affecting the final set of cross edges after pruning.

\subsection{Scalability of GNNs and LLMs}

The scalability of our proposed method across different sizes of GNNs and LLMs is an important consideration. However, the primary focus of this work is to introduce multimodal prompt learning for GNNs and validate the effectiveness of our novel paradigm, Morpher.

Regarding GNN scalability, the scale of a GNN is highly dependent on the total available samples for self-supervised pretraining. In our experiments, we employed GNNs with up to 10M parameters, which is already relatively large for the datasets we used. Notably, many recent works achieving state-of-the-art performance on similar classification tasks use GNNs of comparable or smaller scale \cite{DBLP:journals/corr/abs-2309-16014, DBLP:journals/corr/abs-2306-03561, DBLP:conf/nips/KreuzerBHLT21, DBLP:conf/nips/FrascaBBM22, DBLP:journals/pami/BouritsasFZB23}.

For LLM scalability, as mentioned in the limitation section of our paper, our method requires a language encoder and has not yet been integrated with very large decoder-only LLMs. Nonetheless, our experiments with RoBERTa (0.1B parameters), ELECTRA (0.3B parameters), and DistilBERT (0.06B parameters) in our ablation study demonstrate that Morpher is robust across a variety of language encoders with different designs and scales.

Our contributions in this paper include improving the design of graph prompts and introducing a multimodal prompt learning paradigm for GNNs, tailored to real-world scenarios where text supervision is extremely weak. Through extensive experiments, we have demonstrated that our proposed methods outperform state-of-the-art baselines, establishing a robust and effective framework for multimodal learning.
While scalability across larger GNNs and LLMs is an important direction, it is beyond the scope of this work and we highlight scalability as a potential future research to extend Morpher to even larger models.

\subsection{Model and Data Scaling Laws}
Scaling laws and emergence ability have attracted much research interest recently \cite{DBLP:journals/corr/abs-2001-08361}. To this end, we conduct additional experiments regarding the data scaling capability and model scaling capability using GraphCL+GCN on the MUTAG dataset. 

\textbf{Scaling law with respect to model size.} Since our Morpher does not require the pretraining of large language models (LLMs), we report the performance using various LLMs of different sizes in Table \ref{TB: scaling_with_model_size}. Next, we pre-train GNNs of varying sizes by adjusting parameters such as the hidden dimension, while keeping the language model fixed to RoBERTa, and report the results in Figure \ref{TB: scaling_with_gnn_size}. Based on these results, we did not observe significant scaling laws with respect to the size of the LLM or GNN. We hypothesize that this may be because the sizes of the LLM and GNN have not yet reached the threshold for exhibiting emergent capabilities. Further investigation into this phenomenon is an interesting direction for future work.

\begin{table}[h]
    \centering
    \resizebox{0.47\textwidth}{!}{%
    \begin{tabular}{lcc}
        \toprule
        Language Model Size & MUTAG Acc & MUTAG F1 \\
        \midrule
        DistilBERT 0.06B  & 78.00 & 78.61 \\
        RoBERTa 0.1B       & 78.67 & 78.09 \\
        ELECTRA 0.3B       & 78,00 & 78.17 \\
        \bottomrule
    \end{tabular}
    }
    \caption{Scaling law with respect to language model size.}
    \label{TB: scaling_with_model_size}
\end{table}

\begin{table}[h]
    \centering
    \resizebox{0.40\textwidth}{!}{%
    \begin{tabular}{ccc}
        \toprule
        GCN Model Size & MUTAG Acc & MUTAG F1 \\
        \midrule
        1M  & 78.00 & 78.56 \\
        3M       & 78.67 & 78.97 \\
        10M       & 78.67 & 78.09 \\
        \bottomrule
    \end{tabular}
    }
    \caption{Scaling law with respect to GCN model size.}
    \label{TB: scaling_with_gnn_size}
\end{table}

\textbf{Scaling law with respect to data size}
We conducted additional studies by adjusting either the pre-training data size or the downstream few-shot data size. To evaluate the effect of pre-training data size, we used GraphCL+GCN on the MUTAG dataset, randomly selecting k\% of the samples for GNN pre-training. The results are presented in Table \ref{TB: scaling_with_data_size}. The results suggest that increasing the size of either the pre-training data or the downstream few-shot data generally improves the performance of our Morpher. This observation is consistent with typical data scaling laws. While the current findings provide valuable insights, further investigation is required to explore the detailed effects of data scaling, which we leave for future work.

\begin{table}[h]
    \centering
    \resizebox{0.43\textwidth}{!}{%
    \begin{tabular}{ccc}
        \toprule
        Pretrain Data Ratio & MUTAG Acc & MUTAG F1 \\
        \midrule
        10\%  & 72.00 & 68.80 \\
        30\%       & 75.33 & 75.11 \\
        100\%       & 78.67 & 78.09 \\
        \bottomrule
    \end{tabular}
    }
    \caption{Scaling law with respect to data size.}
    \label{TB: scaling_with_data_size}
\end{table}

\subsection{Tunable pre-trained GNN/LLM Scenario}

Generally, in NLP, prompt or prefix tuning is often employed as a parameter-efficient alternative to fine-tuning, especially when fine-tuning the entire model is computationally expensive. While prompt tuning and fine-tuning are not technically incompatible, they are typically not used simultaneously, as the goal of prompt tuning is to achieve strong performance without the need to update the entire model.

That being said, making the pre-trained GNN tunable could potentially further enhance performance in specific scenarios. However, this would come at the cost of increased computational complexity and resource requirements, which goes against the motivation of our proposed method. We designed Morpher with efficiency in mind, ensuring that it achieves strong performance without requiring extensive updates to the pre-trained GNN. Furthermore, given the weak text supervision in our setting, increasing the parameter space by making the GNN tunable could reduce the sample efficiency of prompt tuning, potentially hindering the model's ability to learn effectively from limited supervision.

Therefore, while the scenario where the GNN—and potentially the language model—are also tunable is an interesting direction, it falls beyond the scope of this paper. Nonetheless, we acknowledge this as an open question that warrants further exploration in future research, particularly to better understand the trade-offs between parameter efficiency and model expressiveness.

\subsection{Broader Impact and Future Directions}
Learning on graphs has been a long-standing goal in the machine learning community, evolving from pattern-based mining \cite{DBLP:conf/kdd/LiFAH25} to modern graph neural network models \cite{zheng2024drgnn, DBLP:conf/iclr/WangHZFYCHWYL25}, with broad applications in social network analysis \cite{eveppr}, natural sciences \cite{climatebenchm}, and beyond \cite{DBLP:conf/sigir/LiAH24, llmgraph, DBLP:journals/corr/abs-2409-11585, lin2025moralise, jin2024scam}. 
In addition to advancing graph-language alignment and graph prompt learning, we hope this work can also inspire future research in the following directions.

\textbf{Distribution Shift in Graph Data}.
Real-world graph data often undergoes distribution shifts in both node features and graph structures, which can severely degrade the performance of GNNs. Before foundational GNNs, to address this challenge, graph domain adaptation aims to adapt a pretrained GNN model to a target graph via either model adaptation~\cite{baomatcha, wu2023non,guo2022learning, tieu2025invariant} or data adaptation~\cite{liu2023structural,wei2022augmentations,DBLP:conf/nips/Lin0FQT24,zeng2025pave, DBLP:conf/iclr/0003FTMH25}.

\textbf{Graph Foundation Models versus Domain Specific GNNs.}
In the era of big data and AI \cite{liu2020self,wei2020fast,liu2025selfelicit, DBLP:conf/iclr/FuHXFZSWMHL24, DBLP:journals/corr/abs-2504-01346}, graph foundation models play an important role in many applications, such as network alignment \cite{yan2021dynamic,yan2022dissecting,zeng2024hierarchical,yu2025joint}, spectral graph signal processing \cite{xu2024slog,liu2024class}, anomaly detection~\cite{DBLP:conf/www/ZhengBWZH25, DBLP:conf/www/ZhengCHC24}, multi-layered network embedding \cite{yan2024thegcn,yan2024pacer,yan2024topological,jing2022coin,jing2024sterling}, information retrieval~\cite{wei2021model,yoo2024ensuring,liu2024collaborative} and time series analysis \cite{roach2020canon,DBLP:conf/kdd/FuFMTH22, wang2023networked,lin2025cats,DBLP:conf/nips/BanZLQFKTH24, DBLP:conf/nips/TieuFZHH24, DBLP:journals/corr/abs-2408-04254}. However, as noted earlier, graph foundation models still face challenges under distribution shifts. In practice, many real-world applications continue to rely on domain-specific GNN solutions when sufficient training data is available. While this work introduces a method for training graph foundation models under extremely weak text supervision, we still recommend evaluating domain-specific GNNs as a baseline when developing real-world systems.

\textbf{Multimodal Learning with Graphs.}
Multi-modality learning~\cite{DBLP:conf/www/ZhengCYCH21, DBLP:conf/sdm/ZhengZH23, li2025language,wei2024towards} has been studied for decades. Inspired by the success of large language model~\cite{DBLP:conf/icml/RadfordKHRGASAM21, DBLP:conf/emnlp/LanZMK24,ai2025resmoe}, a new trend of multimodal learning with graphs~\cite{DBLP:conf/kdd/ZhengJLTH24} is to align the text representation with graph structure for text-attributed graphs~\cite{DBLP:conf/sigir/Wen023} or bridge molecular graphs and text data with semantic alignment for text-paired graphs~\cite{DBLP:conf/emnlp/LiuLL00K0C23}. Our work presents an alignment with graphs and languages, and similar frameworks can be expanded into graphs with other modalities. 

\textbf{Semantic-aware Graph Generative Modeling.}
Graph generation models~\cite{kong2023autoregressive,vignac2022digress,xu2024discrete,zeng2023generative} have a longstanding history with wide applications in many domains. These methods aim to capture and reproduce one or more important structure properties, such as community structure~\cite{DBLP:journals/fdata/ZhouZXH19}, motif distribution~\cite{DBLP:conf/icde/Zheng0TXZH24}, and densification in graph evolution~\cite{DBLP:conf/kdd/ZhouZ0H20}. In this work, we present a prototype of a semantic-aware graph predictive model, and semantic-aware graph generative models could also be a future direction, which may hold great potential in conditioned generation.

\section{Full-Resolution Figures}
\label{ap: full resolution figures}
Due to space constraints in the main text, we resized the figures for a more compact presentation in Figure \ref{fig: zero and emb}. We provide the full-resolution versions of the figures here for finer details.

\clearpage
\begin{figure*}
    \centering
    \includegraphics[width=0.9\linewidth]{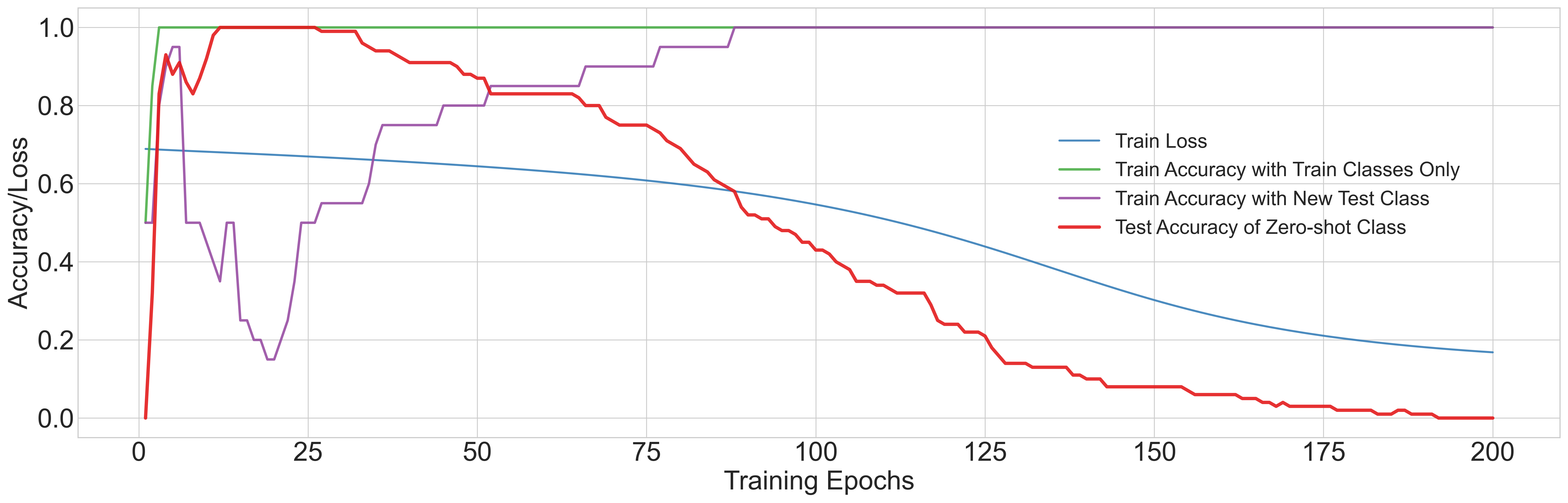}
    \caption{Novel class generalization result for our ZERO-Cora dataset.}
\end{figure*}

\begin{figure*}
    \centering
    \includegraphics[width=0.9\linewidth]{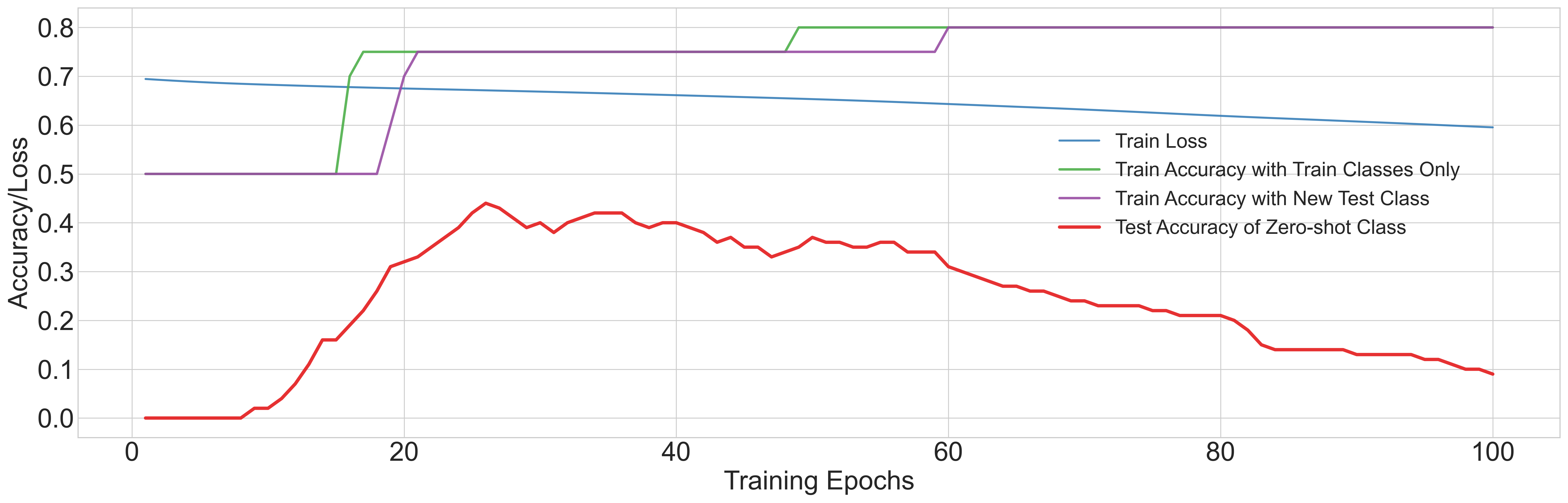}
    \caption{Novel class generalization result for our ZERO-CiteSeer dataset.}
\end{figure*}

\begin{figure*}
    \centering
    \includegraphics[width=0.9\linewidth]{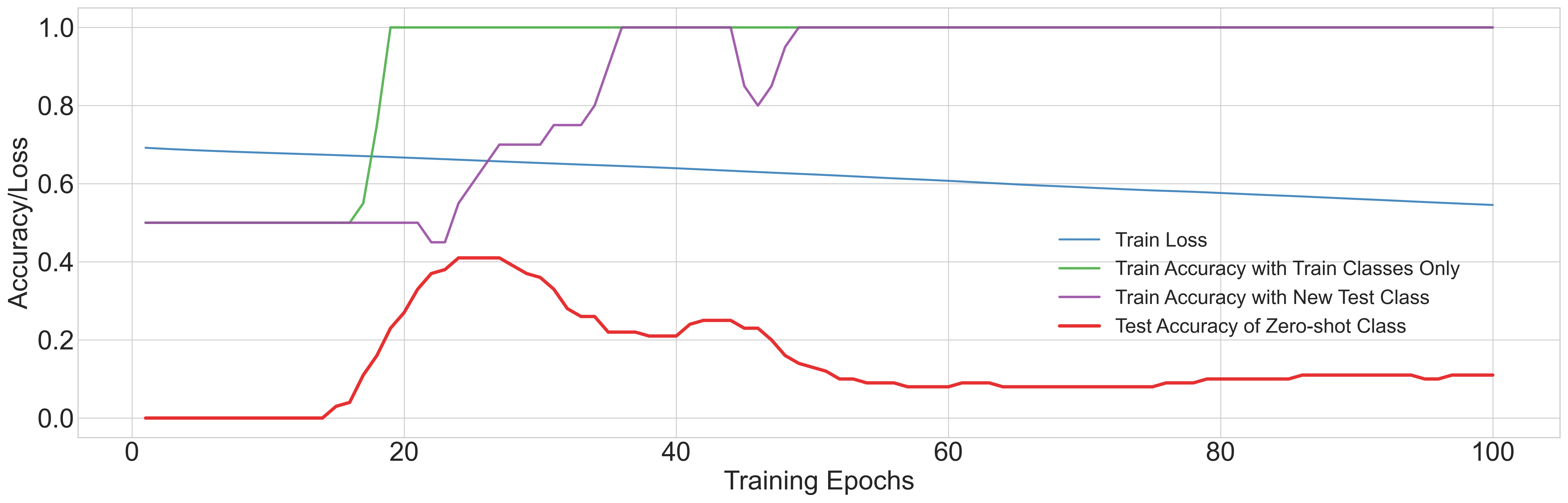}
    \caption{Novel class generalization result for our ZERO-PubMed dataset.}
\end{figure*}

\clearpage


\begin{figure*}
    \centering
    \begin{tabular}{cc}
        \subcaptionbox{Fine-tune on CiteSeer}{\includegraphics[width=0.43\textwidth]{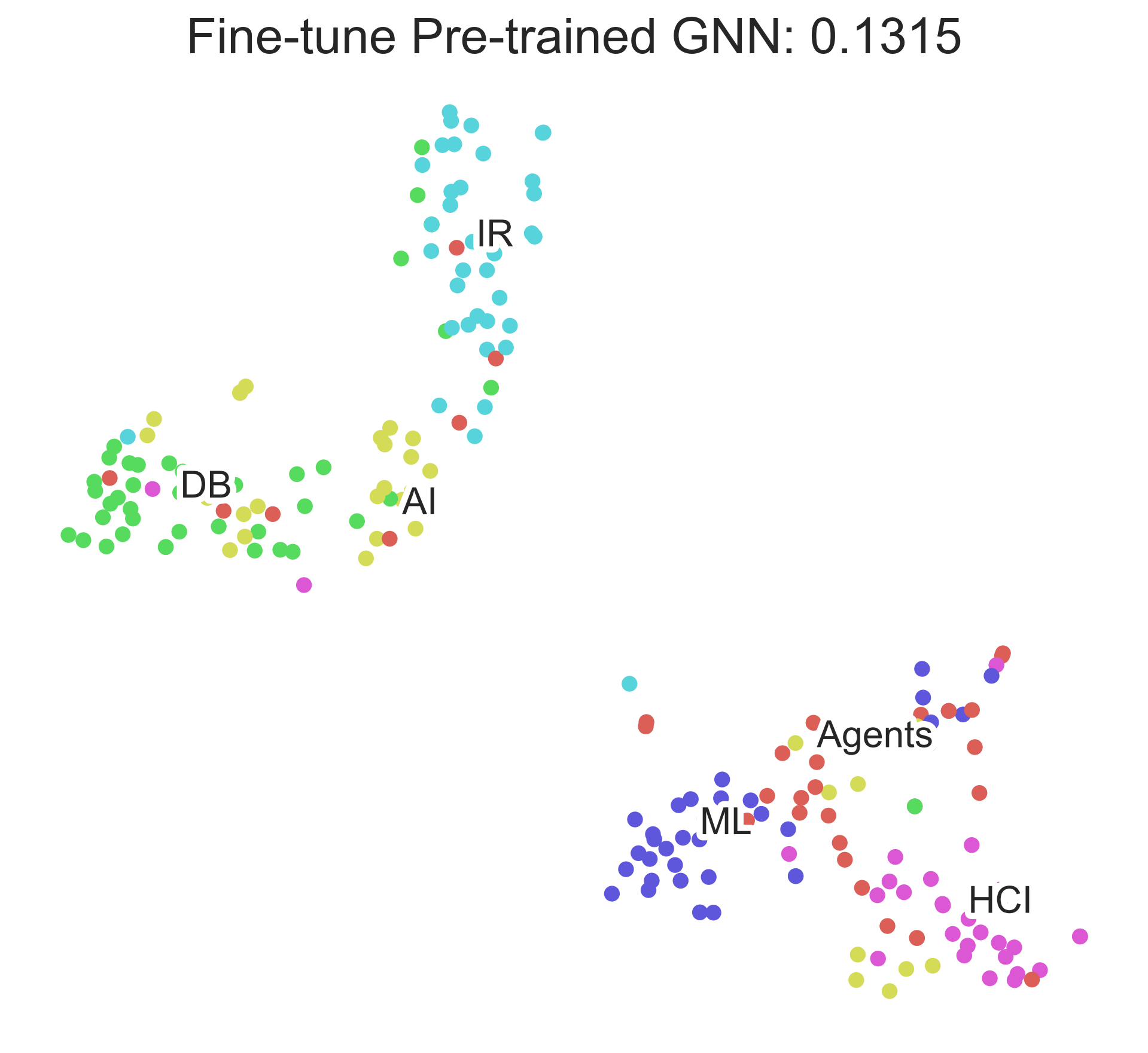}} &
        \subcaptionbox{Fine-tune on MSRC\_21C}{\includegraphics[width=0.43\textwidth]{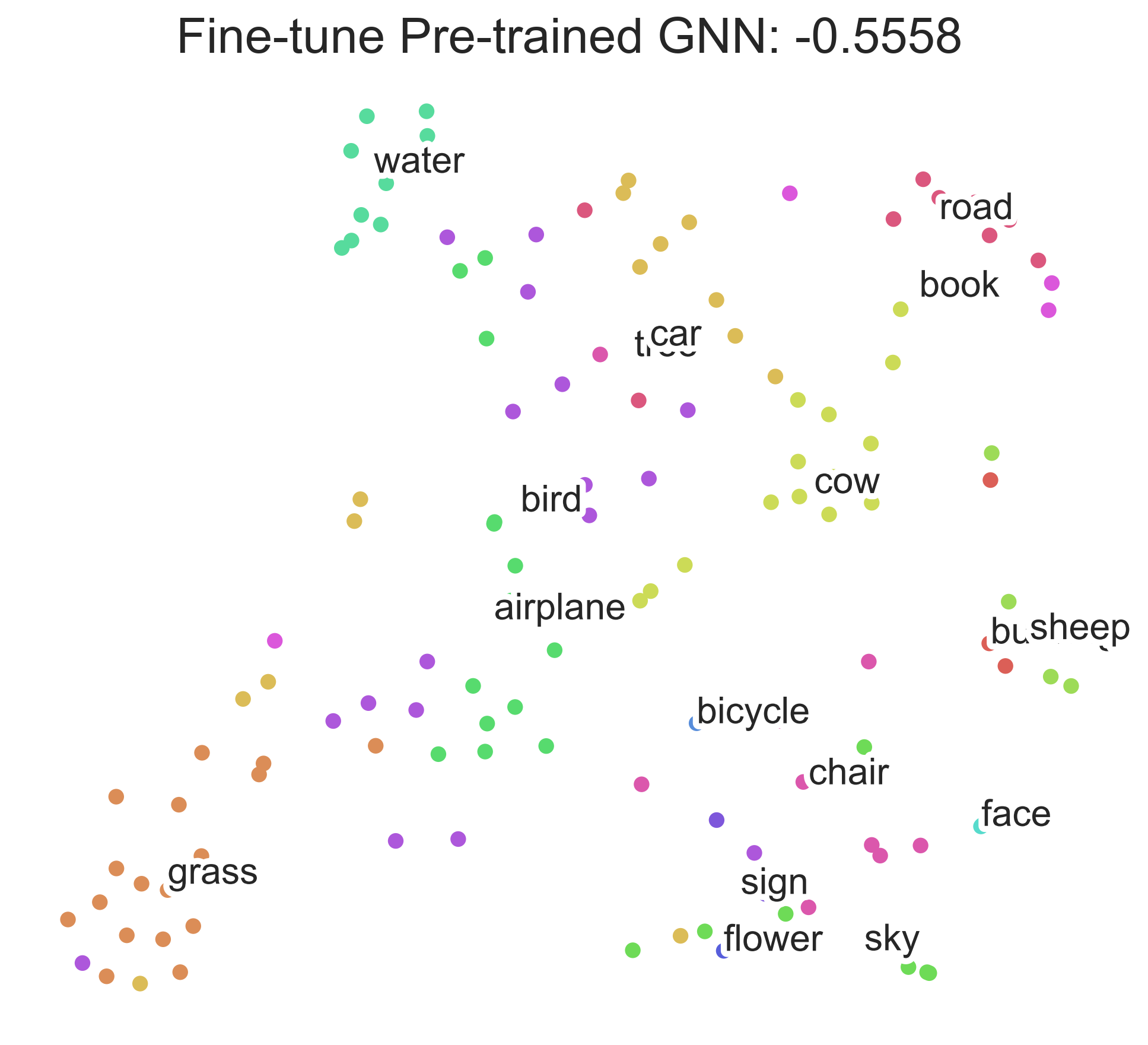}} \vspace{10mm}\\
        
        \subcaptionbox{ImprovedAIO on CiteSeer}{\includegraphics[width=0.43\textwidth]{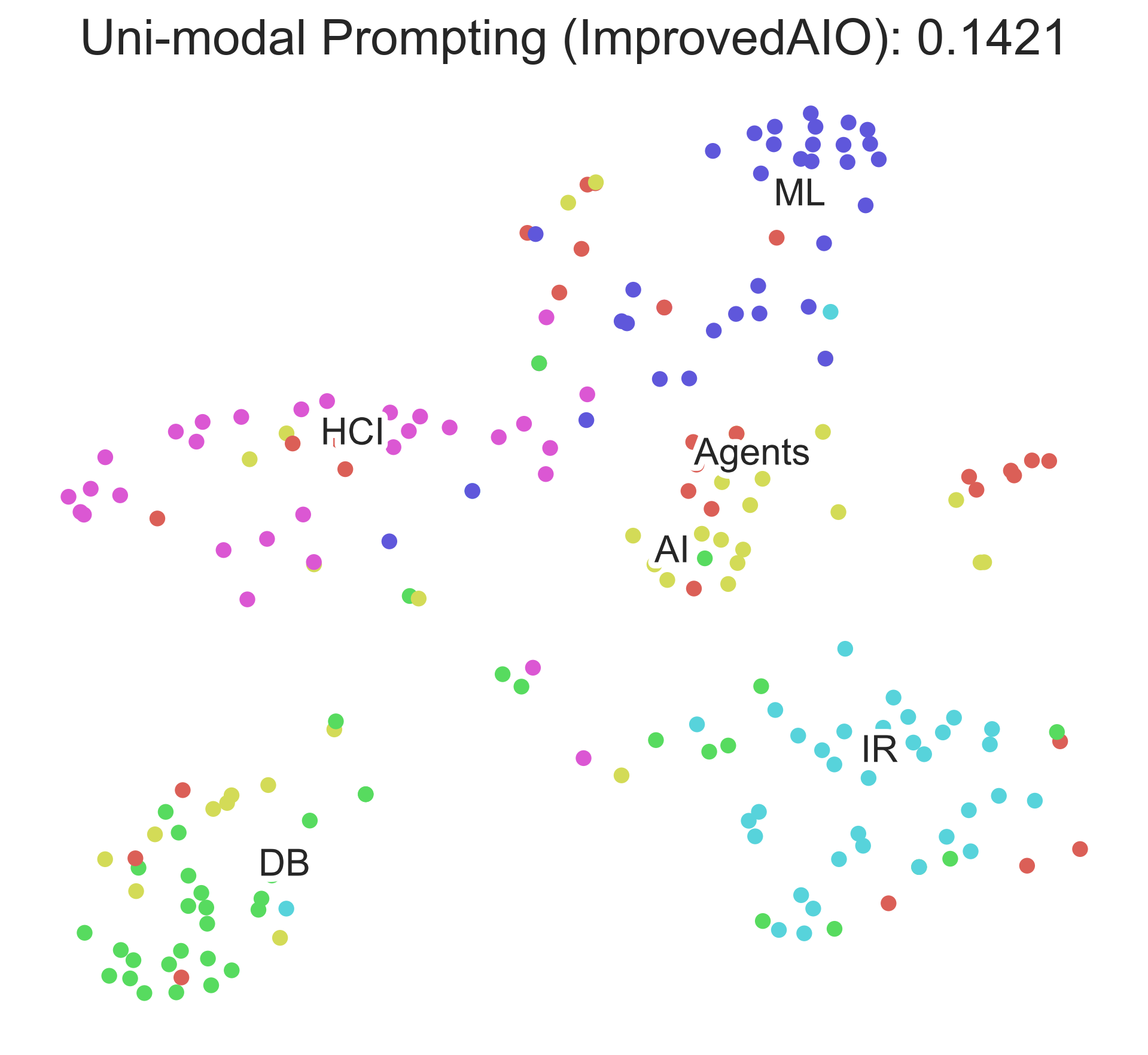}} &
        \subcaptionbox{ImprovedAIO on MSRC\_21C}{\includegraphics[width=0.43\textwidth]{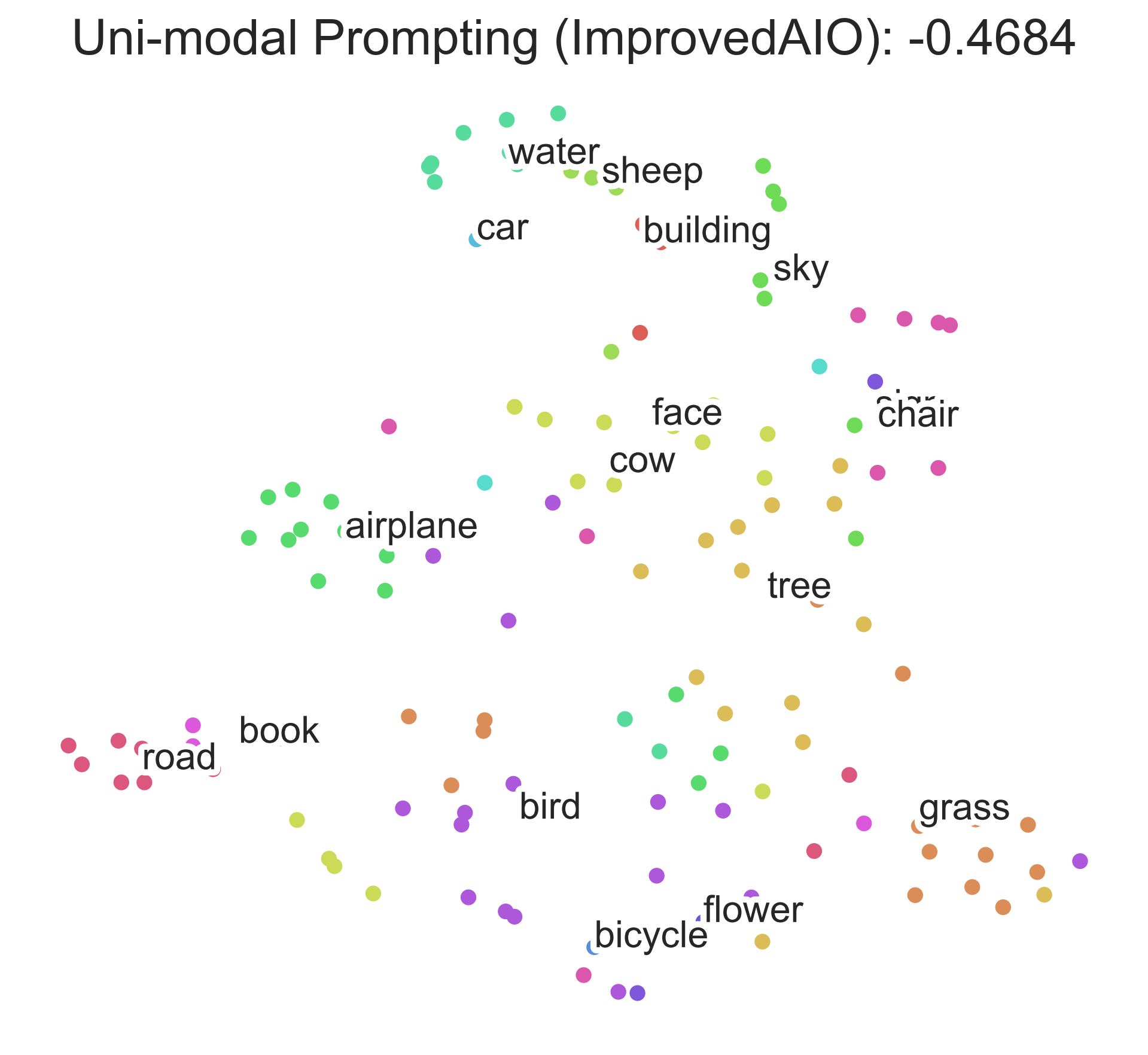}} \vspace{10mm}\\
        
        \subcaptionbox{Morpher on CiteSeer}{\includegraphics[width=0.43\textwidth]{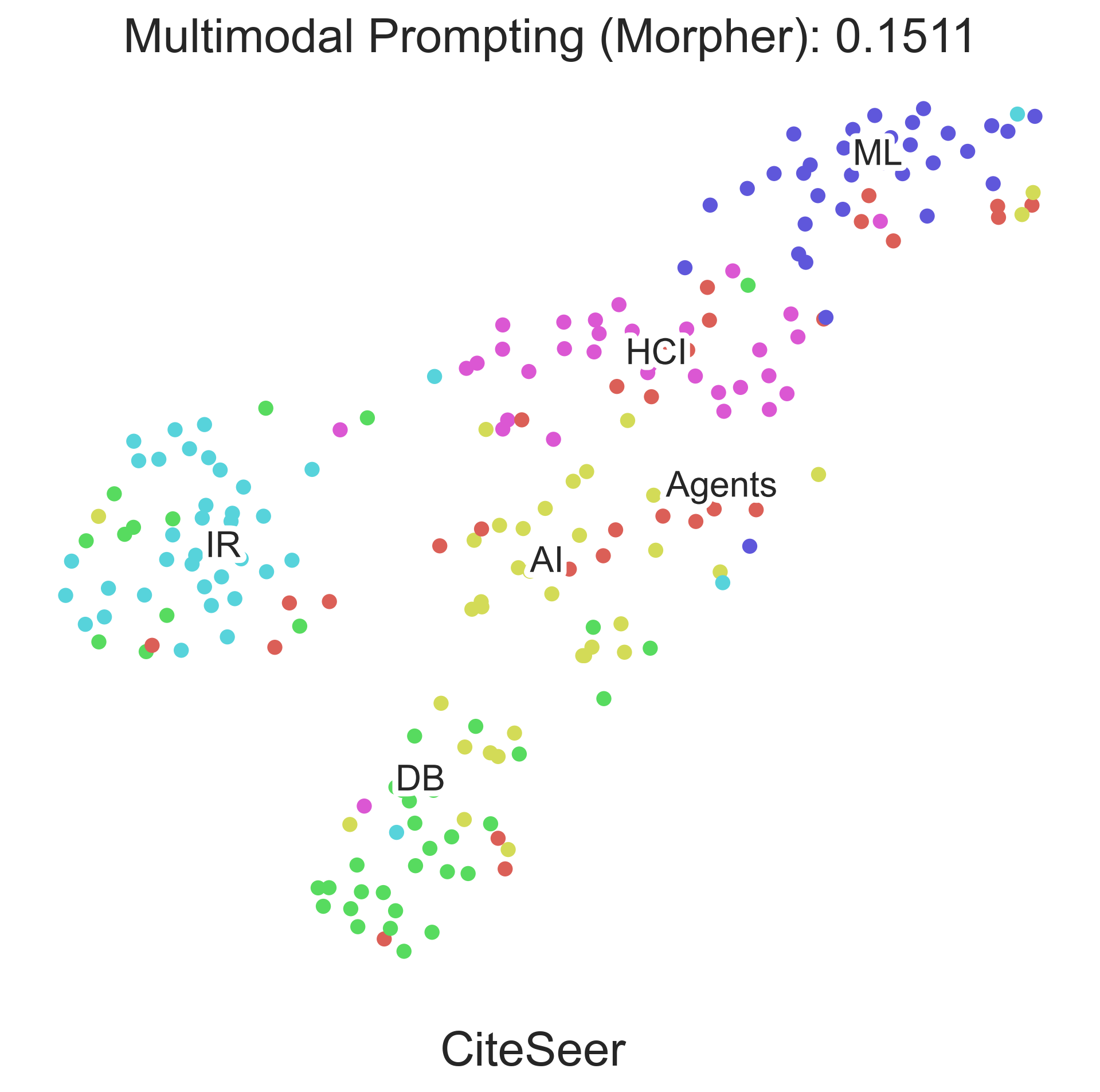}} &
        \subcaptionbox{Morpher on MSRC\_21C}{\includegraphics[width=0.43\textwidth]{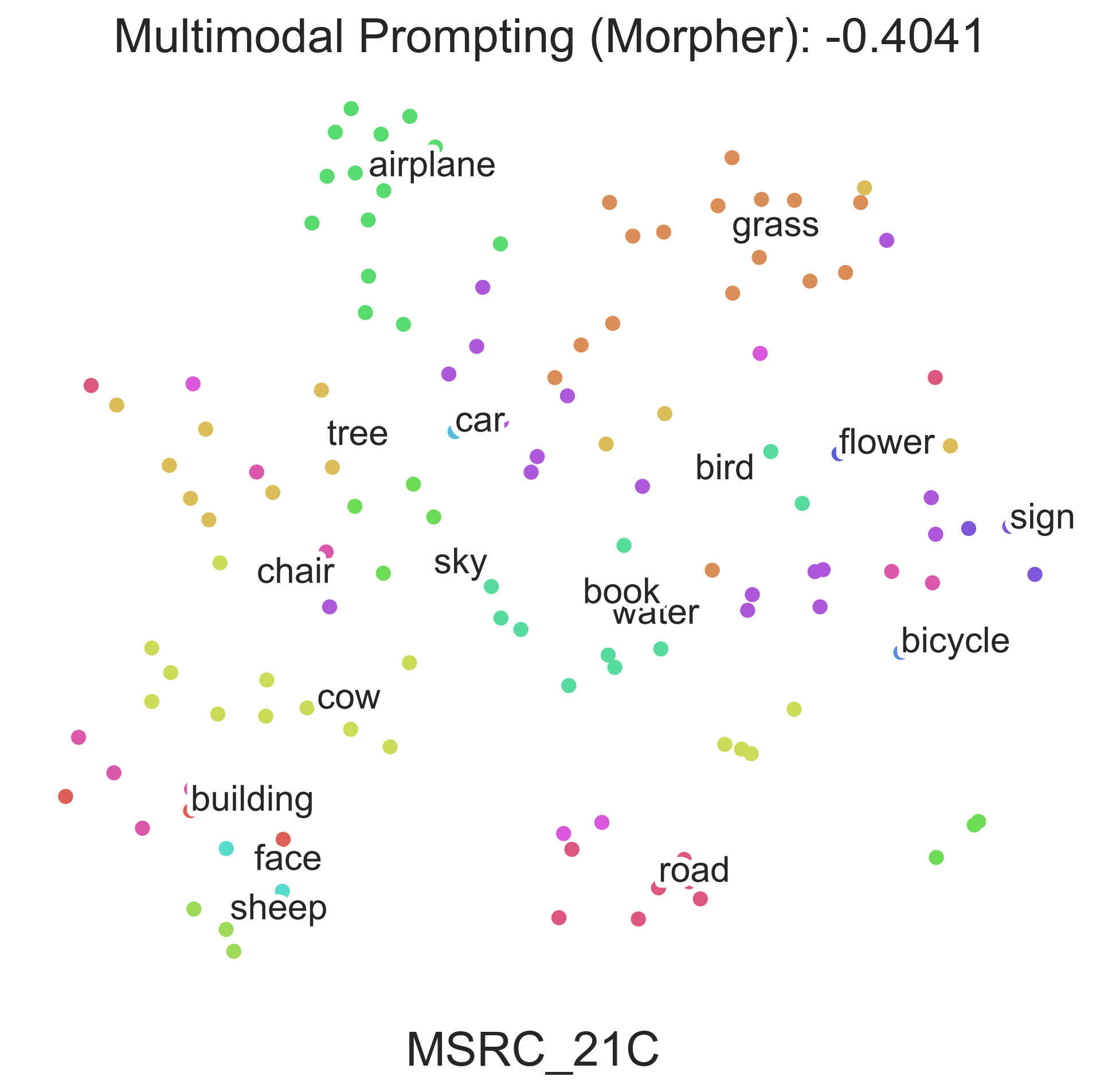}} \\
    \end{tabular}
    \caption{t-SNE embedding plots on CiteSeer (left) and MSRC\_21C (right). We calculate the silhouette score, a metric for cluster quality ($\uparrow$) ranged in $[-1, 1]$. It turns out that our Morpher leads to better adaptation.}
\end{figure*}

%% file: tables/few_shot_learning_full.tex
\begin{table*}
\centering
\resizebox{0.85\textwidth}{!}{%
\begin{tabular}{
cccccccccc
}
\toprule
\multirow{2}{*}{\makecell[c]{Training\\ schemes}}     & \multirow{2}{*}{GNN pretraining} & \multicolumn{2}{c}{MUTAG}       & \multicolumn{2}{c}{ENZYMES}    & \multicolumn{2}{c}{PROTEINS}     & \multicolumn{2}{c}{MSRC\_21C}  \\
                                      &                          & Acc & F1 & Acc & F1 & Acc & F1 & Acc & F1\\ \midrule
\multirow{3}{*}{Supervised}           
& N/A + GCN      & 66.00    & 66.67    & 16.67    & 8.68    & 65.89    & 60.77      & 38.85    & 35.32 \\
& N/A + GAT      & 66.00    & 65.69    & 16.45    & 4.65    & 64.75    & 64.08    & 41.14    & 39.86 \\
& N/A + GT       & 66.66    & 66.26    & 15.62    & 4.22    & 62.81    & 57.12    & 38.28    & 41.62 \\
\midrule
\multirow{6}{*}{\makecell[c]{Pre-train \\+\\  Fine-tune}} 
& GraphCL+GCN   & 70.00    & 70.23    & 17.91     & 11.82    & 65.89    & 61.23     & 40.00     & 43.89 \\
& GraphCL+GAT   & 70.00    & 69.73    & 17.91     & 10.46    & 65.16    & 63.92     & 44.57     & 45.74 \\
& GraphCL+GT    & 68.00    & 67.81    & 17.70     & 8.99    & 63.28    & 56.41     & 41.71    & 43.73 \\
& SimGRACE+GCN  & 66.67    & 67.27    & 17.29     & 8.78    & 66.82    & 64.70 & 40.57    & 43.84 \\
& SimGRACE+GAT  & 70.67    & 69.10    & 16.87     & 7.18    & 65.42    & 63.65     & 42.85    & 42.37 \\
& SimGRACE+GT   & 69.33    & 69.77    & 16.24     & 6.08    & 65.98    & 62.31     & 39.42    & 40.78 \\
\midrule
\multirow{6}{*}{\makecell{AIO\\ \citep{DBLP:conf/kdd/SunCLLG23}}}               
& GraphCL+GCN   & 64.67 & 39.27 & 17.50 & 4.97  & 61.35 & 44.93 & 3.59 & 10.09\\
& GraphCL+GAT   & 64.67 & 39.27 & 17.50 & 4.97  & 59.21 & 37.19 & 14.37 & 3.11\\
& GraphCL+GT    & 73.33 & 72.06 & 18.33 & 9.09 & 40.79 & 28.97 & 17.96 & 8.30\\
& SimGRACE+GCN  & 64.67 & 39.27 & 16.04 & 4.61  & 67.42 & 60.87 & 34.73 & 18.16\\
& SimGRACE+GAT  & 64.67 & 39.27 & 16.04 & 4.61  & 59.21 & 37.19 & 7.78 & 1.79\\
& SimGRACE+GT   & 36.00 & 27.26 & 17.50 & 8.15 & 50.56 & 49.34 & 32.34 & 15.13\\
\midrule
\multirow{6}{*}{\makecell{GPF-plus\\ \citep{DBLP:conf/nips/FangZYWC23}}}               
& GraphCL+GCN   & 68.67 & 67.27 & 16.88 & 15.48 & 64.75 & 61.45 & 47.42 & 29.02\\
& GraphCL+GAT   & 68.67 & 62.84 & 16.45 & 13.23 & 65.89 & 60.07 & 47.42 & 26.28\\
& GraphCL+GT    & 69.33 & 67.87 & 18.12 & 15.56 & 59.66 & 37.37 & 41.71 & 21.35\\
& SimGRACE+GCN  & 65.33 & 39.52 & 18.96 & 15.83 & 65.16 & 58.80 & 45.71 & 23.32\\
& SimGRACE+GAT  & 69.33 & 66.72 & 18.54 & 12.58 & 63.28 & 53.50 & 42.85 & 21.40\\
& SimGRACE+GT   & 70.00 & 67.31 & 17.91 & 14.69 & 64.83 & 52.97 & 34.13 & 20.13\\
\midrule
\multirow{6}{*}{\makecell{Gprompt\\ \citep{DBLP:conf/www/LiuY0023}}}               
& GraphCL+GCN   & 73.33 & 66.93 & 17.91 & 8.44  & 61.01 & 60.01 & 1.80 & 0.21\\
& GraphCL+GAT   & 64.67 & 62.63 & 17.08 & 14.18 & 50.56 & 50.55 & 1.80 & 0.22 \\
& GraphCL+GT    & 70.67 & 70.02 & 17.91 & 9.64  & 63.28 & 58.65 & 1.80 & 0.21\\
& SimGRACE+GCN  & 65.33 & 39.52 & 17.29 & 14.48 & 52.70 & 52.68 & 1.80 & 0.21\\
& SimGRACE+GAT  & 67.33 & 65.88 & 16.25 & 11.31 & 59.10 & 58.72 & 1.80 & 0.21\\
& SimGRACE+GT   & 73.33 & 67.84 & 16.87 & 13.54 & 64.75 & 62.37 & 1.80 & 0.223\\
\midrule
\multirow{6}{*}{\makecell[c]{Improved\\AIO (Ours)}}               
& GraphCL+GCN   & 77.33    & 77.74    & 18.13     & 11.98    & 65.89    & 65.97     & 42.85     & 45.91 \\
& GraphCL+GAT   & 74.67    & 75.51    & 18.33     & 11.26    & 65.76    & 66.05     & 46.85     & 51.39 \\
& GraphCL+GT    & 74.67    & 74.67    & 19.16     & 9.04    & 68.12    & 68.18     & 42.85     & 43.54 \\
& SimGRACE+GCN  & 68.00    & 69.01    & 17.91     & 9.02    & 66.82    & 66.40     & 44.57     & 49.24 \\
& SimGRACE+GAT  & 77.33    & 77.20    & 18.75     & 9.39    & 66.91    & 65.49     & 45.14     & 42.31 \\
& SimGRACE+GT   & 71.33    & 72.06    & 18.95     & 11.25    & 68.59    & 68.84     & 40.57     & 42.82\\
\midrule
\multirow{6}{*}{\makecell[c]{Morpher\\(Ours)}}               
& GraphCL+GCN   & 78.67    & 78.09    & 20.41     & 15.20    & 67.47    & 66.40     & 45.14     & 49.62 \\
& GraphCL+GAT   & 79.33    & 79.15    & 23.12     & 18.01    & 70.89    & 70.30     & 50.85     & 54.48 \\
& GraphCL+GT    & 76.00    & 76.51    & 19.58     & 13.28    & 73.53    & 72.48     & 45.71    & 48.41 \\
& SimGRACE+GCN  & 69.33    & 70.27    & 19.79     & 14.94    & 67.10    & 66.15     & 45.71    & 51.24 \\
& SimGRACE+GAT  & 78.00    & 77.65    & 20.21     & 16.27    & 68.12    & 67.26     & 45.71   & 51.13 \\
& SimGRACE+GT   & 74.00    & 74.84    & 19.16     & 14.29    & 71.76    & 71.75     & 44.00    & 48.16 \\
\midrule
\multicolumn{2}{c}{IMP of ImprovedAIO}   & 2.00 $\uparrow$ &	5.01 $\uparrow$ &	0.52 $\uparrow$ &	4.41 $\downarrow$  &	2.01 $\uparrow$ &	4.37 $\uparrow$ &	0.28 $\downarrow$ &	2.50  $\uparrow$ \\     
\midrule
\multicolumn{2}{c}{IMP of Morpher}     &                    4.00 $\uparrow$ & 6.73 $\uparrow$  & 2.36 $\uparrow$ & 	0.60 $\uparrow$  & 	4.81 $\uparrow$ & 	6.61 $\uparrow$ & 	2.66 $\uparrow$ & 	7.14  $\uparrow$\\  
\bottomrule
\end{tabular}%
}
\caption{Few-shot graph classification performance (\%). IMP (\%): the average improvement (absolute value) compared to the \textbf{best result} among all the baseline methods.}
\label{TB: main comparison full}
\end{table*}

%% file: acl_latex.bbl
\begin{thebibliography}{144}
\providecommand{\natexlab}[1]{#1}

\bibitem[{Ai et~al.(2025)Ai, Wei, Chen, Zeng, Zhao, Varatkar, Rouhani, Tang, Tong, and He}]{ai2025resmoe}
Mengting Ai, Tianxin Wei, Yifan Chen, Zhichen Zeng, Ritchie Zhao, Girish Varatkar, Bita~Darvish Rouhani, Xianfeng Tang, Hanghang Tong, and Jingrui He. 2025.
\newblock Resmoe: Space-efficient compression of mixture of experts llms via residual restoration.
\newblock \emph{arXiv preprint arXiv:2503.06881}.

\bibitem[{Ban et~al.(2024)Ban, Zou, Li, Qi, Fu, Kang, Tong, and He}]{DBLP:conf/nips/BanZLQFKTH24}
Yikun Ban, Jiaru Zou, Zihao Li, Yunzhe Qi, Dongqi Fu, Jian Kang, Hanghang Tong, and Jingrui He. 2024.
\newblock Pagerank bandits for link prediction.
\newblock In \emph{NeurIPS}.

\bibitem[{Bao et~al.()Bao, Zeng, Liu, Tong, and He}]{baomatcha}
Wenxuan Bao, Zhichen Zeng, Zhining Liu, Hanghang Tong, and Jingrui He.
\newblock Matcha: Mitigating graph structure shifts with test-time adaptation.
\newblock In \emph{The Thirteenth International Conference on Learning Representations}.

\bibitem[{Borgwardt et~al.(2005)Borgwardt, Ong, Sch{\"{o}}nauer, Vishwanathan, Smola, and Kriegel}]{DBLP:conf/ismb/BorgwardtOSVSK05}
Karsten~M. Borgwardt, Cheng~Soon Ong, Stefan Sch{\"{o}}nauer, S.~V.~N. Vishwanathan, Alexander~J. Smola, and Hans{-}Peter Kriegel. 2005.
\newblock \href {https://doi.org/10.1093/BIOINFORMATICS/BTI1007} {Protein function prediction via graph kernels}.
\newblock In \emph{Proceedings Thirteenth International Conference on Intelligent Systems for Molecular Biology 2005, Detroit, MI, USA, 25-29 June 2005}, pages 47--56.

\bibitem[{Bouritsas et~al.(2023)Bouritsas, Frasca, Zafeiriou, and Bronstein}]{DBLP:journals/pami/BouritsasFZB23}
Giorgos Bouritsas, Fabrizio Frasca, Stefanos Zafeiriou, and Michael~M. Bronstein. 2023.
\newblock \href {https://doi.org/10.1109/TPAMI.2022.3154319} {Improving graph neural network expressivity via subgraph isomorphism counting}.
\newblock \emph{{IEEE} Trans. Pattern Anal. Mach. Intell.}, 45(1):657--668.

\bibitem[{Chen et~al.(2023{\natexlab{a}})Chen, Liu, Liu, Li, Mao, and Sun}]{DBLP:journals/corr/abs-2310-14845}
Mouxiang Chen, Zemin Liu, Chenghao Liu, Jundong Li, Qiheng Mao, and Jianling Sun. 2023{\natexlab{a}}.
\newblock \href {https://doi.org/10.48550/ARXIV.2310.14845} {{ULTRA-DP:} unifying graph pre-training with multi-task graph dual prompt}.
\newblock \emph{CoRR}, abs/2310.14845.

\bibitem[{Chen et~al.(2023{\natexlab{b}})Chen, Gan, Wu, Hu, and Lin}]{DBLP:journals/corr/abs-2312-10073}
Zefeng Chen, Wensheng Gan, Jiayang Wu, Kaixia Hu, and Hong Lin. 2023{\natexlab{b}}.
\newblock \href {https://doi.org/10.48550/ARXIV.2312.10073} {Data scarcity in recommendation systems: {A} survey}.
\newblock \emph{CoRR}, abs/2312.10073.

\bibitem[{Chien et~al.(2021)Chien, Chang, Hsieh, Yu, Zhang, Milenkovic, and Dhillon}]{chien2021node}
Eli Chien, Wei-Cheng Chang, Cho-Jui Hsieh, Hsiang-Fu Yu, Jiong Zhang, Olgica Milenkovic, and Inderjit~S Dhillon. 2021.
\newblock Node feature extraction by self-supervised multi-scale neighborhood prediction.
\newblock \emph{arXiv preprint arXiv:2111.00064}.

\bibitem[{Clark et~al.(2020)Clark, Luong, Le, and Manning}]{DBLP:conf/iclr/ClarkLLM20}
Kevin Clark, Minh{-}Thang Luong, Quoc~V. Le, and Christopher~D. Manning. 2020.
\newblock \href {https://openreview.net/forum?id=r1xMH1BtvB} {{ELECTRA:} pre-training text encoders as discriminators rather than generators}.
\newblock In \emph{8th International Conference on Learning Representations, {ICLR} 2020, Addis Ababa, Ethiopia, April 26-30, 2020}. OpenReview.net.

\bibitem[{Duan et~al.(2023)Duan, Liu, Chua, Yan, Ooi, Xie, and He}]{duan2023simteg}
Keyu Duan, Qian Liu, Tat-Seng Chua, Shuicheng Yan, Wei~Tsang Ooi, Qizhe Xie, and Junxian He. 2023.
\newblock Simteg: A frustratingly simple approach improves textual graph learning.
\newblock \emph{arXiv preprint arXiv:2308.02565}.

\bibitem[{Dwivedi et~al.(2023)Dwivedi, Joshi, Luu, Laurent, Bengio, and Bresson}]{DBLP:journals/jmlr/DwivediJL0BB23}
Vijay~Prakash Dwivedi, Chaitanya~K. Joshi, Anh~Tuan Luu, Thomas Laurent, Yoshua Bengio, and Xavier Bresson. 2023.
\newblock \href {http://jmlr.org/papers/v24/22-0567.html} {Benchmarking graph neural networks}.
\newblock \emph{J. Mach. Learn. Res.}, 24:43:1--43:48.

\bibitem[{Fang et~al.(2023)Fang, Zhang, Yang, Wang, and Chen}]{DBLP:conf/nips/FangZYWC23}
Taoran Fang, Yunchao Zhang, Yang Yang, Chunping Wang, and Lei Chen. 2023.
\newblock \href {http://papers.nips.cc/paper\_files/paper/2023/hash/a4a1ee071ce0fe63b83bce507c9dc4d7-Abstract-Conference.html} {Universal prompt tuning for graph neural networks}.
\newblock In \emph{Advances in Neural Information Processing Systems 36: Annual Conference on Neural Information Processing Systems 2023, NeurIPS 2023, New Orleans, LA, USA, December 10 - 16, 2023}.

\bibitem[{Fey and Lenssen(2019)}]{DBLP:journals/corr/abs-1903-02428}
Matthias Fey and Jan~Eric Lenssen. 2019.
\newblock \href {https://arxiv.org/abs/1903.02428} {Fast graph representation learning with pytorch geometric}.
\newblock \emph{CoRR}, abs/1903.02428.

\bibitem[{Frasca et~al.(2022)Frasca, Bevilacqua, Bronstein, and Maron}]{DBLP:conf/nips/FrascaBBM22}
Fabrizio Frasca, Beatrice Bevilacqua, Michael~M. Bronstein, and Haggai Maron. 2022.
\newblock \href {http://papers.nips.cc/paper\_files/paper/2022/hash/cb2a4cc70db72ea779abd01107782c7b-Abstract-Conference.html} {Understanding and extending subgraph gnns by rethinking their symmetries}.
\newblock In \emph{Advances in Neural Information Processing Systems 35: Annual Conference on Neural Information Processing Systems 2022, NeurIPS 2022, New Orleans, LA, USA, November 28 - December 9, 2022}.

\bibitem[{Fu et~al.(2024{\natexlab{a}})Fu, Fang, Li, Tong, Torvik, and He}]{llmgraph}
Dongqi Fu, Liri Fang, Zihao Li, Hanghang Tong, Vetle~I. Torvik, and Jingrui He. 2024{\natexlab{a}}.
\newblock \href {https://doi.org/10.48550/ARXIV.2410.12126} {Parametric graph representations in the era of foundation models: {A} survey and position}.
\newblock \emph{CoRR}, abs/2410.12126.

\bibitem[{Fu et~al.(2022)Fu, Fang, Maciejewski, Torvik, and He}]{DBLP:conf/kdd/FuFMTH22}
Dongqi Fu, Liri Fang, Ross Maciejewski, Vetle~I. Torvik, and Jingrui He. 2022.
\newblock Meta-learned metrics over multi-evolution temporal graphs.
\newblock In \emph{{KDD}}.

\bibitem[{Fu et~al.(2024{\natexlab{b}})Fu, Hua, Xie, Fang, Zhang, Sancak, Wu, Malevich, He, and Long}]{DBLP:conf/iclr/FuHXFZSWMHL24}
Dongqi Fu, Zhigang Hua, Yan Xie, Jin Fang, Si~Zhang, Kaan Sancak, Hao Wu, Andrey Malevich, Jingrui He, and Bo~Long. 2024{\natexlab{b}}.
\newblock Vcr-graphormer: {A} mini-batch graph transformer via virtual connections.
\newblock In \emph{{ICLR}}.

\bibitem[{Fu et~al.(2025)Fu, Zhu, Liu, Zheng, Lin, Li, Fang, Tieu, Bhardwaj, Weldemariam, Tong, Hamann, and He}]{climatebenchm}
Dongqi Fu, Yada Zhu, Zhining Liu, Lecheng Zheng, Xiao Lin, Zihao Li, Liri Fang, Katherine Tieu, Onkar Bhardwaj, Kommy Weldemariam, Hanghang Tong, Hendrik~F. Hamann, and Jingrui He. 2025.
\newblock \href {https://doi.org/10.48550/ARXIV.2504.07394} {Climatebench-m: {A} multi-modal climate data benchmark with a simple generative method}.
\newblock \emph{CoRR}, abs/2504.07394.

\bibitem[{Fu et~al.(2024{\natexlab{c}})Fu, Zhu, Tong, Weldemariam, Bhardwaj, and He}]{DBLP:journals/corr/abs-2408-04254}
Dongqi Fu, Yada Zhu, Hanghang Tong, Kommy Weldemariam, Onkar Bhardwaj, and Jingrui He. 2024{\natexlab{c}}.
\newblock Generating fine-grained causality in climate time series data for forecasting and anomaly detection.
\newblock \emph{CoRR}.

\bibitem[{Ge et~al.(2023)Ge, Zhao, Liu, Cheng, Li, Wang, and Yin}]{DBLP:journals/corr/abs-2310-17394}
Qingqing Ge, Zeyuan Zhao, Yiding Liu, Anfeng Cheng, Xiang Li, Shuaiqiang Wang, and Dawei Yin. 2023.
\newblock \href {https://doi.org/10.48550/ARXIV.2310.17394} {Enhancing graph neural networks with structure-based prompt}.
\newblock \emph{CoRR}, abs/2310.17394.

\bibitem[{Giusti et~al.(2023)Giusti, Reu, Ceccarelli, Bodnar, and Li{\`{o}}}]{DBLP:journals/corr/abs-2306-03561}
Lorenzo Giusti, Teodora Reu, Francesco Ceccarelli, Cristian Bodnar, and Pietro Li{\`{o}}. 2023.
\newblock \href {https://doi.org/10.48550/ARXIV.2306.03561} {{CIN++:} enhancing topological message passing}.
\newblock \emph{CoRR}, abs/2306.03561.

\bibitem[{Guo et~al.(2022)Guo, Wang, Yan, Lou, Feng, Zhu, Chen, He, and Yu}]{guo2022learning}
Gaoyang Guo, Chaokun Wang, Bencheng Yan, Yunkai Lou, Hao Feng, Junchao Zhu, Jun Chen, Fei He, and Philip~S Yu. 2022.
\newblock Learning adaptive node embeddings across graphs.
\newblock \emph{IEEE Transactions on Knowledge and Data Engineering}, 35(6):6028--6042.

\bibitem[{Guzhov et~al.(2022)Guzhov, Raue, Hees, and Dengel}]{DBLP:conf/icassp/GuzhovRHD22}
Andrey Guzhov, Federico Raue, J{\"{o}}rn Hees, and Andreas Dengel. 2022.
\newblock \href {https://doi.org/10.1109/ICASSP43922.2022.9747631} {Audioclip: Extending clip to image, text and audio}.
\newblock In \emph{{IEEE} International Conference on Acoustics, Speech and Signal Processing, {ICASSP} 2022, Virtual and Singapore, 23-27 May 2022}, pages 976--980. {IEEE}.

\bibitem[{Hassani and Ahmadi(2020)}]{DBLP:conf/icml/HassaniA20}
Kaveh Hassani and Amir Hosein~Khas Ahmadi. 2020.
\newblock \href {http://proceedings.mlr.press/v119/hassani20a.html} {Contrastive multi-view representation learning on graphs}.
\newblock In \emph{Proceedings of the 37th International Conference on Machine Learning, {ICML} 2020, 13-18 July 2020, Virtual Event}, volume 119 of \emph{Proceedings of Machine Learning Research}, pages 4116--4126. {PMLR}.

\bibitem[{He et~al.(2025)He, Fu, Tong, Maciejewski, and He}]{DBLP:conf/iclr/0003FTMH25}
Xinyu He, Dongqi Fu, Hanghang Tong, Ross Maciejewski, and Jingrui He. 2025.
\newblock Temporal heterogeneous graph generation with privacy, utility, and efficiency.
\newblock In \emph{{ICLR}}.

\bibitem[{Hess et~al.(2024)Hess, Tonderski, Petersson, {\AA}str{\"{o}}m, and Svensson}]{DBLP:conf/wacv/HessTPAS24}
Georg Hess, Adam Tonderski, Christoffer Petersson, Kalle {\AA}str{\"{o}}m, and Lennart Svensson. 2024.
\newblock \href {https://doi.org/10.1109/WACV57701.2024.00727} {Lidarclip or: How {I} learned to talk to point clouds}.
\newblock In \emph{{IEEE/CVF} Winter Conference on Applications of Computer Vision, {WACV} 2024, Waikoloa, HI, USA, January 3-8, 2024}, pages 7423--7432. {IEEE}.

\bibitem[{Hou et~al.(2022)Hou, Liu, Cen, Dong, Yang, Wang, and Tang}]{DBLP:conf/kdd/HouLCDYW022}
Zhenyu Hou, Xiao Liu, Yukuo Cen, Yuxiao Dong, Hongxia Yang, Chunjie Wang, and Jie Tang. 2022.
\newblock \href {https://doi.org/10.1145/3534678.3539321} {Graphmae: Self-supervised masked graph autoencoders}.
\newblock In \emph{{KDD} '22: The 28th {ACM} {SIGKDD} Conference on Knowledge Discovery and Data Mining, Washington, DC, USA, August 14 - 18, 2022}, pages 594--604. {ACM}.

\bibitem[{Houlsby et~al.(2019)Houlsby, Giurgiu, Jastrzebski, Morrone, de~Laroussilhe, Gesmundo, Attariyan, and Gelly}]{DBLP:conf/icml/HoulsbyGJMLGAG19}
Neil Houlsby, Andrei Giurgiu, Stanislaw Jastrzebski, Bruna Morrone, Quentin de~Laroussilhe, Andrea Gesmundo, Mona Attariyan, and Sylvain Gelly. 2019.
\newblock \href {http://proceedings.mlr.press/v97/houlsby19a.html} {Parameter-efficient transfer learning for {NLP}}.
\newblock In \emph{Proceedings of the 36th International Conference on Machine Learning, {ICML} 2019, 9-15 June 2019, Long Beach, California, {USA}}, volume~97 of \emph{Proceedings of Machine Learning Research}, pages 2790--2799. {PMLR}.

\bibitem[{Hu et~al.(2020{\natexlab{a}})Hu, Liu, Gomes, Zitnik, Liang, Pande, and Leskovec}]{DBLP:conf/iclr/HuLGZLPL20}
Weihua Hu, Bowen Liu, Joseph Gomes, Marinka Zitnik, Percy Liang, Vijay~S. Pande, and Jure Leskovec. 2020{\natexlab{a}}.
\newblock \href {https://openreview.net/forum?id=HJlWWJSFDH} {Strategies for pre-training graph neural networks}.
\newblock In \emph{8th International Conference on Learning Representations, {ICLR} 2020, Addis Ababa, Ethiopia, April 26-30, 2020}. OpenReview.net.

\bibitem[{Hu et~al.(2020{\natexlab{b}})Hu, Dong, Wang, Chang, and Sun}]{DBLP:conf/kdd/HuZiniu20}
Ziniu Hu, Yuxiao Dong, Kuansan Wang, Kai{-}Wei Chang, and Yizhou Sun. 2020{\natexlab{b}}.
\newblock {GPT-GNN:} generative pre-training of graph neural networks.
\newblock In \emph{{KDD} '20: The 26th {ACM} {SIGKDD} Conference on Knowledge Discovery and Data Mining, Virtual Event, CA, USA, August 23-27, 2020}, pages 1857--1867. {ACM}.

\bibitem[{Huang et~al.(2023)Huang, Ren, Chen, Krzmanc, Zeng, Liang, and Leskovec}]{DBLP:conf/nips/HuangRCK0LL23}
Qian Huang, Hongyu Ren, Peng Chen, Gregor Krzmanc, Daniel Zeng, Percy Liang, and Jure Leskovec. 2023.
\newblock \href {http://papers.nips.cc/paper\_files/paper/2023/hash/34dce0dc3121951dd0399ba02c0f0d06-Abstract-Conference.html} {{PRODIGY:} enabling in-context learning over graphs}.
\newblock In \emph{Advances in Neural Information Processing Systems 36: Annual Conference on Neural Information Processing Systems 2023, NeurIPS 2023, New Orleans, LA, USA, December 10 - 16, 2023}.

\bibitem[{Jia et~al.(2021)Jia, Yang, Xia, Chen, Parekh, Pham, Le, Sung, Li, and Duerig}]{DBLP:conf/icml/JiaYXCPPLSLD21}
Chao Jia, Yinfei Yang, Ye~Xia, Yi{-}Ting Chen, Zarana Parekh, Hieu Pham, Quoc~V. Le, Yun{-}Hsuan Sung, Zhen Li, and Tom Duerig. 2021.
\newblock \href {http://proceedings.mlr.press/v139/jia21b.html} {Scaling up visual and vision-language representation learning with noisy text supervision}.
\newblock In \emph{Proceedings of the 38th International Conference on Machine Learning, {ICML} 2021, 18-24 July 2021, Virtual Event}, volume 139 of \emph{Proceedings of Machine Learning Research}, pages 4904--4916. {PMLR}.

\bibitem[{Jin et~al.(2023{\natexlab{a}})Jin, Liu, Han, Jiang, Ji, and Han}]{jin2023large}
Bowen Jin, Gang Liu, Chi Han, Meng Jiang, Heng Ji, and Jiawei Han. 2023{\natexlab{a}}.
\newblock Large language models on graphs: A comprehensive survey.
\newblock \emph{arXiv preprint arXiv:2312.02783}.

\bibitem[{Jin et~al.(2023{\natexlab{b}})Jin, Zhang, Zhang, Meng, Zhang, Zhu, and Han}]{DBLP:conf/acl/JinZZ000023}
Bowen Jin, Wentao Zhang, Yu~Zhang, Yu~Meng, Xinyang Zhang, Qi~Zhu, and Jiawei Han. 2023{\natexlab{b}}.
\newblock \href {https://doi.org/10.18653/V1/2023.ACL-LONG.387} {Patton: Language model pretraining on text-rich networks}.
\newblock In \emph{Proceedings of the 61st Annual Meeting of the Association for Computational Linguistics (Volume 1: Long Papers), {ACL} 2023, Toronto, Canada, July 9-14, 2023}, pages 7005--7020. Association for Computational Linguistics.

\bibitem[{Jin et~al.(2023{\natexlab{c}})Jin, Zhang, Zhang, Meng, Zhang, Zhu, and Han}]{jin2023patton}
Bowen Jin, Wentao Zhang, Yu~Zhang, Yu~Meng, Xinyang Zhang, Qi~Zhu, and Jiawei Han. 2023{\natexlab{c}}.
\newblock Patton: Language model pretraining on text-rich networks.
\newblock \emph{arXiv preprint arXiv:2305.12268}.

\bibitem[{Jin et~al.(2023{\natexlab{d}})Jin, Zhang, Zhu, and Han}]{jin2023heterformer}
Bowen Jin, Yu~Zhang, Qi~Zhu, and Jiawei Han. 2023{\natexlab{d}}.
\newblock Heterformer: Transformer-based deep node representation learning on heterogeneous text-rich networks.
\newblock In \emph{Proceedings of the 29th ACM SIGKDD Conference on Knowledge Discovery and Data Mining}, pages 1020--1031.

\bibitem[{Jin et~al.(2024)Jin, Yang, and Xu}]{jin2024scam}
Yihong Jin, Ze~Yang, and Xinhe Xu. 2024.
\newblock Scam detection for ethereum smart contracts: Leveraging graph representation learning for secure blockchain.
\newblock \emph{arXiv preprint arXiv:2412.12370}.

\bibitem[{Jing et~al.(2021)Jing, Park, and Tong}]{jing2021hdmi}
Baoyu Jing, Chanyoung Park, and Hanghang Tong. 2021.
\newblock {HDMI:} high-order deep multiplex infomax.
\newblock In \emph{{WWW} '21: The Web Conference 2021, Virtual Event / Ljubljana, Slovenia, April 19-23, 2021}, pages 2414--2424. {ACM} / {IW3C2}.

\bibitem[{Jing et~al.(2024)Jing, Yan, Ding, Park, Zhu, Liu, and Tong}]{jing2024sterling}
Baoyu Jing, Yuchen Yan, Kaize Ding, Chanyoung Park, Yada Zhu, Huan Liu, and Hanghang Tong. 2024.
\newblock Sterling: Synergistic representation learning on bipartite graphs.
\newblock In \emph{Proceedings of the AAAI Conference on Artificial Intelligence}, volume~38, pages 12976--12984.

\bibitem[{Jing et~al.(2022)Jing, Yan, Zhu, and Tong}]{jing2022coin}
Baoyu Jing, Yuchen Yan, Yada Zhu, and Hanghang Tong. 2022.
\newblock Coin: Co-cluster infomax for bipartite graphs.
\newblock \emph{arXiv preprint arXiv:2206.00006}.

\bibitem[{Kaplan et~al.(2020)Kaplan, McCandlish, Henighan, Brown, Chess, Child, Gray, Radford, Wu, and Amodei}]{DBLP:journals/corr/abs-2001-08361}
Jared Kaplan, Sam McCandlish, Tom Henighan, Tom~B. Brown, Benjamin Chess, Rewon Child, Scott Gray, Alec Radford, Jeffrey Wu, and Dario Amodei. 2020.
\newblock \href {https://arxiv.org/abs/2001.08361} {Scaling laws for neural language models}.
\newblock \emph{CoRR}, abs/2001.08361.

\bibitem[{Khattak et~al.(2023)Khattak, Rasheed, Maaz, Khan, and Khan}]{DBLP:conf/cvpr/KhattakR0KK23}
Muhammad~Uzair Khattak, Hanoona~Abdul Rasheed, Muhammad Maaz, Salman~H. Khan, and Fahad~Shahbaz Khan. 2023.
\newblock \href {https://doi.org/10.1109/CVPR52729.2023.01832} {Maple: Multi-modal prompt learning}.
\newblock In \emph{{IEEE/CVF} Conference on Computer Vision and Pattern Recognition, {CVPR} 2023, Vancouver, BC, Canada, June 17-24, 2023}, pages 19113--19122. {IEEE}.

\bibitem[{Khosla et~al.(2020)Khosla, Teterwak, Wang, Sarna, Tian, Isola, Maschinot, Liu, and Krishnan}]{DBLP:conf/nips/KhoslaTWSTIMLK20}
Prannay Khosla, Piotr Teterwak, Chen Wang, Aaron Sarna, Yonglong Tian, Phillip Isola, Aaron Maschinot, Ce~Liu, and Dilip Krishnan. 2020.
\newblock Supervised contrastive learning.
\newblock In \emph{Advances in Neural Information Processing Systems 33: Annual Conference on Neural Information Processing Systems 2020, NeurIPS 2020, December 6-12, 2020, virtual}.

\bibitem[{Kipf and Welling(2016)}]{DBLP:journals/corr/KipfW16}
Thomas~N. Kipf and Max Welling. 2016.
\newblock \href {https://arxiv.org/abs/1609.02907} {Semi-supervised classification with graph convolutional networks}.
\newblock \emph{CoRR}, abs/1609.02907.

\bibitem[{Kong et~al.(2023)Kong, Cui, Sun, Zhuang, Prakash, and Zhang}]{kong2023autoregressive}
Lingkai Kong, Jiaming Cui, Haotian Sun, Yuchen Zhuang, B~Aditya Prakash, and Chao Zhang. 2023.
\newblock Autoregressive diffusion model for graph generation.
\newblock In \emph{International conference on machine learning}, pages 17391--17408. PMLR.

\bibitem[{Kreuzer et~al.(2021)Kreuzer, Beaini, Hamilton, L{\'{e}}tourneau, and Tossou}]{DBLP:conf/nips/KreuzerBHLT21}
Devin Kreuzer, Dominique Beaini, William~L. Hamilton, Vincent L{\'{e}}tourneau, and Prudencio Tossou. 2021.
\newblock \href {https://proceedings.neurips.cc/paper/2021/hash/b4fd1d2cb085390fbbadae65e07876a7-Abstract.html} {Rethinking graph transformers with spectral attention}.
\newblock In \emph{Advances in Neural Information Processing Systems 34: Annual Conference on Neural Information Processing Systems 2021, NeurIPS 2021, December 6-14, 2021, virtual}, pages 21618--21629.

\bibitem[{Lan et~al.(2024)Lan, Zheng, Ming, and Kilicoglu}]{DBLP:conf/emnlp/LanZMK24}
Mengfei Lan, Lecheng Zheng, Shufan Ming, and Halil Kilicoglu. 2024.
\newblock Multi-label sequential sentence classification via large language model.
\newblock In \emph{Findings of the Association for Computational Linguistics: {EMNLP} 2024, Miami, Florida, USA, November 12-16, 2024}, pages 16086--16104. Association for Computational Linguistics.

\bibitem[{Lester et~al.(2021)Lester, Al{-}Rfou, and Constant}]{DBLP:conf/emnlp/LesterAC21}
Brian Lester, Rami Al{-}Rfou, and Noah Constant. 2021.
\newblock \href {https://doi.org/10.18653/V1/2021.EMNLP-MAIN.243} {The power of scale for parameter-efficient prompt tuning}.
\newblock In \emph{Proceedings of the 2021 Conference on Empirical Methods in Natural Language Processing, {EMNLP} 2021, Virtual Event / Punta Cana, Dominican Republic, 7-11 November, 2021}, pages 3045--3059. Association for Computational Linguistics.

\bibitem[{Li et~al.(2022)Li, Li, Xiong, and Hoi}]{DBLP:conf/icml/0001LXH22}
Junnan Li, Dongxu Li, Caiming Xiong, and Steven C.~H. Hoi. 2022.
\newblock \href {https://proceedings.mlr.press/v162/li22n.html} {{BLIP:} bootstrapping language-image pre-training for unified vision-language understanding and generation}.
\newblock In \emph{International Conference on Machine Learning, {ICML} 2022, 17-23 July 2022, Baltimore, Maryland, {USA}}, volume 162 of \emph{Proceedings of Machine Learning Research}, pages 12888--12900. {PMLR}.

\bibitem[{Li and Liang(2021)}]{DBLP:conf/acl/LiL20}
Xiang~Lisa Li and Percy Liang. 2021.
\newblock \href {https://doi.org/10.18653/V1/2021.ACL-LONG.353} {Prefix-tuning: Optimizing continuous prompts for generation}.
\newblock In \emph{Proceedings of the 59th Annual Meeting of the Association for Computational Linguistics and the 11th International Joint Conference on Natural Language Processing, {ACL/IJCNLP} 2021, (Volume 1: Long Papers), Virtual Event, August 1-6, 2021}, pages 4582--4597. Association for Computational Linguistics.

\bibitem[{Li et~al.(2023{\natexlab{a}})Li, Ding, and Lee}]{DBLP:conf/emnlp/0001DL23}
Yichuan Li, Kaize Ding, and Kyumin Lee. 2023{\natexlab{a}}.
\newblock \href {https://doi.org/10.18653/V1/2023.FINDINGS-EMNLP.181} {{GRENADE:} graph-centric language model for self-supervised representation learning on text-attributed graphs}.
\newblock In \emph{Findings of the Association for Computational Linguistics: {EMNLP} 2023, Singapore, December 6-10, 2023}, pages 2745--2757. Association for Computational Linguistics.

\bibitem[{Li et~al.(2024{\natexlab{a}})Li, Ao, and He}]{DBLP:conf/sigir/LiAH24}
Zihao Li, Yuyi Ao, and Jingrui He. 2024{\natexlab{a}}.
\newblock \href {https://doi.org/10.1145/3626772.3657910} {Sphere: Expressive and interpretable knowledge graph embedding for set retrieval}.
\newblock In \emph{Proceedings of the 47th International {ACM} {SIGIR} Conference on Research and Development in Information Retrieval, {SIGIR} 2024, Washington DC, USA, July 14-18, 2024}, pages 2629--2634. {ACM}.

\bibitem[{Li et~al.(2025{\natexlab{a}})Li, Fu, Ai, and He}]{DBLP:conf/kdd/LiFAH25}
Zihao Li, Dongqi Fu, Mengting Ai, and Jingrui He. 2025{\natexlab{a}}.
\newblock \href {https://doi.org/10.1145/3690624.3709213} {Apex\({}^{\mbox{2}}\): Adaptive and extreme summarization for personalized knowledge graphs}.
\newblock In \emph{Proceedings of the 31st {ACM} {SIGKDD} Conference on Knowledge Discovery and Data Mining, V.1, {KDD} 2025, Toronto, ON, Canada, August 3-7, 2025}, pages 741--752. {ACM}.

\bibitem[{Li et~al.(2023{\natexlab{b}})Li, Fu, and He}]{eveppr}
Zihao Li, Dongqi Fu, and Jingrui He. 2023{\natexlab{b}}.
\newblock \href {https://doi.org/10.1145/3543507.3583474} {Everything evolves in personalized pagerank}.
\newblock In \emph{Proceedings of the {ACM} Web Conference 2023, {WWW} 2023, Austin, TX, USA, 30 April 2023 - 4 May 2023}, pages 3342--3352. {ACM}.

\bibitem[{Li et~al.(2025{\natexlab{b}})Li, Lin, Liu, Zou, Wu, Zheng, Fu, Zhu, Hamann, Tong et~al.}]{li2025language}
Zihao Li, Xiao Lin, Zhining Liu, Jiaru Zou, Ziwei Wu, Lecheng Zheng, Dongqi Fu, Yada Zhu, Hendrik Hamann, Hanghang Tong, et~al. 2025{\natexlab{b}}.
\newblock Language in the flow of time: Time-series-paired texts weaved into a unified temporal narrative.
\newblock \emph{arXiv preprint arXiv:2502.08942}.

\bibitem[{Li et~al.(2024{\natexlab{b}})Li, He, Yang, Ryu, Kim, and Madduri}]{DBLP:journals/corr/abs-2409-11585}
Zilinghan Li, Shilan He, Ze~Yang, Minseok Ryu, Kibaek Kim, and Ravi~K. Madduri. 2024{\natexlab{b}}.
\newblock \href {https://doi.org/10.48550/ARXIV.2409.11585} {Advances in {APPFL:} {A} comprehensive and extensible federated learning framework}.
\newblock \emph{CoRR}, abs/2409.11585.

\bibitem[{Lin et~al.(2024)Lin, Liu, Fu, Qiu, and Tong}]{DBLP:conf/nips/Lin0FQT24}
Xiao Lin, Zhining Liu, Dongqi Fu, Ruizhong Qiu, and Hanghang Tong. 2024.
\newblock Backtime: Backdoor attacks on multivariate time series forecasting.
\newblock In \emph{NeurIPS 2024}.

\bibitem[{Lin et~al.(2025{\natexlab{a}})Lin, Liu, Yang, Li, Qiu, Wang, Liu, Li, Keswani, Pardeshi et~al.}]{lin2025moralise}
Xiao Lin, Zhining Liu, Ze~Yang, Gaotang Li, Ruizhong Qiu, Shuke Wang, Hui Liu, Haotian Li, Sumit Keswani, Vishwa Pardeshi, et~al. 2025{\natexlab{a}}.
\newblock Moralise: A structured benchmark for moral alignment in visual language models.
\newblock \emph{arXiv preprint arXiv:2505.14728}.

\bibitem[{Lin et~al.(2025{\natexlab{b}})Lin, Zeng, Wei, Liu, Tong et~al.}]{lin2025cats}
Xiao Lin, Zhichen Zeng, Tianxin Wei, Zhining Liu, Hanghang Tong, et~al. 2025{\natexlab{b}}.
\newblock Cats: Mitigating correlation shift for multivariate time series classification.
\newblock \emph{arXiv preprint arXiv:2504.04283}.

\bibitem[{Liu et~al.(2023{\natexlab{a}})Liu, Yang, Lu, Chen, Li, Zhang, Bai, Fang, Sun, Yu, and Shi}]{DBLP:journals/corr/abs-2310-11829}
Jiawei Liu, Cheng Yang, Zhiyuan Lu, Junze Chen, Yibo Li, Mengmei Zhang, Ting Bai, Yuan Fang, Lichao Sun, Philip~S. Yu, and Chuan Shi. 2023{\natexlab{a}}.
\newblock \href {https://doi.org/10.48550/ARXIV.2310.11829} {Towards graph foundation models: {A} survey and beyond}.
\newblock \emph{CoRR}, abs/2310.11829.

\bibitem[{Liu et~al.(2023{\natexlab{b}})Liu, Li, Feng, Tran, Zhao, Qiu, and Li}]{liu2023structural}
Shikun Liu, Tianchun Li, Yongbin Feng, Nhan Tran, Han Zhao, Qiang Qiu, and Pan Li. 2023{\natexlab{b}}.
\newblock Structural re-weighting improves graph domain adaptation.
\newblock In \emph{International Conference on Machine Learning}, pages 21778--21793. PMLR.

\bibitem[{Liu et~al.(2024{\natexlab{a}})Liu, Zeng, Liu, Yuan, Song, Hang, Liu, Yang, Kim, Chen et~al.}]{liu2024collaborative}
Xiaolong Liu, Zhichen Zeng, Xiaoyi Liu, Siyang Yuan, Weinan Song, Mengyue Hang, Yiqun Liu, Chaofei Yang, Donghyun Kim, Wen-Yen Chen, et~al. 2024{\natexlab{a}}.
\newblock A collaborative ensemble framework for ctr prediction.
\newblock \emph{arXiv preprint arXiv:2411.13700}.

\bibitem[{Liu et~al.(2019)Liu, Ott, Goyal, Du, Joshi, Chen, Levy, Lewis, Zettlemoyer, and Stoyanov}]{DBLP:journals/corr/abs-1907-11692}
Yinhan Liu, Myle Ott, Naman Goyal, Jingfei Du, Mandar Joshi, Danqi Chen, Omer Levy, Mike Lewis, Luke Zettlemoyer, and Veselin Stoyanov. 2019.
\newblock \href {https://arxiv.org/abs/1907.11692} {Roberta: {A} robustly optimized {BERT} pretraining approach}.
\newblock \emph{CoRR}, abs/1907.11692.

\bibitem[{Liu et~al.(2023{\natexlab{c}})Liu, Jin, Pan, Zhou, Zheng, Xia, and Yu}]{DBLP:journals/tkde/LiuJPZZXY23}
Yixin Liu, Ming Jin, Shirui Pan, Chuan Zhou, Yu~Zheng, Feng Xia, and Philip~S. Yu. 2023{\natexlab{c}}.
\newblock \href {https://doi.org/10.1109/TKDE.2022.3172903} {Graph self-supervised learning: {A} survey}.
\newblock \emph{{IEEE} Trans. Knowl. Data Eng.}, 35(6):5879--5900.

\bibitem[{Liu et~al.(2023{\natexlab{d}})Liu, Yu, Fang, and Zhang}]{DBLP:conf/www/LiuY0023}
Zemin Liu, Xingtong Yu, Yuan Fang, and Xinming Zhang. 2023{\natexlab{d}}.
\newblock \href {https://doi.org/10.1145/3543507.3583386} {Graphprompt: Unifying pre-training and downstream tasks for graph neural networks}.
\newblock In \emph{Proceedings of the {ACM} Web Conference 2023, {WWW} 2023, Austin, TX, USA, 30 April 2023 - 4 May 2023}, pages 417--428. {ACM}.

\bibitem[{Liu et~al.(2025)Liu, Amjad, Adkathimar, Wei, and Tong}]{liu2025selfelicit}
Zhining Liu, Rana~Ali Amjad, Ravinarayana Adkathimar, Tianxin Wei, and Hanghang Tong. 2025.
\newblock Selfelicit: Your language model secretly knows where is the relevant evidence.
\newblock \emph{arXiv preprint arXiv:2502.08767}.

\bibitem[{Liu et~al.(2020)Liu, Cao, Gao, Bian, Chen, Chang, and Liu}]{liu2020self}
Zhining Liu, Wei Cao, Zhifeng Gao, Jiang Bian, Hechang Chen, Yi~Chang, and Tie-Yan Liu. 2020.
\newblock Self-paced ensemble for highly imbalanced massive data classification.
\newblock In \emph{2020 IEEE 36th international conference on data engineering (ICDE)}, pages 841--852. IEEE.

\bibitem[{Liu et~al.(2024{\natexlab{b}})Liu, Qiu, Zeng, Yoo, Zhou, Xu, Zhu, Weldemariam, He, and Tong}]{liu2024class}
Zhining Liu, Ruizhong Qiu, Zhichen Zeng, Hyunsik Yoo, David Zhou, Zhe Xu, Yada Zhu, Kommy Weldemariam, Jingrui He, and Hanghang Tong. 2024{\natexlab{b}}.
\newblock Class-imbalanced graph learning without class rebalancing.
\newblock In \emph{Forty-first International Conference on Machine Learning}.

\bibitem[{Liu et~al.(2023{\natexlab{e}})Liu, Li, Luo, Fei, Cao, Kawaguchi, Wang, and Chua}]{DBLP:conf/emnlp/LiuLL00K0C23}
Zhiyuan Liu, Sihang Li, Yanchen Luo, Hao Fei, Yixin Cao, Kenji Kawaguchi, Xiang Wang, and Tat{-}Seng Chua. 2023{\natexlab{e}}.
\newblock \href {https://doi.org/10.18653/V1/2023.EMNLP-MAIN.966} {Molca: Molecular graph-language modeling with cross-modal projector and uni-modal adapter}.
\newblock In \emph{Proceedings of the 2023 Conference on Empirical Methods in Natural Language Processing, {EMNLP} 2023, Singapore, December 6-10, 2023}, pages 15623--15638. Association for Computational Linguistics.

\bibitem[{Lu et~al.(2021)Lu, Jiang, Fang, and Shi}]{DBLP:conf/aaai/LuJ0S21}
Yuanfu Lu, Xunqiang Jiang, Yuan Fang, and Chuan Shi. 2021.
\newblock Learning to pre-train graph neural networks.
\newblock In \emph{Thirty-Fifth {AAAI} Conference on Artificial Intelligence, {AAAI} 2021, Thirty-Third Conference on Innovative Applications of Artificial Intelligence, {IAAI} 2021, The Eleventh Symposium on Educational Advances in Artificial Intelligence, {EAAI} 2021, Virtual Event, February 2-9, 2021}, pages 4276--4284. {AAAI} Press.

\bibitem[{Luo et~al.(2023)Luo, Yang, Hong, Liu, and Nie}]{DBLP:journals/corr/abs-2307-09484}
Yizhen Luo, Kai Yang, Massimo Hong, Xing~Yi Liu, and Zaiqing Nie. 2023.
\newblock \href {https://doi.org/10.48550/ARXIV.2307.09484} {Molfm: {A} multimodal molecular foundation model}.
\newblock \emph{CoRR}, abs/2307.09484.

\bibitem[{Madani et~al.(2020)Madani, McCann, Naik, Keskar, Anand, Eguchi, Huang, and Socher}]{DBLP:journals/corr/abs-2004-03497}
Ali Madani, Bryan McCann, Nikhil Naik, Nitish~Shirish Keskar, Namrata Anand, Raphael~R. Eguchi, Po{-}Ssu Huang, and Richard Socher. 2020.
\newblock \href {https://arxiv.org/abs/2004.03497} {Progen: Language modeling for protein generation}.
\newblock \emph{CoRR}, abs/2004.03497.

\bibitem[{Manchanda et~al.(2023)Manchanda, Gupta, Ranu, and Bedathur}]{DBLP:conf/log/ManchandaGRB23}
Sahil Manchanda, Shubham Gupta, Sayan Ranu, and Srikanta~J. Bedathur. 2023.
\newblock \href {https://proceedings.mlr.press/v231/manchanda24a.html} {Generative modeling of labeled graphs under data scarcity}.
\newblock In \emph{Learning on Graphs Conference, 27-30 November 2023, Virtual Event}, volume 231 of \emph{Proceedings of Machine Learning Research}, page~32. {PMLR}.

\bibitem[{Morris et~al.(2020)Morris, Kriege, Bause, Kersting, Mutzel, and Neumann}]{DBLP:journals/corr/abs-2007-08663}
Christopher Morris, Nils~M. Kriege, Franka Bause, Kristian Kersting, Petra Mutzel, and Marion Neumann. 2020.
\newblock \href {https://arxiv.org/abs/2007.08663} {Tudataset: {A} collection of benchmark datasets for learning with graphs}.
\newblock \emph{CoRR}, abs/2007.08663.

\bibitem[{Neumann et~al.(2016)Neumann, Garnett, Bauckhage, and Kersting}]{DBLP:journals/ml/NeumannGBK16}
Marion Neumann, Roman Garnett, Christian Bauckhage, and Kristian Kersting. 2016.
\newblock \href {https://doi.org/10.1007/S10994-015-5517-9} {Propagation kernels: efficient graph kernels from propagated information}.
\newblock \emph{Mach. Learn.}, 102(2):209--245.

\bibitem[{Nguyen et~al.(2023)Nguyen, Poli, Faizi, Thomas, Wornow, Birch{-}Sykes, Massaroli, Patel, Rabideau, Bengio, Ermon, R{\'{e}}, and Baccus}]{DBLP:conf/nips/NguyenPFTWBMPRB23}
Eric Nguyen, Michael Poli, Marjan Faizi, Armin~W. Thomas, Michael Wornow, Callum Birch{-}Sykes, Stefano Massaroli, Aman Patel, Clayton~M. Rabideau, Yoshua Bengio, Stefano Ermon, Christopher R{\'{e}}, and Stephen Baccus. 2023.
\newblock \href {http://papers.nips.cc/paper\_files/paper/2023/hash/86ab6927ee4ae9bde4247793c46797c7-Abstract-Conference.html} {Hyenadna: Long-range genomic sequence modeling at single nucleotide resolution}.
\newblock In \emph{Advances in Neural Information Processing Systems 36: Annual Conference on Neural Information Processing Systems 2023, NeurIPS 2023, New Orleans, LA, USA, December 10 - 16, 2023}.

\bibitem[{Park et~al.(2022)Park, Bae, Kim, Kim, and Choi}]{DBLP:conf/chil/ParkBKKC22}
Sungjin Park, Seongsu Bae, Jiho Kim, Tackeun Kim, and Edward Choi. 2022.
\newblock \href {https://proceedings.mlr.press/v174/park22a.html} {Graph-text multi-modal pre-training for medical representation learning}.
\newblock In \emph{Conference on Health, Inference, and Learning, {CHIL} 2022, 7-8 April 2022, Virtual Event}, volume 174 of \emph{Proceedings of Machine Learning Research}, pages 261--281. {PMLR}.

\bibitem[{Qiu et~al.(2023)Qiu, Li, Sun, Peng, Shi, Zhang, Dong, Lam, Lo, Xiao, Yuan, Wang, Xu, and Lo}]{DBLP:journals/titb/QiuLSPSZDLLXYWXL23}
Jianing Qiu, Lin Li, Jiankai Sun, Jiachuan Peng, Peilun Shi, Ruiyang Zhang, Yinzhao Dong, Kyle Lam, Frank~P.{-}W. Lo, Bo~Xiao, Wu~Yuan, Ningli Wang, Dong Xu, and Benny P.~L. Lo. 2023.
\newblock \href {https://doi.org/10.1109/JBHI.2023.3316750} {Large {AI} models in health informatics: Applications, challenges, and the future}.
\newblock \emph{{IEEE} J. Biomed. Health Informatics}, 27(12):6074--6087.

\bibitem[{Radford et~al.(2021)Radford, Kim, Hallacy, Ramesh, Goh, Agarwal, Sastry, Askell, Mishkin, Clark, Krueger, and Sutskever}]{DBLP:conf/icml/RadfordKHRGASAM21}
Alec Radford, Jong~Wook Kim, Chris Hallacy, Aditya Ramesh, Gabriel Goh, Sandhini Agarwal, Girish Sastry, Amanda Askell, Pamela Mishkin, Jack Clark, Gretchen Krueger, and Ilya Sutskever. 2021.
\newblock \href {http://proceedings.mlr.press/v139/radford21a.html} {Learning transferable visual models from natural language supervision}.
\newblock In \emph{Proceedings of the 38th International Conference on Machine Learning, {ICML} 2021, 18-24 July 2021, Virtual Event}, volume 139 of \emph{Proceedings of Machine Learning Research}, pages 8748--8763. {PMLR}.

\bibitem[{Roach et~al.(2020)Roach, Ni, Kopylov, Lu, Xu, Zhang, Du, Zhou, Wu, Liu et~al.}]{roach2020canon}
Shane Roach, Connie Ni, Alexei Kopylov, Tsai-Ching Lu, Jiejun Xu, Si~Zhang, Boxin Du, Dawei Zhou, Jun Wu, Lihui Liu, et~al. 2020.
\newblock Canon: Complex analytics of network of networks for modeling adversarial activities.
\newblock In \emph{2020 IEEE International Conference on Big Data (Big Data)}, pages 1634--1643. IEEE.

\bibitem[{Sanh et~al.(2019)Sanh, Debut, Chaumond, and Wolf}]{DBLP:journals/corr/abs-1910-01108}
Victor Sanh, Lysandre Debut, Julien Chaumond, and Thomas Wolf. 2019.
\newblock \href {https://arxiv.org/abs/1910.01108} {Distilbert, a distilled version of {BERT:} smaller, faster, cheaper and lighter}.
\newblock \emph{CoRR}, abs/1910.01108.

\bibitem[{Shih et~al.(2022)Shih, Wang, Chang, Berry, Lee, and Harwath}]{DBLP:conf/slt/ShihWCBLH22}
Yi{-}Jen Shih, Hsuan{-}Fu Wang, Heng{-}Jui Chang, Layne Berry, Hung{-}yi Lee, and David Harwath. 2022.
\newblock \href {https://doi.org/10.1109/SLT54892.2023.10022954} {Speechclip: Integrating speech with pre-trained vision and language model}.
\newblock In \emph{{IEEE} Spoken Language Technology Workshop, {SLT} 2022, Doha, Qatar, January 9-12, 2023}, pages 715--722. {IEEE}.

\bibitem[{Singhal et~al.(2022)Singhal, Azizi, Tu, Mahdavi, Wei, Chung, Scales, Tanwani, Cole{-}Lewis, Pfohl, Payne, Seneviratne, Gamble, Kelly, Sch{\"{a}}rli, Chowdhery, Mansfield, y~Arcas, Webster, Corrado, Matias, Chou, Gottweis, Tomasev, Liu, Rajkomar, Barral, Semturs, Karthikesalingam, and Natarajan}]{DBLP:journals/corr/abs-2212-13138}
Karan Singhal, Shekoofeh Azizi, Tao Tu, S.~Sara Mahdavi, Jason Wei, Hyung~Won Chung, Nathan Scales, Ajay~Kumar Tanwani, Heather Cole{-}Lewis, Stephen Pfohl, Perry Payne, Martin Seneviratne, Paul Gamble, Chris Kelly, Nathaneal Sch{\"{a}}rli, Aakanksha Chowdhery, Philip~Andrew Mansfield, Blaise~Ag{\"{u}}era y~Arcas, Dale~R. Webster, Gregory~S. Corrado, Yossi Matias, Katherine Chou, Juraj Gottweis, Nenad Tomasev, Yun Liu, Alvin Rajkomar, Joelle~K. Barral, Christopher Semturs, Alan Karthikesalingam, and Vivek Natarajan. 2022.
\newblock \href {https://doi.org/10.48550/ARXIV.2212.13138} {Large language models encode clinical knowledge}.
\newblock \emph{CoRR}, abs/2212.13138.

\bibitem[{Skenderi et~al.(2023)Skenderi, Li, Tang, and Cristani}]{DBLP:journals/corr/abs-2309-16014}
Geri Skenderi, Hang Li, Jiliang Tang, and Marco Cristani. 2023.
\newblock \href {https://doi.org/10.48550/ARXIV.2309.16014} {Graph-level representation learning with joint-embedding predictive architectures}.
\newblock \emph{CoRR}, abs/2309.16014.

\bibitem[{Su et~al.(2022)Su, Du, Yang, Zhou, Li, Rao, Sun, Lu, and Wen}]{DBLP:journals/corr/abs-2209-05481}
Bing Su, Dazhao Du, Zhao Yang, Yujie Zhou, Jiangmeng Li, Anyi Rao, Hao Sun, Zhiwu Lu, and Ji{-}Rong Wen. 2022.
\newblock \href {https://doi.org/10.48550/ARXIV.2209.05481} {A molecular multimodal foundation model associating molecule graphs with natural language}.
\newblock \emph{CoRR}, abs/2209.05481.

\bibitem[{Sun et~al.(2020)Sun, Hoffmann, Verma, and Tang}]{DBLP:conf/iclr/Infograph}
Fan{-}Yun Sun, Jordan Hoffmann, Vikas Verma, and Jian Tang. 2020.
\newblock Infograph: Unsupervised and semi-supervised graph-level representation learning via mutual information maximization.
\newblock In \emph{8th International Conference on Learning Representations, {ICLR} 2020, Addis Ababa, Ethiopia, April 26-30, 2020}. OpenReview.net.

\bibitem[{Sun et~al.(2022)Sun, Zhou, He, Wang, and Wang}]{DBLP:conf/kdd/SunZHWW22}
Mingchen Sun, Kaixiong Zhou, Xin He, Ying Wang, and Xin Wang. 2022.
\newblock \href {https://doi.org/10.1145/3534678.3539249} {{GPPT:} graph pre-training and prompt tuning to generalize graph neural networks}.
\newblock In \emph{{KDD} '22: The 28th {ACM} {SIGKDD} Conference on Knowledge Discovery and Data Mining, Washington, DC, USA, August 14 - 18, 2022}, pages 1717--1727. {ACM}.

\bibitem[{Sun et~al.(2023{\natexlab{a}})Sun, Cheng, Li, Liu, and Guan}]{DBLP:conf/kdd/SunCLLG23}
Xiangguo Sun, Hong Cheng, Jia Li, Bo~Liu, and Jihong Guan. 2023{\natexlab{a}}.
\newblock \href {https://doi.org/10.1145/3580305.3599256} {All in one: Multi-task prompting for graph neural networks}.
\newblock In \emph{Proceedings of the 29th {ACM} {SIGKDD} Conference on Knowledge Discovery and Data Mining, {KDD} 2023, Long Beach, CA, USA, August 6-10, 2023}, pages 2120--2131. {ACM}.

\bibitem[{Sun et~al.(2023{\natexlab{b}})Sun, Zhang, Wu, Cheng, Xiong, and Li}]{DBLP:journals/corr/abs-2311-16534}
Xiangguo Sun, Jiawen Zhang, Xixi Wu, Hong Cheng, Yun Xiong, and Jia Li. 2023{\natexlab{b}}.
\newblock \href {https://doi.org/10.48550/ARXIV.2311.16534} {Graph prompt learning: {A} comprehensive survey and beyond}.
\newblock \emph{CoRR}, abs/2311.16534.

\bibitem[{Tan et~al.(2023)Tan, Guo, Ding, and Liu}]{DBLP:conf/kdd/TanGDL23}
Zhen Tan, Ruocheng Guo, Kaize Ding, and Huan Liu. 2023.
\newblock \href {https://doi.org/10.1145/3580305.3599541} {Virtual node tuning for few-shot node classification}.
\newblock In \emph{Proceedings of the 29th {ACM} {SIGKDD} Conference on Knowledge Discovery and Data Mining, {KDD} 2023, Long Beach, CA, USA, August 6-10, 2023}, pages 2177--2188. {ACM}.

\bibitem[{Taylor et~al.(2022)Taylor, Kardas, Cucurull, Scialom, Hartshorn, Saravia, Poulton, Kerkez, and Stojnic}]{DBLP:journals/corr/abs-2211-09085}
Ross Taylor, Marcin Kardas, Guillem Cucurull, Thomas Scialom, Anthony Hartshorn, Elvis Saravia, Andrew Poulton, Viktor Kerkez, and Robert Stojnic. 2022.
\newblock \href {https://doi.org/10.48550/ARXIV.2211.09085} {Galactica: {A} large language model for science}.
\newblock \emph{CoRR}, abs/2211.09085.

\bibitem[{Tieu et~al.(2025)Tieu, Fu, Wu, and He}]{tieu2025invariant}
Katherine Tieu, Dongqi Fu, Jun Wu, and Jingrui He. 2025.
\newblock Invariant link selector for spatial-temporal out-of-distribution problem.
\newblock In \emph{The 28th International Conference on Artificial Intelligence and Statistics}.

\bibitem[{Tieu et~al.(2024)Tieu, Fu, Zhu, Hamann, and He}]{DBLP:conf/nips/TieuFZHH24}
Katherine Tieu, Dongqi Fu, Yada Zhu, Hendrik~F. Hamann, and Jingrui He. 2024.
\newblock Temporal graph neural tangent kernel with graphon-guaranteed.
\newblock In \emph{NeurIPS}.

\bibitem[{Vaswani et~al.(2017)Vaswani, Shazeer, Parmar, Uszkoreit, Jones, Gomez, Kaiser, and Polosukhin}]{DBLP:conf/nips/VaswaniSPUJGKP17}
Ashish Vaswani, Noam Shazeer, Niki Parmar, Jakob Uszkoreit, Llion Jones, Aidan~N. Gomez, Lukasz Kaiser, and Illia Polosukhin. 2017.
\newblock \href {https://proceedings.neurips.cc/paper/2017/hash/3f5ee243547dee91fbd053c1c4a845aa-Abstract.html} {Attention is all you need}.
\newblock In \emph{Advances in Neural Information Processing Systems 30: Annual Conference on Neural Information Processing Systems 2017, December 4-9, 2017, Long Beach, CA, {USA}}, pages 5998--6008.

\bibitem[{Vignac et~al.(2022)Vignac, Krawczuk, Siraudin, Wang, Cevher, and Frossard}]{vignac2022digress}
Clement Vignac, Igor Krawczuk, Antoine Siraudin, Bohan Wang, Volkan Cevher, and Pascal Frossard. 2022.
\newblock Digress: Discrete denoising diffusion for graph generation.
\newblock \emph{arXiv preprint arXiv:2209.14734}.

\bibitem[{Wang et~al.(2023)Wang, Yan, Qiu, Zhu, Guan, Margenot, and Tong}]{wang2023networked}
Dingsu Wang, Yuchen Yan, Ruizhong Qiu, Yada Zhu, Kaiyu Guan, Andrew Margenot, and Hanghang Tong. 2023.
\newblock Networked time series imputation via position-aware graph enhanced variational autoencoders.
\newblock In \emph{Proceedings of the 29th ACM SIGKDD Conference on Knowledge Discovery and Data Mining}, pages 2256--2268.

\bibitem[{Wang et~al.(2025)Wang, Hassani, Zhang, Fu, Yuan, Cong, Hua, Wu, Yao, and Long}]{DBLP:conf/iclr/WangHZFYCHWYL25}
Limei Wang, Kaveh Hassani, Si~Zhang, Dongqi Fu, Baichuan Yuan, Weilin Cong, Zhigang Hua, Hao Wu, Ning Yao, and Bo~Long. 2025.
\newblock Learning graph quantized tokenizers.
\newblock In \emph{{ICLR}}.

\bibitem[{Wei et~al.(2021)Wei, Feng, Chen, Wu, Yi, and He}]{wei2021model}
Tianxin Wei, Fuli Feng, Jiawei Chen, Ziwei Wu, Jinfeng Yi, and Xiangnan He. 2021.
\newblock Model-agnostic counterfactual reasoning for eliminating popularity bias in recommender system.
\newblock In \emph{Proceedings of the 27th ACM SIGKDD conference on knowledge discovery \& data mining}, pages 1791--1800.

\bibitem[{Wei et~al.(2024)Wei, Jin, Li, Zeng, Wang, Sun, Yin, Lu, Wang, He et~al.}]{wei2024towards}
Tianxin Wei, Bowen Jin, Ruirui Li, Hansi Zeng, Zhengyang Wang, Jianhui Sun, Qingyu Yin, Hanqing Lu, Suhang Wang, Jingrui He, et~al. 2024.
\newblock Towards unified multi-modal personalization: Large vision-language models for generative recommendation and beyond.
\newblock \emph{arXiv preprint arXiv:2403.10667}.

\bibitem[{Wei et~al.(2020)Wei, Wu, Li, Hu, Feng, He, Sun, and Wang}]{wei2020fast}
Tianxin Wei, Ziwei Wu, Ruirui Li, Ziniu Hu, Fuli Feng, Xiangnan He, Yizhou Sun, and Wei Wang. 2020.
\newblock Fast adaptation for cold-start collaborative filtering with meta-learning.
\newblock In \emph{2020 IEEE International Conference on Data Mining (ICDM)}, pages 661--670. IEEE.

\bibitem[{Wei et~al.(2022)Wei, You, Chen, Shen, He, and Wang}]{wei2022augmentations}
Tianxin Wei, Yuning You, Tianlong Chen, Yang Shen, Jingrui He, and Zhangyang Wang. 2022.
\newblock Augmentations in hypergraph contrastive learning: Fabricated and generative.
\newblock \emph{Advances in neural information processing systems}, 35:1909--1922.

\bibitem[{Wen and Fang(2023)}]{DBLP:conf/sigir/Wen023}
Zhihao Wen and Yuan Fang. 2023.
\newblock \href {https://doi.org/10.1145/3539618.3591641} {Augmenting low-resource text classification with graph-grounded pre-training and prompting}.
\newblock In \emph{Proceedings of the 46th International {ACM} {SIGIR} Conference on Research and Development in Information Retrieval, {SIGIR} 2023, Taipei, Taiwan, July 23-27, 2023}, pages 506--516. {ACM}.

\bibitem[{Wu et~al.(2023{\natexlab{a}})Wu, He, and Ainsworth}]{wu2023non}
Jun Wu, Jingrui He, and Elizabeth Ainsworth. 2023{\natexlab{a}}.
\newblock Non-iid transfer learning on graphs.
\newblock In \emph{Proceedings of the AAAI Conference on Artificial Intelligence}, volume~37, pages 10342--10350.

\bibitem[{Wu et~al.(2023{\natexlab{b}})Wu, Zhou, Sun, Wang, and Liu}]{DBLP:journals/corr/abs-2303-07275}
Xuansheng Wu, Kaixiong Zhou, Mingchen Sun, Xin Wang, and Ninghao Liu. 2023{\natexlab{b}}.
\newblock \href {https://doi.org/10.48550/ARXIV.2303.07275} {A survey of graph prompting methods: Techniques, applications, and challenges}.
\newblock \emph{CoRR}, abs/2303.07275.

\bibitem[{Wu et~al.(2017)Wu, Ramsundar, Feinberg, Gomes, Geniesse, Pappu, Leswing, and Pande}]{DBLP:journals/corr/WuRFGGPLP17}
Zhenqin Wu, Bharath Ramsundar, Evan~N. Feinberg, Joseph Gomes, Caleb Geniesse, Aneesh~S. Pappu, Karl Leswing, and Vijay~S. Pande. 2017.
\newblock \href {https://arxiv.org/abs/1703.00564} {Moleculenet: {A} benchmark for molecular machine learning}.
\newblock \emph{CoRR}, abs/1703.00564.

\bibitem[{Xia et~al.(2022)Xia, Wu, Chen, Hu, and Li}]{DBLP:conf/www/XiaWCHL22}
Jun Xia, Lirong Wu, Jintao Chen, Bozhen Hu, and Stan~Z. Li. 2022.
\newblock Simgrace: {A} simple framework for graph contrastive learning without data augmentation.
\newblock In \emph{{WWW} '22: The {ACM} Web Conference 2022, Virtual Event, Lyon, France, April 25 - 29, 2022}, pages 1070--1079. {ACM}.

\bibitem[{Xu et~al.()Xu, Yan, Wang, Xu, Zeng, Abdelzaher, Han, and Tong}]{xu2024slog}
Haobo Xu, Yuchen Yan, Dingsu Wang, Zhe Xu, Zhichen Zeng, Tarek~F Abdelzaher, Jiawei Han, and Hanghang Tong.
\newblock Slog: An inductive spectral graph neural network beyond polynomial filter.
\newblock In \emph{Forty-first International Conference on Machine Learning}.

\bibitem[{Xu et~al.(2021)Xu, Ghosh, Huang, Okhonko, Aghajanyan, Metze, Zettlemoyer, and Feichtenhofer}]{DBLP:conf/emnlp/XuG0OAMZF21}
Hu~Xu, Gargi Ghosh, Po{-}Yao Huang, Dmytro Okhonko, Armen Aghajanyan, Florian Metze, Luke Zettlemoyer, and Christoph Feichtenhofer. 2021.
\newblock \href {https://doi.org/10.18653/V1/2021.EMNLP-MAIN.544} {Videoclip: Contrastive pre-training for zero-shot video-text understanding}.
\newblock In \emph{Proceedings of the 2021 Conference on Empirical Methods in Natural Language Processing, {EMNLP} 2021, Virtual Event / Punta Cana, Dominican Republic, 7-11 November, 2021}, pages 6787--6800. Association for Computational Linguistics.

\bibitem[{Xu et~al.(2024)Xu, Qiu, Chen, Chen, Fan, Pan, Zeng, Das, and Tong}]{xu2024discrete}
Zhe Xu, Ruizhong Qiu, Yuzhong Chen, Huiyuan Chen, Xiran Fan, Menghai Pan, Zhichen Zeng, Mahashweta Das, and Hanghang Tong. 2024.
\newblock Discrete-state continuous-time diffusion for graph generation.
\newblock \emph{arXiv preprint arXiv:2405.11416}.

\bibitem[{Yan et~al.(2023)Yan, Li, Long, Yan, Zhao, Zhuang, Yin, Zhang, Han, Sun, Deng, Zhang, Sun, Xie, and Wang}]{DBLP:conf/nips/YanLLY0ZYZHSDZ023}
Hao Yan, Chaozhuo Li, Ruosong Long, Chao Yan, Jianan Zhao, Wenwen Zhuang, Jun Yin, Peiyan Zhang, Weihao Han, Hao Sun, Weiwei Deng, Qi~Zhang, Lichao Sun, Xing Xie, and Senzhang Wang. 2023.
\newblock \href {http://papers.nips.cc/paper\_files/paper/2023/hash/37d00f567a18b478065f1a91b95622a0-Abstract-Datasets\_and\_Benchmarks.html} {A comprehensive study on text-attributed graphs: Benchmarking and rethinking}.
\newblock In \emph{Advances in Neural Information Processing Systems 36: Annual Conference on Neural Information Processing Systems 2023, NeurIPS 2023, New Orleans, LA, USA, December 10 - 16, 2023}.

\bibitem[{Yan et~al.(2024{\natexlab{a}})Yan, Chen, Chen, Li, Xu, Zeng, Liu, Liu, and Tong}]{yan2024thegcn}
Yuchen Yan, Yuzhong Chen, Huiyuan Chen, Xiaoting Li, Zhe Xu, Zhichen Zeng, Lihui Liu, Zhining Liu, and Hanghang Tong. 2024{\natexlab{a}}.
\newblock Thegcn: Temporal heterophilic graph convolutional network.
\newblock \emph{arXiv preprint arXiv:2412.16435}.

\bibitem[{Yan et~al.(2024{\natexlab{b}})Yan, Hu, Zhou, Liu, Zeng, Chen, Pan, Chen, Das, and Tong}]{yan2024pacer}
Yuchen Yan, Yongyi Hu, Qinghai Zhou, Lihui Liu, Zhichen Zeng, Yuzhong Chen, Menghai Pan, Huiyuan Chen, Mahashweta Das, and Hanghang Tong. 2024{\natexlab{b}}.
\newblock Pacer: Network embedding from positional to structural.
\newblock In \emph{Proceedings of the ACM Web Conference 2024}, pages 2485--2496.

\bibitem[{Yan et~al.(2024{\natexlab{c}})Yan, Hu, Zhou, Wu, Wang, and Tong}]{yan2024topological}
Yuchen Yan, Yongyi Hu, Qinghai Zhou, Shurang Wu, Dingsu Wang, and Hanghang Tong. 2024{\natexlab{c}}.
\newblock Topological anonymous walk embedding: A new structural node embedding approach.
\newblock In \emph{Proceedings of the 33rd ACM International Conference on Information and Knowledge Management}, pages 2796--2806.

\bibitem[{Yan et~al.(2021)Yan, Liu, Ban, Jing, and Tong}]{yan2021dynamic}
Yuchen Yan, Lihui Liu, Yikun Ban, Baoyu Jing, and Hanghang Tong. 2021.
\newblock Dynamic knowledge graph alignment.
\newblock In \emph{Proceedings of the AAAI conference on artificial intelligence}, volume~35, pages 4564--4572.

\bibitem[{Yan et~al.(2022)Yan, Zhou, Li, Abdelzaher, and Tong}]{yan2022dissecting}
Yuchen Yan, Qinghai Zhou, Jinning Li, Tarek Abdelzaher, and Hanghang Tong. 2022.
\newblock Dissecting cross-layer dependency inference on multi-layered inter-dependent networks.
\newblock In \emph{Proceedings of the 31st ACM International Conference on Information \& Knowledge Management}, pages 2341--2351.

\bibitem[{Yang et~al.(2021)Yang, Liu, Xiao, Li, Lian, Agrawal, Singh, Sun, and Xie}]{yang2021graphformers}
Junhan Yang, Zheng Liu, Shitao Xiao, Chaozhuo Li, Defu Lian, Sanjay Agrawal, Amit Singh, Guangzhong Sun, and Xing Xie. 2021.
\newblock Graphformers: Gnn-nested transformers for representation learning on textual graph.
\newblock \emph{Advances in Neural Information Processing Systems}, 34:28798--28810.

\bibitem[{Yang et~al.(2016)Yang, Cohen, and Salakhutdinov}]{DBLP:conf/icml/YangCS16}
Zhilin Yang, William~W. Cohen, and Ruslan Salakhutdinov. 2016.
\newblock \href {http://proceedings.mlr.press/v48/yanga16.html} {Revisiting semi-supervised learning with graph embeddings}.
\newblock In \emph{Proceedings of the 33nd International Conference on Machine Learning, {ICML} 2016, New York City, NY, USA, June 19-24, 2016}, volume~48 of \emph{{JMLR} Workshop and Conference Proceedings}, pages 40--48. JMLR.org.

\bibitem[{Yoo et~al.(2024)Yoo, Zeng, Kang, Qiu, Zhou, Liu, Wang, Xu, Chan, and Tong}]{yoo2024ensuring}
Hyunsik Yoo, Zhichen Zeng, Jian Kang, Ruizhong Qiu, David Zhou, Zhining Liu, Fei Wang, Charlie Xu, Eunice Chan, and Hanghang Tong. 2024.
\newblock Ensuring user-side fairness in dynamic recommender systems.
\newblock In \emph{Proceedings of the ACM Web Conference 2024}, pages 3667--3678.

\bibitem[{You et~al.(2020)You, Chen, Sui, Chen, Wang, and Shen}]{DBLP:conf/nips/YouCSCWS20}
Yuning You, Tianlong Chen, Yongduo Sui, Ting Chen, Zhangyang Wang, and Yang Shen. 2020.
\newblock \href {https://proceedings.neurips.cc/paper/2020/hash/3fe230348e9a12c13120749e3f9fa4cd-Abstract.html} {Graph contrastive learning with augmentations}.
\newblock In \emph{Advances in Neural Information Processing Systems 33: Annual Conference on Neural Information Processing Systems 2020, NeurIPS 2020, December 6-12, 2020, virtual}.

\bibitem[{Yu et~al.(2025)Yu, Zeng, Yan, Ying, Srikant, and Tong}]{yu2025joint}
Qi~Yu, Zhichen Zeng, Yuchen Yan, Lei Ying, R~Srikant, and Hanghang Tong. 2025.
\newblock Joint optimal transport and embedding for network alignment.
\newblock In \emph{Proceedings of the ACM on Web Conference 2025}, pages 2064--2075.

\bibitem[{Yun et~al.(2019)Yun, Jeong, Kim, Kang, and Kim}]{DBLP:conf/nips/YunJKKK19}
Seongjun Yun, Minbyul Jeong, Raehyun Kim, Jaewoo Kang, and Hyunwoo~J. Kim. 2019.
\newblock \href {https://proceedings.neurips.cc/paper/2019/hash/9d63484abb477c97640154d40595a3bb-Abstract.html} {Graph transformer networks}.
\newblock In \emph{Advances in Neural Information Processing Systems 32: Annual Conference on Neural Information Processing Systems 2019, NeurIPS 2019, December 8-14, 2019, Vancouver, BC, Canada}, pages 11960--11970.

\bibitem[{Zeng et~al.(2022)Zeng, Yao, Liu, and Sun}]{zeng2022deep}
Zheni Zeng, Yuan Yao, Zhiyuan Liu, and Maosong Sun. 2022.
\newblock A deep-learning system bridging molecule structure and biomedical text with comprehension comparable to human professionals.
\newblock \emph{Nature communications}, 13(1):862.

\bibitem[{Zeng et~al.(2024)Zeng, Du, Zhang, Xia, Liu, and Tong}]{zeng2024hierarchical}
Zhichen Zeng, Boxin Du, Si~Zhang, Yinglong Xia, Zhining Liu, and Hanghang Tong. 2024.
\newblock Hierarchical multi-marginal optimal transport for network alignment.
\newblock In \emph{Proceedings of the AAAI Conference on Artificial Intelligence}, volume~38, pages 16660--16668.

\bibitem[{Zeng et~al.(2025)Zeng, Qiu, Bao, Wei, Lin, Yan, Abdelzaher, Han, and Tong}]{zeng2025pave}
Zhichen Zeng, Ruizhong Qiu, Wenxuan Bao, Tianxin Wei, Xiao Lin, Yuchen Yan, Tarek~F Abdelzaher, Jiawei Han, and Hanghang Tong. 2025.
\newblock Pave your own path: Graph gradual domain adaptation on fused gromov-wasserstein geodesics.
\newblock \emph{arXiv preprint arXiv:2505.12709}.

\bibitem[{Zeng et~al.(2023)Zeng, Zhu, Xia, Zeng, and Tong}]{zeng2023generative}
Zhichen Zeng, Ruike Zhu, Yinglong Xia, Hanqing Zeng, and Hanghang Tong. 2023.
\newblock Generative graph dictionary learning.
\newblock In \emph{International Conference on Machine Learning}, pages 40749--40769. PMLR.

\bibitem[{Zhang et~al.(2022{\natexlab{a}})Zhang, Li, Chen, Deng, Bi, Tan, Huang, and Chen}]{DBLP:conf/iclr/ZhangLCDBTHC22}
Ningyu Zhang, Luoqiu Li, Xiang Chen, Shumin Deng, Zhen Bi, Chuanqi Tan, Fei Huang, and Huajun Chen. 2022{\natexlab{a}}.
\newblock \href {https://openreview.net/forum?id=ek9a0qIafW} {Differentiable prompt makes pre-trained language models better few-shot learners}.
\newblock In \emph{The Tenth International Conference on Learning Representations, {ICLR} 2022, Virtual Event, April 25-29, 2022}. OpenReview.net.

\bibitem[{Zhang et~al.(2022{\natexlab{b}})Zhang, Bosselut, Yasunaga, Ren, Liang, Manning, and Leskovec}]{zhang2022greaselm}
Xikun Zhang, Antoine Bosselut, Michihiro Yasunaga, Hongyu Ren, Percy Liang, Christopher~D Manning, and Jure Leskovec. 2022{\natexlab{b}}.
\newblock Greaselm: Graph reasoning enhanced language models for question answering.
\newblock \emph{arXiv preprint arXiv:2201.08860}.

\bibitem[{Zhao et~al.(2023)Zhao, Liu, Ma, Xu, Fu, Deng, Kong, and Liu}]{DBLP:conf/nips/ZhaoLMXFDKL23}
Haiteng Zhao, Shengchao Liu, Chang Ma, Hannan Xu, Jie Fu, Zhihong Deng, Lingpeng Kong, and Qi~Liu. 2023.
\newblock \href {http://papers.nips.cc/paper\_files/paper/2023/hash/129033c7c08be683059559e8d6bfd460-Abstract-Conference.html} {{GIMLET:} {A} unified graph-text model for instruction-based molecule zero-shot learning}.
\newblock In \emph{Advances in Neural Information Processing Systems 36: Annual Conference on Neural Information Processing Systems 2023, NeurIPS 2023, New Orleans, LA, USA, December 10 - 16, 2023}.

\bibitem[{Zhao et~al.(2024)Zhao, Yang, Cen, Ren, Zhang, Dong, Kharlamov, Zhao, and Tang}]{DBLP:conf/kdd/ZhaoYCRZDKZ024}
Huanjing Zhao, Beining Yang, Yukuo Cen, Junyu Ren, Chenhui Zhang, Yuxiao Dong, Evgeny Kharlamov, Shu Zhao, and Jie Tang. 2024.
\newblock \href {https://doi.org/10.1145/3637528.3671952} {Pre-training and prompting for few-shot node classification on text-attributed graphs}.
\newblock In \emph{Proceedings of the 30th {ACM} {SIGKDD} Conference on Knowledge Discovery and Data Mining, {KDD} 2024, Barcelona, Spain, August 25-29, 2024}, pages 4467--4478. {ACM}.

\bibitem[{Zheng et~al.(2025)Zheng, Birge, Wu, Zhang, and He}]{DBLP:conf/www/ZhengBWZH25}
Lecheng Zheng, John~R. Birge, Haiyue Wu, Yifang Zhang, and Jingrui He. 2025.
\newblock Cluster aware graph anomaly detection.
\newblock In \emph{Proceedings of the {ACM} on Web Conference 2025, {WWW} 2025, Sydney, NSW, Australia, 28 April 2025- 2 May 2025}, pages 1771--1782. {ACM}.

\bibitem[{Zheng et~al.(2024{\natexlab{a}})Zheng, Chen, He, and Chen}]{DBLP:conf/www/ZhengCHC24}
Lecheng Zheng, Zhengzhang Chen, Jingrui He, and Haifeng Chen. 2024{\natexlab{a}}.
\newblock {MULAN:} multi-modal causal structure learning and root cause analysis for microservice systems.
\newblock In \emph{Proceedings of the {ACM} on Web Conference 2024, {WWW} 2024, Singapore, May 13-17, 2024}, pages 4107--4116. {ACM}.

\bibitem[{Zheng et~al.(2021)Zheng, Cheng, Yang, Cao, and He}]{DBLP:conf/www/ZhengCYCH21}
Lecheng Zheng, Yu~Cheng, Hongxia Yang, Nan Cao, and Jingrui He. 2021.
\newblock Deep co-attention network for multi-view subspace learning.
\newblock In \emph{{WWW} '21: The Web Conference 2021, Virtual Event / Ljubljana, Slovenia, April 19-23, 2021}, pages 1528--1539. {ACM} / {IW3C2}.

\bibitem[{Zheng et~al.(2024{\natexlab{b}})Zheng, Fu, Maciejewski, and He}]{zheng2024drgnn}
Lecheng Zheng, Dongqi Fu, Ross Maciejewski, and Jingrui He. 2024{\natexlab{b}}.
\newblock Drgnn: Deep residual graph neural network with contrastive learning.
\newblock \emph{Transactions on Machine Learning Research}.

\bibitem[{Zheng et~al.(2024{\natexlab{c}})Zheng, Jing, Li, Tong, and He}]{DBLP:conf/kdd/ZhengJLTH24}
Lecheng Zheng, Baoyu Jing, Zihao Li, Hanghang Tong, and Jingrui He. 2024{\natexlab{c}}.
\newblock Heterogeneous contrastive learning for foundation models and beyond.
\newblock In \emph{Proceedings of the 30th {ACM} {SIGKDD} Conference on Knowledge Discovery and Data Mining, {KDD} 2024, Barcelona, Spain, August 25-29, 2024}, pages 6666--6676. {ACM}.

\bibitem[{Zheng et~al.(2024{\natexlab{d}})Zheng, Jing, Li, Zeng, Wei, Ai, He, Liu, Fu, You, Tong, and He}]{DBLP:journals/corr/abs-2412-21151}
Lecheng Zheng, Baoyu Jing, Zihao Li, Zhichen Zeng, Tianxin Wei, Mengting Ai, Xinrui He, Lihui Liu, Dongqi Fu, Jiaxuan You, Hanghang Tong, and Jingrui He. 2024{\natexlab{d}}.
\newblock \href {https://doi.org/10.48550/ARXIV.2412.21151} {Pyg-ssl: {A} graph self-supervised learning toolkit}.
\newblock \emph{CoRR}, abs/2412.21151.

\bibitem[{Zheng et~al.(2024{\natexlab{e}})Zheng, Zhou, Tong, Xu, Zhu, and He}]{DBLP:conf/icde/Zheng0TXZH24}
Lecheng Zheng, Dawei Zhou, Hanghang Tong, Jiejun Xu, Yada Zhu, and Jingrui He. 2024{\natexlab{e}}.
\newblock Fairgen: Towards fair graph generation.
\newblock In \emph{40th {IEEE} International Conference on Data Engineering, {ICDE} 2024, Utrecht, The Netherlands, May 13-16, 2024}, pages 2285--2297. {IEEE}.

\bibitem[{Zheng et~al.(2023)Zheng, Zhu, and He}]{DBLP:conf/sdm/ZhengZH23}
Lecheng Zheng, Yada Zhu, and Jingrui He. 2023.
\newblock Fairness-aware multi-view clustering.
\newblock In \emph{Proceedings of the 2023 {SIAM} International Conference on Data Mining, {SDM} 2023, Minneapolis-St. Paul Twin Cities, MN, USA, April 27-29, 2023}, pages 856--864. {SIAM}.

\bibitem[{Zhou et~al.(2022{\natexlab{a}})Zhou, Zheng, Fu, Han, and He}]{DBLP:conf/cikm/ZhouZF0H22}
Dawei Zhou, Lecheng Zheng, Dongqi Fu, Jiawei Han, and Jingrui He. 2022{\natexlab{a}}.
\newblock Mentorgnn: Deriving curriculum for pre-training gnns.
\newblock In \emph{Proceedings of the 31st {ACM} International Conference on Information {\&} Knowledge Management, Atlanta, GA, USA, October 17-21, 2022}, pages 2721--2731. {ACM}.

\bibitem[{Zhou et~al.(2020)Zhou, Zheng, Han, and He}]{DBLP:conf/kdd/ZhouZ0H20}
Dawei Zhou, Lecheng Zheng, Jiawei Han, and Jingrui He. 2020.
\newblock A data-driven graph generative model for temporal interaction networks.
\newblock In \emph{{KDD} '20: The 26th {ACM} {SIGKDD} Conference on Knowledge Discovery and Data Mining, Virtual Event, CA, USA, August 23-27, 2020}, pages 401--411. {ACM}.

\bibitem[{Zhou et~al.(2019)Zhou, Zheng, Xu, and He}]{DBLP:journals/fdata/ZhouZXH19}
Dawei Zhou, Lecheng Zheng, Jiejun Xu, and Jingrui He. 2019.
\newblock Misc-gan: {A} multi-scale generative model for graphs.
\newblock \emph{Frontiers Big Data}, 2:3.

\bibitem[{Zhou et~al.(2022{\natexlab{b}})Zhou, Yang, Loy, and Liu}]{DBLP:conf/cvpr/ZhouYL022}
Kaiyang Zhou, Jingkang Yang, Chen~Change Loy, and Ziwei Liu. 2022{\natexlab{b}}.
\newblock \href {https://doi.org/10.1109/CVPR52688.2022.01631} {Conditional prompt learning for vision-language models}.
\newblock In \emph{{IEEE/CVF} Conference on Computer Vision and Pattern Recognition, {CVPR} 2022, New Orleans, LA, USA, June 18-24, 2022}, pages 16795--16804. {IEEE}.

\bibitem[{Zhu et~al.(2021)Zhu, Cui, Liu, Sun, Li, Pelger, Yang, Zhang, Zhang, and Zhao}]{zhu2021textgnn}
Jason Zhu, Yanling Cui, Yuming Liu, Hao Sun, Xue Li, Markus Pelger, Tianqi Yang, Liangjie Zhang, Ruofei Zhang, and Huasha Zhao. 2021.
\newblock Textgnn: Improving text encoder via graph neural network in sponsored search.
\newblock In \emph{Proceedings of the Web Conference 2021}, pages 2848--2857.

\bibitem[{Zhu et~al.(2023)Zhu, Guo, and Tang}]{DBLP:journals/corr/abs-2302-12449}
Yun Zhu, Jianhao Guo, and Siliang Tang. 2023.
\newblock \href {https://doi.org/10.48550/ARXIV.2302.12449} {{SGL-PT:} {A} strong graph learner with graph prompt tuning}.
\newblock \emph{CoRR}, abs/2302.12449.

\bibitem[{Zou et~al.(2025)Zou, Fu, Chen, He, Li, Zhu, Han, and He}]{DBLP:journals/corr/abs-2504-01346}
Jiaru Zou, Dongqi Fu, Sirui Chen, Xinrui He, Zihao Li, Yada Zhu, Jiawei Han, and Jingrui He. 2025.
\newblock {GTR:} graph-table-rag for cross-table question answering.
\newblock \emph{CoRR}.

\end{thebibliography}
